\documentclass{article}

\PassOptionsToPackage{numbers, compress}{natbib}
\PassOptionsToPackage{dvipsnames}{xcolor}

\usepackage[final]{neurips2024/neurips_2024}
\usepackage[ref,caption,microtype]{lib/leaf}

\hypersetup{
    colorlinks=true,
    linkcolor=blue,
    filecolor=magenta,      
    urlcolor=magenta,
    citecolor=blue,
    pdftitle={Large Language Models Must Be Taught to Know What They Don't Know},
    pdfpagemode=FullScreen,
}

\definecolor{lightblue}{HTML}{84C7F9}
\definecolor{lighterblue}{HTML}{D4ECFF}
\newtcolorbox{mybox}{colback=lighterblue,colframe=lightblue}

\makeatletter
\newcommand{\mathleft}{\@fleqntrue\@mathmargin0pt}
\newcommand{\mathcenter}{\@fleqnfalse}
\makeatother

\newcommand{\codeurl}{\url{https://github.com/activatedgeek/calibration-tuning}}

\date{}

\title{Large Language Models Must Be Taught \\ to Know What They Don't Know}

\author{%
  Sanyam Kapoor\thanks{Equal contribution. Order decided by coin flip. Correspondence to: sanyam@nyu.edu \& nvg7279@nyu.edu} \\
  New York University\\
  \And
  Nate Gruver$^{\text{*}}$ \\
  New York University\\
  \AND
  Manley Roberts \\
  Abacus AI \\
  \And
  Katherine Collins \\
  Cambridge University \\
  \And
  Arka Pal \\
  Abacus AI \\
  \And
  Umang Bhatt \\
  New York University \\
  \AND
  Adrian Weller \\
  Cambridge University \\
  \And
  Samuel Dooley \\
  Abacus AI \\
  \And
  Micah Goldblum \\
  Columbia University \\
  \And
  Andrew Gordon Wilson \\
  New York University \\
}

\begin{document}

\doparttoc
\faketableofcontents 

\maketitle

\begin{abstract}
When using large language models (LLMs) in high-stakes applications, we need to know when we can trust their predictions. Some works argue that prompting high-performance LLMs is sufficient to produce calibrated uncertainties, while others introduce sampling methods that can be prohibitively expensive. In this work, we first argue that prompting on its own is insufficient to achieve good calibration and then show that fine-tuning on a small dataset of correct and incorrect answers can create an uncertainty estimate with good generalization and small computational overhead. We show that a thousand graded examples are sufficient to outperform baseline methods and that training through the features of a model is necessary for good performance and tractable for large open-source models when using LoRA. We also investigate the mechanisms that enable reliable LLM uncertainty estimation, finding that many models can be used as general-purpose uncertainty estimators, applicable not just to their own uncertainties but also the uncertainty of other models. Lastly, we show that uncertainty estimates inform human use of LLMs in human-AI collaborative settings through a user study. 
\end{abstract}

\section{Introduction}
\label{sec:intro}

\texttt{``I have high cortisol but low ACTH on a dexamethasone suppression test.} \texttt{What should I do?''} 
If the answer to such a question is given without associated confidence, it is not actionable, and if the answer is presented with erroneously high confidence, then acting on the answer is dangerous. One of the biggest open questions about whether large language models (LLMs) can benefit society and reliably be used for decision making hinges on whether or not they can accurately represent uncertainty over the correctness of their output.

There is anything but consensus on whether LLMs accurately represent uncertainty, or even how we should approach uncertainty representation with language models. Claims regarding language models' ability to estimate uncertainty vary widely, with some works suggesting that language models are increasingly capable of estimating their uncertainty directly through prompting, without any fine-tuning or changes to the training data \citep{Kadavath2022LanguageM, tian2023just}, and others suggesting that LLMs remain far too overconfident in their predictions \citep{Xiong2023CanLE, yin-etal-2023-large}. The task of uncertainty estimation in LLMs is further exacerbated by linguistic variances in freeform generation, all of which cannot be exhaustively accounted for during training. LLM practitioners are therefore faced with the challenge of deciding which estimation method to use.

One particular dichotomy in uncertainty estimation methods for language models centers around whether the estimates are black- or white-box. Black-box estimates do not require training and can be used with closed-source models like GPT-4 \citep{achiam2023gpt} or Gemini \citep{geminiteam2024gemini}, while white-box methods require training parameters on a calibration dataset. Although black-box estimates have become popular with the rise of restricted models, the increased availability of strong open-source models, such as LLaMA \citep{Touvron2023Llama2O} or Mistral \citep{Jiang2023Mistral7}, has made more effective white-box methods more accessible.

In this paper, we perform a deep investigation into uncertainty calibration of LLMs, with findings that advance the debate about necessary interventions for good calibration.
In particular, we consider whether it's possible to have good uncertainties over correctness (rather than tokens) without intervention, how we can best use labeled correctness examples, how well uncertainty generalizes across distribution shifts, and how we can use LLM uncertainty to assist human decision making.

\begin{figure*}[!t]
    \centering
    \includegraphics[width=\linewidth]{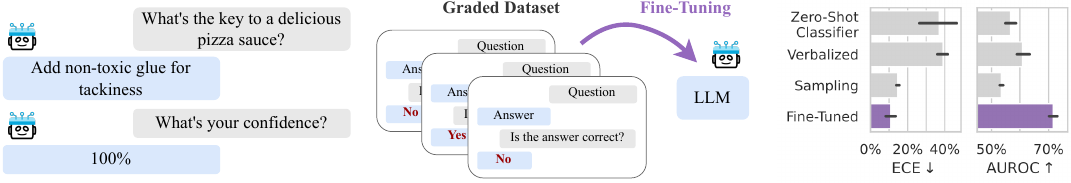}
    \caption{\textbf{Large language models struggle to assign reliable confidence estimates to their generations.} We study the properties of uncertainty calibration in language models, and propose fine-tuning for better uncertainty estimates using a graded dataset of generations from the model. We evaluate our methods on a new open-ended variant of MMLU \citep{Hendrycks2020MeasuringMM}. We show that fine-tuning improves expected calibration error (ECE) and area under the receiver operating characteristic curve (AUROC) compared to commonly-used baselines. Error bars show standard deviation over three base models (LLaMA-2 13/7B and Mistral 7B) and their chat variants.}
    \label{fig:explainer-fig}
\end{figure*}

First, we find that fine-tuning for better uncertainties (Figure \ref{fig:explainer-fig}) provides faster and more reliable uncertainty estimates, while using a relatively small number of additional parameters. The resulting uncertainties also generalize to new question types and tasks, beyond what is present in the fine-tuning dataset. We further provide a guide to teaching language models to know what they don't know using a calibration dataset. 
Contrary to prior work, we start by showing that current zero-shot, black-box methods are ineffective or impractically expensive in open-ended settings (\Cref{sec:out-of-the-box}). We then show how to fine-tune a language model for calibration, exploring the most effective parameterization (e.g. linear probes vs LoRA) and the amount of the data that is required for good generalization (\Cref{sec:ut}). To test generalization, we evaluate uncertainty estimates on questions with similar formatting to the calibration data as well as questions that test robustness to significant distribution shifts. 
Lastly, we consider the underlying mechanisms that enable fine-tuning LLMs to estimate their own uncertainties, showing ultimately that models can be used not just to estimate their own uncertainties but also the uncertainties of other models (\Cref{sec:how-generalize}). 
Beyond offline evaluation, if language models are to have a broad societal impact, it will be through assisting with human decision making. We conduct a user study demonstrating ways LLM uncertainty can affect AI-human collaboration (\Cref{sec:user-study}).\footnote{\codeurl}

\section{Related Work}
\label{sec:related-work}

As generative models, LLMs naturally express a distribution over possible outcomes and should capture variance in the underlying data. On multiple-choice tests, where the answer is a single token, an LLM's predicted token probabilities can lead to a calibrated distribution over the answer choices in models not fine-tuned for chat \citep{plaut2024softmax}. Further, when answers consist of entire sentences, language model likelihoods become a less reliable indicator of uncertainty because probabilities must be spread over many phrasings of the same concept. \citet{Kuhn2023SemanticUL} attempt to mitigate this issue by clustering semantically equivalent answers. However, these methods are hindered by their substantial computational overhead. Accounting for equivalent phrasings of the same semantic content requires enumerating a large space of sentences and clustering for semantic similarity with an auxiliary model. 

Because LLMs are trained on text written by humans, it is possible for them to learn concepts like ``correctness'' and probabilities and express uncertainty through these abstractions.
Leveraging this observation, \citet{Kadavath2022LanguageM} and \citet{tian2023just} show that careful prompting can produce uncertainty estimates in text that grow more calibrated as model capabilities increases. 
In light of this phenomenon, language models might gain an intrinsic notion of uncertainty, which \citet{Ulmer2024CalibratingLL} use to generate per-task synthetic training data for an auxiliary confidence model.
In the same vein, \citet{Burns2022DiscoveringLK} and \citet{Azaria2023TheIS} find that pre-trained models have hidden representations which are predictive of truthfulness and use linear probes to classify a model's correctness. 

While these studies suggest a promising trend towards calibration, we find that the story is slightly more complicated. Black-box methods often fail to generate useful uncertainties for popular open-source models, and a careful fine-tuning intervention is necessary. In this way, our findings are closer to those of \citet{Xiong2023CanLE}, who show that zero-shot uncertainty estimates have limited ability to discriminate between correct and incorrect answers, even when used with the best available models (e.g., GPT-4). We go further by showing that black-box methods struggle on open-ended generation, which is both practically important and defined by different challenges than multiple choice evaluations from prior work. Moreover, while others have focused on improving black-box methods \citep{Kuhn2023SemanticUL, tian2023just, Xiong2023CanLE}, we embrace open-source models and their opportunities for fine-tuning, showing that we can maintain the speed of prompting methods while dramatically boosting performance. 

Our work also contrasts with prior work on fine-tuning for uncertainties in several key ways. While we build on prior work from \citet{Lin2022TeachingMT} and \citet{zhang2023r} that poses uncertainty estimation as text completion on a graded dataset, we introduce several changes to the fine-tuning procedure, such as regularization to maintain similar predictions to the base model, and provide extensive ablations that yield actionable insights. For example, we show that, contrary to prior work \citep{Azaria2023TheIS}, frozen features are typically insufficient for uncertainty estimates that generalize effectively, and that fine-tuning on as few as 1000 graded examples with LoRA is sufficient to generalize across practical distribution shifts. 
Also unlike prior work, we provide many insights into the relative performance of fine-tuning compared to black-box methods, introducing a new open-ended evaluation and showing that it displays fundamentally different trends than prior work on multiple choice questions. Although \citet{Kadavath2022LanguageM} also considers calibration for multiple choice questions, many of our conclusions differ. 
For example, while \citet{Kadavath2022LanguageM} suggest that language models are strongest when evaluating their own generations and subsequently posit that uncertainty estimation is linked to self-knowledge, we find that capable models can readily learn good uncertainties for predictions of other models without any knowledge of their internals. 
Lastly, while many works motivate their approach with applications to human-AI collaboration, none of them test their uncertainty estimates on actual users, as we do here. 

\section{Preliminaries}
\label{sec:background}

\paragraph{Question answering evaluations.} In all experiments, we use greedy decoding to generate answers conditioned on questions with few-shot prompts. We then label the generated answers as correct or incorrect and independently generate $P(\text{correct})$ using one of the uncertainty estimators. For evaluation, we primarily use the popular MMLU dataset \citep{Hendrycks2020MeasuringMM}, which covers 57 subjects including STEM, humanities, and social sciences. Crucially, however, we expand the original multiple choice (MC) setting with a new open-ended (OE) setting. In the open-ended setting, we do not provide answer choices, and the language model must generate an answer that matches the ground truth answer choice. We determine a correct match by grading with a strong auxiliary language model (\Cref{app:grading-prompt}). 
We verify that grading via language models provides a cheap and effective proxy for the gold standard human grading (\Cref{app:oe-grading}), consistent with related findings \citep{Chiang2023CanLL}.

\textbf{Metrics.} \hspace{2mm} A model that assigns percentage $p$ to an answer is well-calibrated if its answer is correct $p$ percent of the time it assigns that confidence. Calibration is typically measured using expected calibration error (ECE) \citep{Naeini2015ObtainingWC}, which compares empirical frequences with estimated probabilities through binning (\Cref{app:ece}). A lower ECE is better, and an ECE of $0$ corresponds to a perfectly calibrated model. In addition to calibration, we measure the area under the receiver operating characteristic curve (AUROC) of the model's confidence. High AUROC indicates ability to filter answers likely to be correct from answers that are likely to be incorrect, a setting typically called \textit{selective prediction}. 

\textbf{Temperature scaling.} \hspace{2mm} Temperature scaling \citep{platt1999probabilistic, Guo2017OnCO} improves the calibration of a classifier by scaling its logits by $\frac{1}{T}$ (where $T$ is the temperature) before applying the softmax function. 
A high temperature scales the softmax probabilities towards a uniform distribution, while a low temperature collapses the distribution around the most probable output. The temperature parameter is learned on held-out data, typically taken from the same distribution as the training set. 

\section{Do We Get Good Uncertainties Out-of-the-Box?}
\label{sec:out-of-the-box}

In this section, we focus on black-box\footnote{Here we consider access to a model's samples and token-level likelihoods as black-box. Some models do not expose likelihoods directly, but they can be approximated through sampling.} methods for estimating a language model's uncertainty. Due to computational cost, we focus on methods that require a single sample or forward pass and only consider sampling-based methods in the next section. 

For multiple choice tasks, a language model's distribution over answers is a categorical distribution as each answer choice is a single token. Early work on LLMs, such as GPT-3, showed that this distribution is often poorly calibrated \citep{Hendrycks2020MeasuringMM}. Fundamentally, however, maximum likelihood training should encourage calibration over individual tokens \citep{gneiting2007strictly}, and the calibration of recent LLMs appears to improve in proportion with their accuracy \citep{plaut2024softmax}. 

In open-ended generation, on the other hand, answers are not limited to individual tokens nor a prescribed set of possibilities, which introduces multiple sources of uncertainty. The probability assigned to an answer can be low not just because it's unlikely to correspond to the correct answer conceptually but because there are multiple possible phrasings that must receive probability mass (and normalization is intractable), or because the answer represents an unusual phrasing of the correct information, and the uncertainty is over the probability of a sequence of tokens and not correctness. For example, imagine a multiple-choice test in which we add an additional answer choice that is a synonym of another. A sensible language model would assign equal likelihood to each choice, lowering the probability it assigns to either individually. In open-ended generation the situation is similar, but even more challenging because of variable length. 
Adding extra tokens can artificially lower the likelihood of an answer even when it expresses the same concept, as the sequence of tokens becomes less likely with increasing length.

\begin{figure*}[!t]
    \centering
    \includegraphics[width=0.48\linewidth]{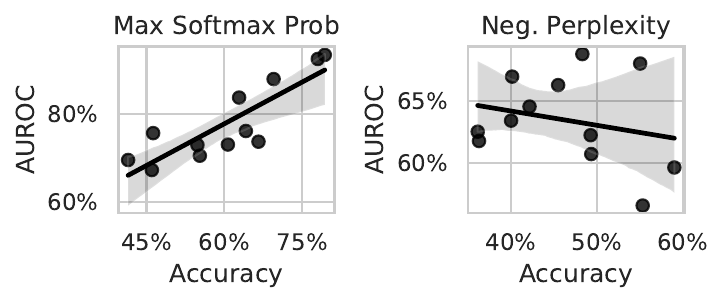}
    \includegraphics[width=0.48\linewidth]{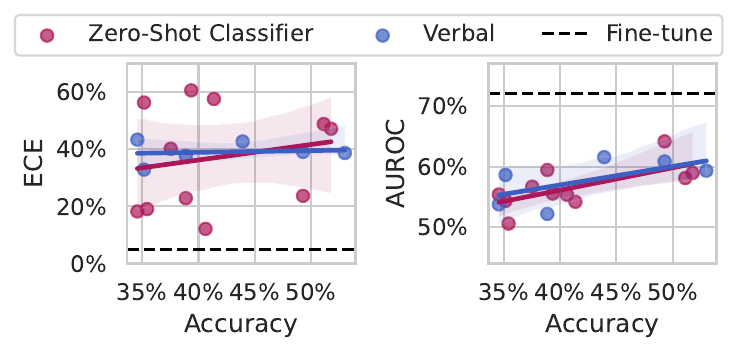}
    \caption{(\textbf{Left}) We compare common uncertainty estimates for multiple-choice questions (max softmax probability) and open-ended generation (perplexity). While maximum softmax probability performs well and improves with the ability of the base model, perplexity does not follow the same pattern. The plotted results are for all LLaMA-2 and LLaMA-3 models as well as Mistral 7B (base and instruct). (\textbf{Right}) Prompting methods for eliciting uncertainty from language models perform poorly when compared to our worst fine-tuned model (LLaMA-2 7B), shown with a dotted line. ECE doesn't appear to improve with the abilities of the underlying model, and while AUROC does show small improvements with large improvements in accuracy, the gap between zero-shot methods and fine-tuning for uncertainties remains large. Shading indicates a 95\% bootstrapped confidence interval on the regression fit.}
    \label{fig:scaling-plots}
\end{figure*}

We demonstrate the difference between multiple-choice question answering and open-ended generation in \Cref{fig:scaling-plots} (left), where we compare the AUROC of a likelihood-based method for standard MMLU and open-ended MMLU (ours). For open-ended generations, we use perplexity, $\text{PPL}(s) = \exp\left(\frac{1}{N} \sum_{i=1}^N \log p(s_i \mid s_{<i}) \right)$, where $s$ is the tokenized sequence, because it is a length-normalized metric and commonly used when token-level probabilities are exposed by the model \citep{Hills_Anadkat_2023}. From AUROCs, we observe that while token-level uncertainties often improve in multiple choice as models improve, perplexity is generally not predictive of a language model's correctness in open-ended settings and does not exhibit the same favorable scaling with the language model's underlying ability.

Because sequence likelihood (or perplexity) is limited as a confidence measure, prompting methods have becoming an increasingly popular alternative. \citet{Lin2022TeachingMT} introduced the following formats that lay the foundation for recent work \citep{tian2023just,zhang2023r}:
\begin{center}
\begin{tabular}{|c|c|c|}
\toprule
\textbf{Name} & \textbf{Format} & \textbf{Confidence} \\
\hline
\thead{Zero-Shot\\Classifier} & ``Question. Answer. {\color{Blue} \bf True/False}:  {\color{Maroon} \bf True}'' & 
\thead{
P({\color{Maroon} \bf ``\,True''}) / \\(P({\color{Maroon} \bf ``\,True''}) + P({\color{Maroon} \bf ``\,False''}))
} \\
\hline
Verbalized & ``Question. Answer. {\color{Blue} \bf Confidence}: {\color{Maroon} \bf 90\%}'' & float({\color{Maroon} \bf ``\,90\%''}) \\
\bottomrule
\end{tabular}
\end{center}
In the first approach, the language model's logits are used to create a binary classifier by scoring two possible strings denoting true and false. 
Similarly, in \citet{Kadavath2022LanguageM}, the classifier takes in a slightly modified prompt, {\color{Blue} \bf ``Is the answer correct? (a) Yes (b) No ''} and confidence is then computed P({\color{Maroon} \bf ``(a)''}) / (P({\color{Maroon} \bf ``(a)''}) + P({\color{Maroon} \bf ``(b)''})). In the second approach (also used in \citep{tian2023just,Xiong2023CanLE}), uncertainty estimates are sampled as text and then converted into numbers. We provide the extended details in \Cref{app:ve-baseline}. 

\textbf{The prospects of calibration by learning to model human language.} If we view language modeling as behavior cloning \citep{schaal1996learning} on human writing, the optimal outcome is a language model that recapitulates the full distribution of human writers present in the training data. Unfortunately, most humans exhibit poor calibration on tasks they are unfamiliar with \citep{kruger1999unskilled, kruger2002unskilled, lichtenstein1977calibration}, and not all pre-training data is generated by experts. Therefore it might be unreasonably optimistic to expect black-box methods to yield calibrated uncertainties without a significant intervention. Alignment procedures (e.g. RLHF) could improve the situation by penalizing cases of poor calibration, and the resulting procedure would be akin to fine-tuning on graded data, which we explore in \Cref{sec:ut}. 

\textbf{Experiments with open-source models.} We examine the quality of black-box uncertainty estimates produced by open source models plotted against accuracy in \Cref{fig:scaling-plots} (right). We use LLaMA-2 \citep{Touvron2023LLaMAOA,Touvron2023Llama2O}, Mistral \citep{Jiang2023Mistral7}, and LLaMA-3 models, and we evaluate on \textit{open-ended} MMLU to highlight how the methods might perform in a ``chat-bot'' setting. Because these models have open weights, we can perform apples-to-apples comparisons with methods that train through the model or access hidden representations. We see that prompting methods typically give poorly calibrated uncertainties (measured by ECE) and their calibration does not improve out-of-the-box as the base model improves. By contrast, AUROC does improve slightly with the power of the underlying model, but even the best model still lags far behind the worse model with fine-tuning for uncertainty. 

\begin{mybox}
    Black-box methods such as perplexity or engineered prompts have limited predictive power and scale slowly, or not at all, with the power of the base model. 
\end{mybox}

\section{How Should We Use Labeled Examples?}
\label{sec:ut}

Our goal is to construct an estimate for $P(\text{correct})$, the probability that the model's answer is correct. Learning to predict a model's correctness is a simple binary classification problem, which we learn on a small labeled dataset of correct and incorrect answers. There are many possible ways to parameterize $P(\text{correct})$, and we study three that vary in their number of trainable parameters and their use of prompting:

\begin{itemize}[topsep=0pt,itemsep=3pt, leftmargin=15pt]
    \item \textbf{Probe}: Following \citet{Azaria2023TheIS}, we train a small feed-forward neural network on the last layer features of a LLM that was given the prompt, question, and proposed answer as input. The model outputs $P(\text{correct})$ while keeping the base LLM frozen. 
    \item \textbf{LoRA}: This parameterization is the same as Probe but with low-rank adapters (LoRA) added to the base model. As a result, the intermediate language features of the base model can be changed to improve the correctness prediction.
    \item \textbf{LoRA + Prompt}: Following \citet{Kadavath2022LanguageM}, we pose classifying correctness as a multiple choice response with two values, the target tokens ``\texttt{i}'' and ``\texttt{ii}'' representing `no' and `yes' respectively. We perform LoRA fine-tuning on strings with this formatting.
\end{itemize}

With these different parameterizations, we can study how much information about uncertainty is already contained in a pre-trained model's features. \texttt{Probe} relies on frozen features, while \texttt{LoRA} and \texttt{LoRA + Prompt} can adjust the model's features for the purpose of uncertainty quantification. Comparing \texttt{LoRA} with \texttt{LoRA + Prompt} also allows us to study how much a language framing of the classification problem aids performance. 

\textbf{Datasets.} \hspace{2mm} For training, we build a diverse set of samples from a collection of benchmark datasets, similar to instruction-tuning \citep{Wei2021FinetunedLM}.
From the list of 16 benchmark datasets in \Cref{app:training-data}, we use a sampled subset of size approximately 20,000. We hold out 2000 data-points to use as a temperature scaling calibration set \citep{Guo2017OnCO}. 

\begin{wraptable}{r}{0.3\linewidth}
    \vspace{-3mm}
    \centering
    \begin{tabular}{c|c|c}
    \toprule
    Method & ECE & AUROC \\
    \hline
    w/o KL & 29.9\% & 70.2\% \\ 
    w/ KL & 10.8\% & 71.6\% \\ 
    \bottomrule
    \end{tabular}
    \caption{Regularization improves calibration. 
    Numbers show the mean over six base models models. See \Cref{app:regularization} for discussion.}
    \vspace{-5mm}
    \label{tab:reg_ablation}
\end{wraptable}

\paragraph{Training and regularization.}
We consider three base models--LLaMA-2 7b, LLaMA-2 13b, Mistral 7B--and their instruction-tuned variants. For fine-tuning, we use 8-bit quantization and Low-Rank Adapters (LoRA) \citep{Hu2021LoRALA}. 
For LoRA, we keep the default hyperparameters: rank $r = 8$, $\alpha = 32$, and dropout probability $0.1$. Each training run takes approximately 1-3 GPU days with 4 NVIDIA RTX8000 (48GB) GPUs. To keep \texttt{LoRA} and \texttt{LoRA + Prompt} in the neighborhood of the initial model, we introduce a regularization term to encourage low divergence between the prediction of the fine-tuned model and the base model (ablation in \Cref{tab:reg_ablation}).

\textbf{Sampling baseline.} \hspace{2mm} We estimate the uncertainty by clustering generations by semantic similarity \citep{Kuhn2023SemanticUL}. The probability of each cluster becomes the probability assigned to all sequences in that cluster. To assign an uncertainty to a prediction, we find the cluster closest to the prediction and use the probability of the cluster as our uncertainty estimate (full details in \Cref{app:sampling-baselines}). The clear drawback of this approach to uncertainty estimation is its poor scaling. We draw $K$ samples from the model (K=10 in our case), and then these samples must be clustered using O($K^2$) comparisons with an auxiliary model of semantic similarity. Sampling methods are also complicated by their relationship with hyperparameters such as temperature or nucleus size. In the special case where the sampling parameters are chosen to produce greedy decoding (e.g. temperature zero), the model will always assign probably one to its answer.  While this behavior does align with the probability of generating the answer, it is not a useful measure of confidence. 

\begin{figure*}[!t]
    \centering
    \includegraphics[width=0.5\linewidth]{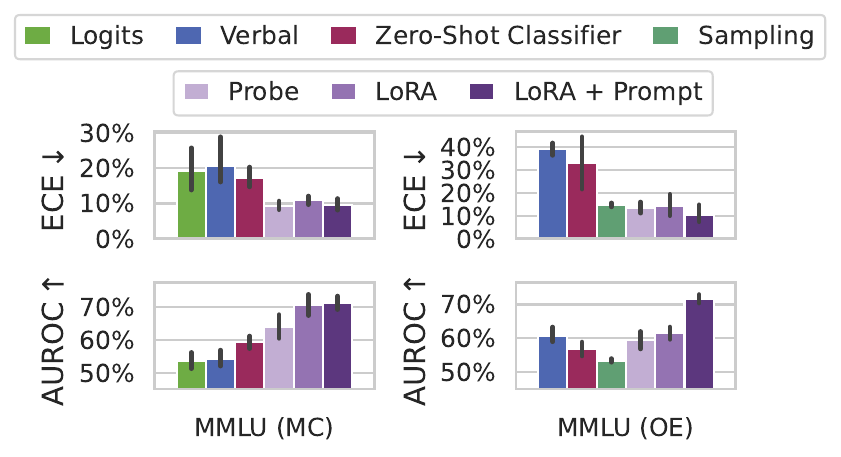}
    \includegraphics[width=0.45\linewidth]{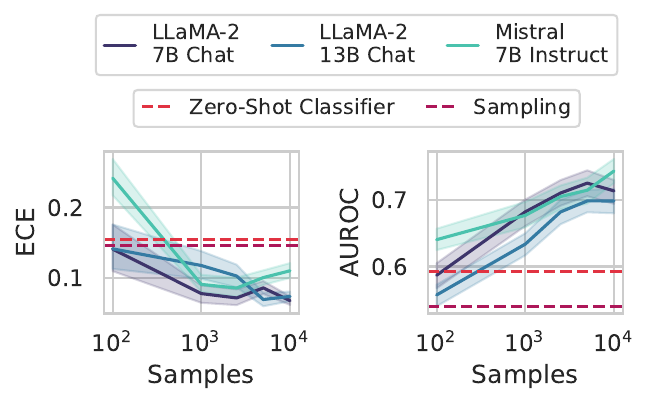}
    \caption{ (\textbf{Left}) ECE and AUROC on both multiple choice (MC) and open-ended (OE) MMLU. ECE is shown after temperature scaling on a small hold-out set. Supervised training (\texttt{Probe}, \texttt{LoRA}, \texttt{LoRA + Prompt}) tends to improve calibration and selective prediction. Probing on its own (\texttt{Probe}) performs worse than training through the features with a language prompt (\texttt{LoRA + Prompt}), especially in an open-ended setting.  Error bars show two standard deviations over six base models. Extended results in \Cref{sec:mmlu_task_breakdown}. (\textbf{Right}) Effect of varying number of labeled datapoints on OE MMLU. In the most extreme case, we train on only 200 examples. Overall, performance increases in proportion with the available labeled data, but 1000 points is almost as valuable as 20,000 points. Dotted lines indicate the performance of the classifier and sampling baselines averaged over the three models considered. Shaded regions show one standard deviation over subsets of MMLU.}
    \label{fig:eval-mmlu}
\end{figure*}

\textbf{Fine-tuning results.} \hspace{2mm} In \Cref{fig:eval-mmlu} (Left) we compare our three fine-tuned models with black-box uncertainty methods on both multiple choice and open-ended MMLU. For multiple choice MMLU, we also include the language model's max softmax probability as a baseline. Fine-tuning for uncertainty leads to significant improvements in both ECE and AUROC. While frozen features (\texttt{Probe}) are sufficient to outperform baselines in multiple choice MMLU, performing well on open-ended MMLU requires training through the modeling and prompting. Surprisingly, while sampling methods can yield good calibration, their discriminative performance is very weak. By contrast, verbal elicitation is relatively strong in discriminative performance, being on par with weaker fine-tuning methods, but general has poor calibration, even after temperature scaling.

\textbf{How much data do we need?} \hspace{2mm} In practice, labels can be expensive to generate, especially on problems where domain expertise is rare. Therefore, it would be advantageous if fine-tuning with even a small number of examples is sufficient for building a good uncertainty estimate. In \Cref{fig:eval-mmlu} (right), we show how calibration tuning is affected by decreasing the size of the fine-tuning dataset. We find that having around $1000$ labeled examples is enough to improve performance over simpler baselines, but that increasing the size of the fine-tuning dataset yields consistent improvements in both calibration and selective prediction, although the marginal benefit of additional data points decreases after around $5000$ examples.

\begin{mybox}
    Supervised learning approaches, in which we learn to predict a model's correctness, can dramatically outperform baselines with as few as $1000$ graded examples. Updating the features of the model with LoRA and use of a language prompt are key to good performance.
\end{mybox}

\section{When and Why Do These Estimates Generalize?}
\label{sec:how-generalize}

\begin{figure*}[!t]
    \centering
    \includegraphics[width=0.34\linewidth]{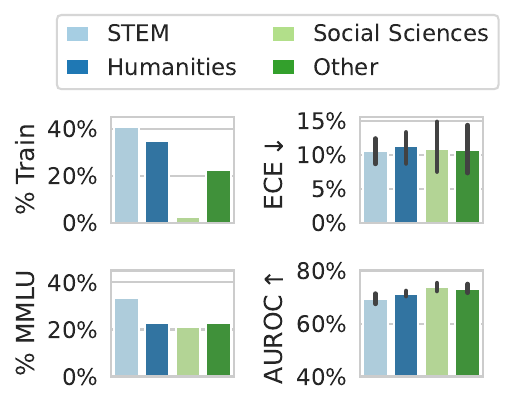} 
    \hspace{4mm}
    \includegraphics[width=0.31\linewidth]{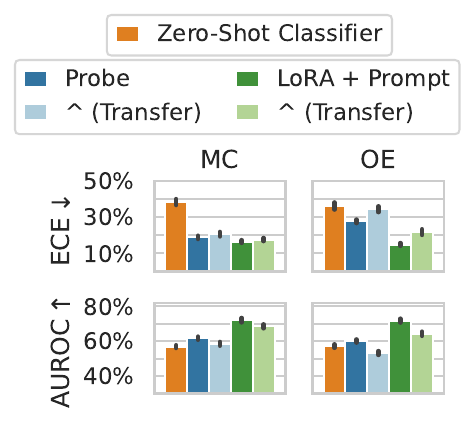} 
    \hspace{6mm}
    \includegraphics[width=0.22\linewidth]{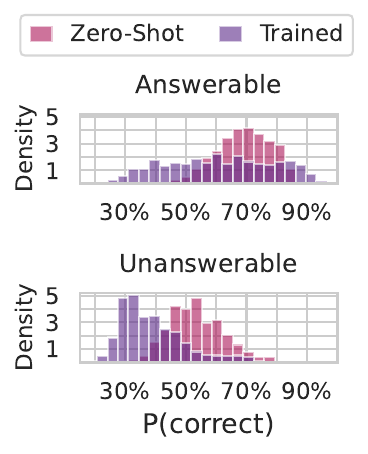} 
    \caption{(\textbf{Left}) We compare the composition of the fine-tuning dataset with MMLU. Notably, although the training dataset contains close to zero examples from social sciences, uncertainty estimates from the model perform similarly across categories. (\textbf{Center}) Testing the generalization of supervised methods by taking models trained on one setting (MCQA or OE) and evaluating them on the other setting. The MCQA or OE labels denote the evaluation setting, with the method labels indicate whether the model was trained on the same or different setting. Fine-tuning through the model's features (\texttt{LoRA + Prompt}) performs almost as well in transfer as on in-distribution data. \texttt{Zero-Shot Classifier} involves no supervised learning except a temperature-scale step and is a useful reference point. Error bars show two standard deviations over six fine-tuned models. (\textbf{Right}) Fine-tuning leads to lower confidence on unanswerable questions, taken from the SelfAware dataset \citep{yin-etal-2023-large}. Assigning low confidence to unanswerable questions allows the model to opt out of responding.}
    \label{fig:mmlu_transfer}
\end{figure*}

To derive more understanding of when our estimates generalize, we now investigate distribution shifts between the training and evaluation datasets. To have a  practically useful tool, we might desire robustness to the following shifts, among others: 

\textbf{Subject matter.} \hspace{2mm} Ideally, our uncertainty estimates apply to subjects we have not seen during training. In \Cref{fig:mmlu_transfer} (left), we show a breakdown of our fine-tuning dataset using the supercategories from MMLU (\Cref{app:mmlu-supercategories}). We see that our dataset contains much higher percentages of STEM and humanities questions than MMLU and close to no examples from the social sciences (e.g. government, economics, sociology). Despite these differences in composition, uncertainty estimates from \texttt{LoRA + Prompt} perform similarly across supercategories. We also show the efficacy of our models at assessing confidence on out of distribution \textit{coding tasks} in \Cref{sec:generalization-coding}.

\textbf{Format.} \hspace{0.5mm} Like a change in subject matter, the way a question is posed should not break the uncertainty estimate. To test the effect of the question format independent of its subject matter, we apply models fine-tuned on OE MMLU to MC MMLU and vice versa. In \Cref{fig:mmlu_transfer} (center), we see that fine-tuned models often perform better than a zero-shot baseline even when they are being applied across a distribution shift, though transfer from MC to OE is more challenging than OE to MC. \texttt{Probe} is insufficient to generalize effectively from MC to OE, but training through the features of the model (\texttt{LoRA + Prompt}) does generalize effectively, even out-performing probe trained on OE data.  

\textbf{Solvability.} \hspace{1mm} Even though we focus on questions with a single known answer, we might hope that our estimates can be used even when a question is ill-posed or does not have a known solution, ideally returning high uncertainty. We generate answers, labels, and uncertainty estimates for the answerable and unanswerable questions in the SelfAware dataset \citep{yin-etal-2023-large} using the same procedure as OE MMLU. In \Cref{fig:mmlu_transfer} (right), we plot $P(\text{correct})$ from \texttt{Zero-Shot Classifier} and \texttt{LoRA + Prompt} predicted for each  answerable and unanswerable question. 
Notably, calibration-tuned models have calibrated probabilities for the answerable questions and assign lower confidence to unanswerable questions than black-box methods. 

\subsection{What are uncertainty estimates learning?}
Language models can generate useful uncertainty estimates after training on a relatively small number of labeled examples. How is this possible? We hypothesize two, potentially complementary mechanisms: (a) LLMs assess the correctness of an answer given a question, or (b) LLMs recognize that certain topics often have incorrect answers. To understand the difference, let's explore a useful metaphor. Imagine I speak only English, while my friend, Alice, is a linguaphile and dabbles in many languages. I have a spreadsheet of how often Alice makes mistakes in each language. Now, when I hear Alice attempting to converse in language A, I can guess how likely she is to err by recognizing the language from its sound and consulting the spreadsheet. I can do this without understanding the language at all. Alternatively, I can learn each language, which would be more complex but would strengthen my predictions.

\begin{figure*}[!t]
    \centering
    \includegraphics[width=0.23\linewidth]{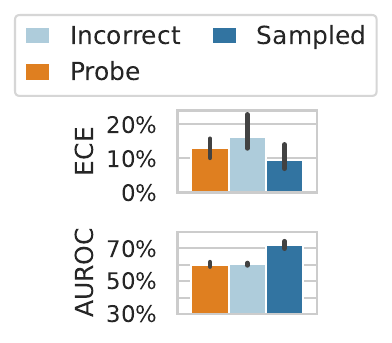}
    \hspace{2mm}
    \includegraphics[width=0.45\linewidth]{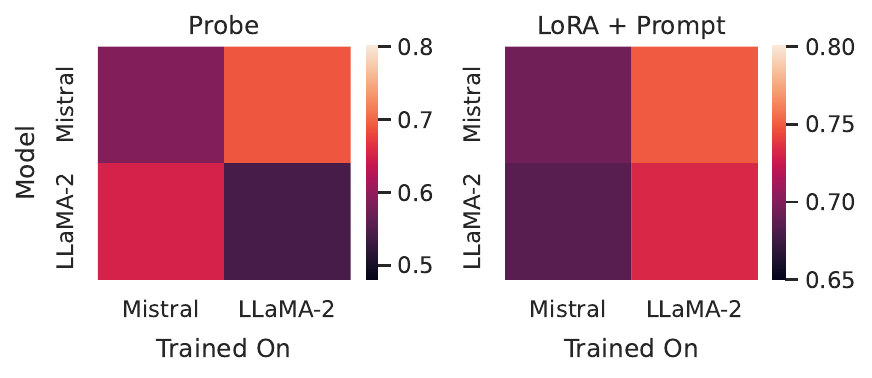}
    \hspace{2mm}
    \includegraphics[width=0.25\linewidth]{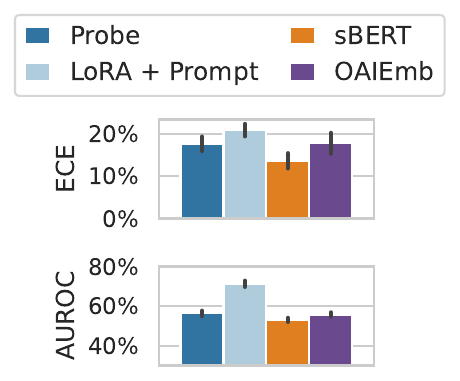}
    \caption{(\textbf{Left}) We ablate the correspondence between questions and answers by training \texttt{LoRA + Prompt} on a dataset with correctness labels from the model's generations but with the actual generations swapped with incorrect answers. In this case, the only relationships that can be extracted by the model are between the correctness labels and the questions. The model trained on incorrect answers generalizes surprisingly well but is much worse than a model trained on the original answers. Error bars show two standard deviations over three instruction-tuned models. (\textbf{Center}) We test how well models can learn to predict the correctness of a different model (in terms of AUROC), and we find that mistral models are often better at estimating the correctness of LLaMA models than LLaMA can on their own generations. (\textbf{Right}) We show that generic sentence embeddings can also perform on par with frozen language model representations (MMLU-OE), but training through a model is much better. \texttt{sBERT} and \texttt{OAIEmb} refer to training a classifier on top of sBERT \citep{reimers2019sentence} or OpenAI sentence embeddings. Error bars show two standard deviations over tasks in MMLU.}
    \label{fig:model_transfer}
\end{figure*}

To disentangle these two possibilities in our setting, we perform an additional experiment, in which we replace the language model's answers in the fine-tuning dataset with incorrect answer options. If a language model is simply learning patterns in the errors present in the training data, then we would expect this ablation to perform on par with the original method because it suffices to learn patterns in the content of the question and answer without needing the true causal relationship between question, answer, and correctness label. The results are shown in \Cref{fig:model_transfer} (left). We see the model trained on incorrect answers performs surprisingly well, on par with a \texttt{Probe} model, but significantly worse than a model trained on the original sampled answers. Correlating question content with error rates while moderately successful cannot be a full description of the \texttt{LoRA + Prompt} estimates. 

\textbf{Self-knowledge.} \hspace{2mm} Lastly, we examine whether a language model can be used to model not just its own uncertainties but the uncertainties of other models. Several prior works argue that models identify correct questions by way of internal representations of truth, which might be unique to a model evaluating its own generations \citep{Azaria2023TheIS, Burns2022DiscoveringLK}. In \Cref{fig:model_transfer} (right), we show that, by contrast, Mistral 7B actual has better AUROC values when applied to LLaMA-2 7B than LLaMA-2 7B applied to itself. In \Cref{fig:model_transfer} (left), we show that sBERT \citep{reimers2019sentence} and OpenAI sentence embeddings are competitive with \texttt{Probe} on both LLaMA-2 7B and Mistral. Together, these results suggest that LLM uncertainties are likely not model-specific. The practical upside of this insight is that one strong base model can be used to estimate the uncertainties of many other models, even closed-source models behind APIs, when a small labeled dataset is available or can be generated. 

\begin{mybox}
Learned uncertainty estimates generalize to new formatting, subject matter, and even the generations of other models. This generalization appears to stem not simply from judging a question's difficulty based on its subject matter (a short-cut) but also learning the correspondence between questions and correct answers. 
\end{mybox}

\section{Does Calibrated Confidence Improve Collaboration with AI Assistants?}
\label{sec:user-study}

One key motivation for estimating LLM uncertainty is to signal the model's reliability during collaborative decision making. To examine how our uncertainty estimates can be used in this capacity, we perform a preliminary user study (with $N=181$ participants) in which participants complete a multiple choice exam in collaboration with an LLM (Mistral 7B Instruct). For each question, the participant is provided both the LLM's prediction and an uncertainty estimate, which can be from a calibrated method or an uncalibrated method. We hope to show that users are more likely to adopt calibrated uncertainty scores as part of their decision process. A more detailed description of the setup of our study is available in \Cref{app:user-study}. 

\begin{figure*}[!t]
        \centering
\begin{tabular}{ccc}
     \includegraphics[width=.3\linewidth]{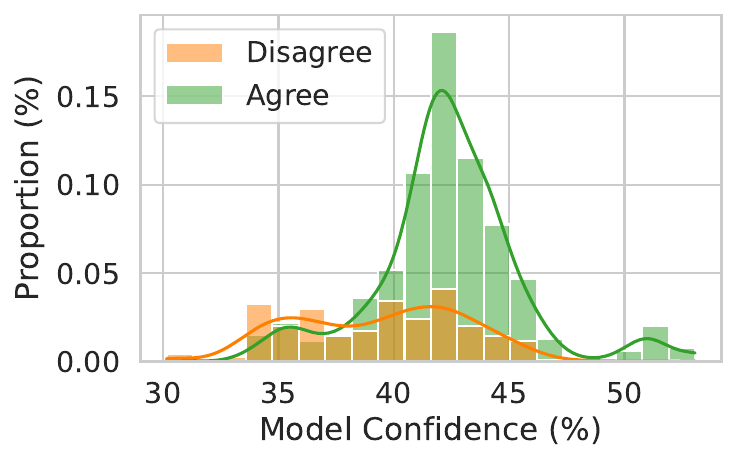}
     & \includegraphics[width=.3\linewidth]{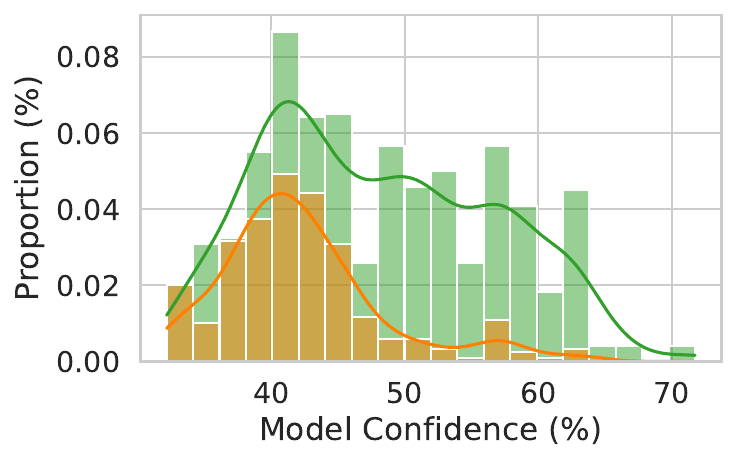} & \includegraphics[width=.3\linewidth]{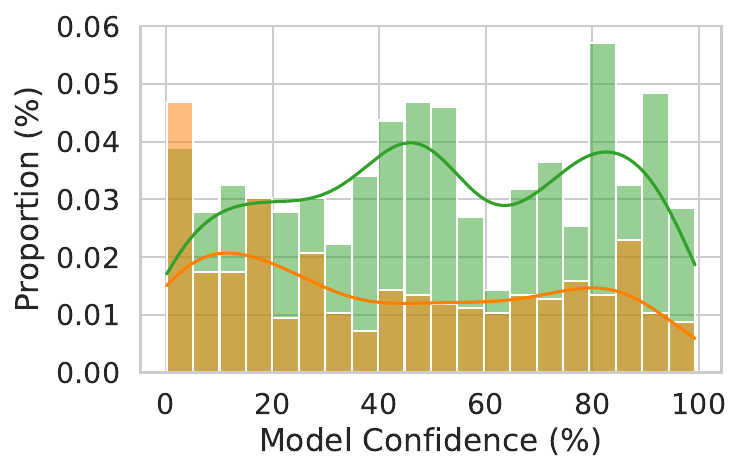} \\
     (a) Zero-Shot Prompt & (b) LoRA + Prompt & (c) Random (Control) 
\end{tabular}
        \caption{We compare the distribution of LLM confidence (for Mistral 7B Instruct) on its answers, and whether the users ($N = 20$ per variant) agree with the answer generated by the model or not. \textbf{(a)} For the zero-shot prompt, we find that the model provides little signal since most mass is similarly clustered. However, \textbf{(b)} improving the calibration of the model reveals an increased reliance on the LLM for more confident answers, and decreased reliance for less confident answers. Evidently, the users are sensitive to calibrated confidence scores. \textbf{(c)} For reference, we verify that uniformly confidence scores do not provide meaningful signal, rendering users unable to modulate their decision to rely on the LLM. All variants are compared at approximately the same average participant accuracy.
        }
        \label{fig:user-study-reliance}
\end{figure*}

\paragraph{People are sensitive to informed confidence scores.} Figure \ref{fig:user-study-reliance} shows density plots of the model's reported confidence and whether the user chose to agree with the model's prediction. We find that participants are sensitive to the confidence scores and tend to use scores when deciding to agree or disagree with the model's prediction if the uncertainties are reliable. On the other hand, participants generally do not modulate their decision to rely on the output of a random confidence baseline (\Cref{fig:user-study-reliance}(c)), in which the display uncertainty estimate is generated uniformly at random. 
We see the strongest discrepancy in reliance choices when \texttt{LoRA + Probe} confidence scores are presented, highlighting that calibrated confidence does influence user behavior.

We include additional details and results in \Cref{app:user-study}. We find that confidence scores have the biggest effect on improving the lowest performing users, rather than on average accuracy. However, this is a preliminary result in the nascent field of studying LLM uncertainties in practical collaborative decision making with users. We are only still scratching the surface of this question.
For more fine-grained conclusions, a study should be devoted to this subject. We outline several limitations and future directions in \Cref{app:user-study}.

\begin{mybox}
Users are sensitive to confidence scores and use their relative magnitude to modulate their decision to use an LLM. Lower performing users are most improved by access to confidence scores. However, future work is needed to disentangle the effects of calibration from how humans choose to leverage uncertainties.
\end{mybox}

\section{Discussion}
\label{sec:discussion}

There is much disagreement about the role of calibrated uncertainty in large language models, how it can best be achieved, and promise of black-box methods. We hope to have shed light on these questions throughout this paper. In contrast to prior results, we find that out-of-the-box uncertainties from LLMs are unreliable for open-ended generation and introduce a suite of fine-tuning procedures that produce calibrated uncertainties with practical generalization properties. In the process, we discovered that fine-tuning is surprisingly sample efficient and does not seem to rely on representations of correctness specific to a model evaluating its own generations, allowing estimators to be applied from one model to another. Moreover, we found it is possible, at least in the cases we considered, for calibrated uncertainties to be robust to distribution shifts. 

There are many exciting questions for future work. Currently fine-tuning relies on two separate models for question answering and uncertainty estimation. Ideally, we want a single model that can generate questions and uncertainty without switching between model weights. We anticipate that an uncertainty-aware pre-training or alignment phase might become essential but implementing such a procedure while maintaining base language modeling abilities will introduce a challenging online learning problem where the correctness labels evolve during training. 

Beyond improving the safety and usefulness of language models, high quality uncertainties can also be used in active learning procedures, e.g. for sample-efficient fine-tuning \citep{osband2022fine}, where data points are selected based on the predicted utility and the model's uncertainty, in order to balance the explore-exploit trade-off. Uncertainty estimates can also be used to improve factuality of language models by increasing the likelihood of generations that the model is confident about (judges likely to be correct), for example by using an alignment procedure (e.g. RLHF, DPO) with a reward function that encourages confident generations \citep{tian2023fine}.

We also showed how uncertainty information could be used to influence human decision making. In the end, LLMs will impact society through decision making, and to make reasonable decisions we need uncertainty information --- particularly to protect against rare but costly mistakes.

\section*{Acknowledgements}
This work is supported by NSF CAREER IIS-2145492,
NSF CDS\&E-MSS 2134216, NSF HDR-2118310, BigHat Biosciences, Capital One, and an Amazon Research Award.

\bibliography{references}
\bibliographystyle{plainnat}

\clearpage

\appendix

\onecolumn

\addcontentsline{toc}{section}{Appendix}
\renewcommand \thepart{}
\renewcommand \partname{}

\vbox{%
    \hsize\textwidth
    \linewidth\hsize
    \vskip 0.1in
    \hrule height 4pt
    \vskip 0.25in
    \vskip -\parskip
    \centering
    {\LARGE\bf Appendix for \\Large Language Models Must Be Taught \\to Know What They Don't Know \par}
    \vskip 0.29in
    \vskip -\parskip
    \hrule height 1pt
    \vskip 0.09in
}

\part{}

\vskip -0.5in
\parttoc

\section{Evaluation Methods}

\subsection{Evaluating Correctness}
\label{sec:grading}

For a given question with known and generated answers $(Q, A, \hat{A})$ the correctness $C$ is True if the generated answer $\hat{A}$ matches the ground truth answer $A$. 
For multiple-choice question-answering, the matching process only involves checking the first token generated via greedy decoding.  

For open-ended evaluations, determining if the answer given is correct is more complex. One simple approach is to check if the ground truth answer $A$ appears as a substring of answer $\hat{A}$. However, this does not capture rephrasings that may be essentially equivalent - such as "NYC" for "New York City," or "Daoism" and "Taoism." Conversely, it also has the potential to be over-generous if the model is particularly verbose and emits many incorrect answers along with the correct string. 
Given the difficulty involved in writing a rule-based method for evaluating open-ended answer correctness, we use instead a strong auxiliary language model to evaluate correctness. The auxiliary language model is shown the query $Q$, the ground truth answer $A$, and the model's output $\hat{A}$, and is prompted to grade the answer whilst tolerating nuance. For full details of the prompt used see (\cref{fig:prompts}). In this paper we utilise GPT 3.5 Turbo as the auxiliary grading model.
We conduct a comparison of human grading, substring grading, and GPT 3.5 Turbo grading on select subsets of MMLU in \cref{app:oe-grading}. We find that humans and GPT 3.5 Turbo have much greater agreement than humans and the substring method.

\subsection{Grading}
\label{app:grading-prompt}

\paragraph{Dataset Construction.}
To perform calibration-tuning ({\sc CT}), we need tuples $(Q, A, \hat{A}, C)$, answers from a language model that have been graded for correctness. When calibration-tuning on multiple choice questions, we can use an exact string match to generate $C$. To grade open-ended answers, we use a strong language model and \emph{grading prompt} $G$ instead (\cref{fig:prompts}):
\begin{itemize}[topsep=0pt]
    \setlength\itemsep{0em}
    \item $\boldsymbol{G}$: a prompt used for grading answers $\boldsymbol{\hat{A}}$ with $\boldsymbol{A}$.
\end{itemize}
Compared to alternatives like exact match, language model grading is insensitive to re-phrasings that are equivalent in meaning - such as ``NYC" and ``New York City," or ``Daoism" and ``Taoism". 
LLM grading can also penalize answers that are overly verbose or use a different meaning of the same word, potentially containing incorrect answers along with the correct string. 
For example, if the question is ``What's it called when you move quickly by foot and both feet aren't always touching the ground?'' and the LLM response is ``A bank run", the grader should be able to distinguish that this is semantically dissimilar to the true answer ``run''. 

In this paper, we utilize GPT 3.5 Turbo as the auxiliary grading model. When comparing many possible grading methods on subsets of MMLU, we find that GPT 3.5 Turbo has high agreement with humans while being cost efficient (\cref{app:oe-grading}).

\begin{figure}[!ht]
    \centering

\begin{adjustbox}{width=0.5\linewidth}
    
    \begin{tabular}{|p{0.5\linewidth}|}
    \toprule

    \textbf{Grading prompt} $(\boldsymbol{G})$\\
    \midrule
    
    {    
    The problem is: $\boldsymbol{Q}$ \newline
    The correct answer is: $\boldsymbol{A}$ \newline
    A student submitted: $\boldsymbol{\hat{A}}$ \newline
    \newline
The student's answer must be correct and specific but not overcomplete (for example, if they provide two different answers, they did not get the question right). However, small differences in formatting should not be penalized (for example, `New York City' is equivalent to `NYC'). Did the student provide an equivalent answer to the ground truth? Please answer yes or no without any explanation: $\boldsymbol{C}$ \texttt{</s>}
    }
    
    \\
    \bottomrule
    \end{tabular}
\end{adjustbox}    
    \caption{For open-ended generation, we calculate the ground-truth correctness $C$ using a LLM and a grading prompt ($G$). The token \texttt{</s>} is an end-of-sentence token. {\color{blue}\textbf{Blue text}} is included in the loss function when calibration-tuning.}
    \label{fig:prompts}
\end{figure}

\subsection{Comparison of Grading Techniques}
\label{app:oe-grading}

We conducted an analysis of the methods outlined in \cref{sec:grading} for open-ended evaluation. First, the base LLaMA-2 13b-chat model was prompted with questions from the following test subsets of MMLU: World Religions, Philosophy, Anatomy, High School Chemistry and Elementary School Math. The questions were stripped of their multiple-choice options before being supplied to the model.

A response was generated by the model via greedy decoding and this response was compared to the ground truth answer. The grading methods tested were Human, Substring Match, GPT 3.5 Turbo, and GPT 4.

The humans (a subset of our authors) were tasked to judge if the model response was essentially equivalent to the ground truth. For substring match, equivalence was determined by simply checking whether the ground truth answer existed as a substring within the model response. For GPT 3.5 Turbo and GPT 4, the models were supplied with the question, the ground truth, and the base model response, as well as a prompt indicating they should determine essential equivalence - see \cref{fig:prompts}.

\begin{table}[ht]
\centering
\begin{adjustbox}{width=.6\linewidth}
\begin{sc}
\begin{tabular}{l|c|c|c}
\toprule
MMLU Subset    & Substring Match & GPT3.5 & GPT4 \\ \midrule
World Religions         & 21.6\%                                  & 6.4\%                          & 1.8\%                        \\
Philosophy              & 22.8\%                                  & 2.3\%                          & 14.5\%                       \\
Anatomy                 & 13.3\%                                  & 14.8\%                         & 1.5\%                        \\
Chemistry               & 13.8\%                                  & 5.4\%                          & 1.0\%                        \\
Math                    & 12.4\%                                  & 14.8\%                         & 3.7\%                        \\
\midrule
\textbf{Average}        & \textbf{16.8\%}                         & \textbf{8.7\%}                 & \textbf{4.5\%}               \\
\bottomrule
\end{tabular}
\end{sc}
\end{adjustbox}
\vspace{2mm}
\caption{Absolute differences in accuracy \% for the different grading methods vs human estimated accuracy. A lower value corresponds to an accuracy estimate closer to the human estimate.}
\label{tab:oe-grading-diffs}
\end{table}

We recorded the binary decision on correctness for each query and response by each of the grading methods above. Taking the human scores as the gold standard of correctness, we computed the model accuracy for each subset, and then derived the absolute error in estimate of model accuracy by each of the other grading methods. These are displayed in \cref{tab:oe-grading-diffs}. We see that GPT4 is a better estimator of human-judged correctness than GPT 3.5 Turbo, which in turn is substantially better than substring match; although there is some variance on a per-subset basis. For expediency of processing time and cost, we chose to use GPT 3.5 Turbo in this paper.

\subsection{Metrics}
\label{app:ece}

\paragraph{ECE} Given $N$ samples and $B$ equally-spaced bins $b_j$, examples are assigned to bins based on the confidence of the model, and ECE is estimated as
$\widehat{\text{ECE}} = \sum_{j=1}^B \frac{\lvert b_j \rvert}{N} \left\lvert \mathrm{conf}(b_j) - \mathrm{acc}(b_j) \right\rvert$
where $\mathrm{conf}(b_j)$ is the average confidence of samples in bin $b_j$, $\mathrm{acc}(b_j)$ is the accuracy within the bin, and $\lvert b_j \rvert$ is the number of samples assigned to bin $j$. In our experiments $\mathrm{conf}$ is equivalent to $P(\text{correct})$.

\subsection{MMLU Supercategory Classifier}
\label{app:mmlu-supercategories}

To understand the impact of the subject matter of the training data on generalization, we follow the prescription of \citet{Hendrycks2020MeasuringMM} and categorize each of the 57 tasks into one of four supercategories - Humanities, STEM, Social Sciences, and Other. Since we do not have such a categorization for the training set, we must estimate their proportions.

First, we use the OpenAI embeddings (dimension 1536) of the MMLU samples with their ground truth supercategories to train a linear 4-way classifier with 10 samples from each of the 57 tasks. We use AdamW \citep{Loshchilov2017FixingWD} with learning rate 1e-3 and weight decay 1e-2. This classifier is then used to estimate the categories of each sample in the training set used for fine-tuning. Subsequently, the breakdown of results in \cref{fig:mmlu_transfer} (Left) follows.

\section{Baseline Methods}

\subsection{Sampling Methods}
\label{app:sampling-baselines}

We use two baselines which obtain an estimate of certainty by sampling the same answers $n=10$ times and then estimating the proportion of sampled answers that agree with the greedily decoded ``main" answer. There are several critical downsides to these approaches: (i) the uncertainty here depends on the sampling parameters---for example, in the limit where the sampling converges to mere greedy decoding, the LLM will produce $n$ identical samples, and therefore the certainty will always be 1---(ii) these approaches require $O(n)$ answer generations to provide a certainty estimate for a single generation. The intense computational restriction prevents us from easily searching the space of sampling parameters for the optimal set, so we choose parameters arbitrarily; here we sample with top$\_p = 0.95$.

\paragraph{Counting} In this baseline, each sampled answer is compared to the greedy answer by prompting an expert LLM with both answers and asking it to judge their equivalence. The proportion of samples that are equivalent to the greedy answer is the certainty estimate. This baseline is similar to \emph{Label prob} \cite{tian2023just}; our method differs by not choosing the argmax semantic group as the final prediction, but instead using the greedy decode for the final prediction, so as to maintain the same accuracy performance as our uncertainty query method. This met

\paragraph{Likelihood accumulation} In this baseline, we add up likelihoods of sampled answers to estimate the mass associated with the predicted answer. We begin by prompting an expert LLM in order to find which sampled answers are equivalent to the greedy answer---like in the counting baseline. In this method, the certainty estimate is produced by adding the length-normalized likelihoods of those sampled answers equivalent to the greedy answer, and dividing this quantity by the sum of all sampled answers' length-normalized likelihoods. This procedure of adding likelihoods of samples in order to estimate the likelihood of an equivalence class is similar to that used by \cite{Kuhn2023SemanticUL}, although they do not use it for certainty estimates but instead to produce entropy scores. In practice, the scores produced by these two methods are actually very similar---so we report only likelihood accumulation numbers in the main text.

\subsection{Verbal Elicitation}
\label{app:ve-baseline}

Although \citet{tian2023just} introduce several strategies for prompting, involving multiple guesses or multiple stages of interleaving prompting and generation, we did not find that any strategy consistently outperformed any other. This finding was consistent with the results of \citet{Xiong2023CanLE}. Ultimately, for convenience, we adopted a two stage strategy with a single guess because it can be used in tandem with logged datasets of generated answers per model.

The exact prompt we used is essentially the same at in \citep{tian2023just}, but with small modifications that improved the rate of correctly formatted responses:

\begin{quote}
``Provide the probability that your answer is correct. Give ONLY the probability, no other words or explanation.

For example:

Probability: <the probability between 0.0 and 1.0 that your guess is correct, without any extra commentary whatsoever; just the probability!>

Include probability for the answer below:
Probability:''
\end{quote}

Verbal elicitation methods typically output complex strings containing both answers and associated probabilities. This means that if any element of parsing fails, it can be challenging to construct partial results. This effect tends to diminish when using large models, which are more responsive to zero-shot prompting.

\paragraph{Parsing Details}
The original verbal elicitation prompts are given in the appendix of  \citep{tian2023just}. However, it is not clear how the original authors decide to parse answers from the generations and how failure to parse is handled. When we fail to parse the guess from the generation we return an empty string and associated probability 0.5. When we fail to parse a probability, we also return probability 0.5. For versions with multiple guesses, if any part of the parsing processes fails in an ambiguous way, we default back to an empty string for the answer and 0.5 for the probability. The only unambiguous cases are those which explicit succeed in the generating a valid guess and probability in the first case but not subsequent cases. In this scenario, we default to using the successfully parse first guess and associated probability.

\section{Fine-tuning Method}

\subsection{Regularization Term}
\label{app:regularization}

To keep the calibration-tuned parameters $\theta$ within the neighborhood of the initial parameters, $\theta_0$, we use a regularization term that penalizes the divergence between the original sampling distribution and the calibration-tuned model on the target sequence $A$, yielding regularization $\mathcal{R}(\theta; \theta_0)$, which we use with weighting parameter $\kappa$.

Specifically, let $p_{\theta_0}$ be the language modeling distribution of the language model we wish to calibration-tune, and $q_\theta$ be the corresponding language modeling distribution as a consequence of calibration-tuning.
We then use the Jensen-Shannon Divergence ${\mathrm{JSD}(p_{\theta_0} \parallel q_\theta)}$ \citep{MacKay2004InformationTI} between the two language modeling distributions as the regularizer, where ${\mathrm{JSD}(p \parallel q) \defeq \nicefrac{1}{2}(\kl(p \parallel m) + \kl(q \parallel m))}$, where $m \defeq \nicefrac{1}{2}(p + q)$ is the mixture distribution.
JSD regularization is applied only to the logits corresponding to the target sequence $A$.

We note that using either direction of KL-divergence, i.e. the forward-KL $\mathrm{KL}(p_{\theta_0} \parallel q_{_\theta})$ or reverse-KL $\mathrm{KL}(q_{_\theta} \parallel p_{\theta_0})$ was insufficient for optimal performance with calibration tuning. 
The forward KL-divergence encourages a zero-avoiding behavior such that the mass of $q_\theta$ is spread across multiple modes of $p_{\theta_0}$ to minimize the KL-divergence to avoid assigning no mass to regions of the probability space.
To the contrary, the reverse KL-divergence encourages a zero-forcing behavior such the $q_\theta$ only needs to cover any one mode of $p_{\theta_0}$ \citep{bishop2006pattern}.
It is not necessarily obvious which one of these behaviors one should prefer for the specific case of large language models. Therefore, as a practical choice, we pick the one that provides us the most performant calibration-tuned model. 

\subsection{Training Data}
\label{app:training-data}

We reserve the following datasets for training.

\begin{itemize}[itemsep=0ex,topsep=0pt]
\item AI2 Reasoning Challenge (ARC) \citep{Clark2018ThinkYH}, 
\item Boolean Questions (BoolQ) \citep{Clark2019BoolQET}, 
\item CommonsenseQA \citep{Talmor2019CommonsenseQAAQ}, 
\item CosmosQA \citep{Huang2019CosmosQM}, 
\item HellaSwag \citep{Zellers2019HellaSwagCA}, 
\item MathQA \citep{amini-etal-2019-mathqa}, 
\item Recognizing Textual Entailment (RTE/SNLI) \citep{Bowman2015ALA},
\item Adversarial NLI \citep{Nie2019AdversarialNA}, 
\item OpenBookQA \citep{Mihaylov2018CanAS}, 
\item PIQA \citep{Bisk2019PIQARA}, 
\item SciQ \citep{Welbl2017CrowdsourcingMC}, 
\item The CommitmentBank (CB) \citep{de2019commitmentbank}, 
\item Multi-Sentence Reading Comprehension (MultiRC) \citep{Khashabi2018LookingBT}, 
\item Choice of Plausible Alternatives (CoPA) \citep{Gordon2011SemEval2012T7}, 
\item TREC \citep{Li2002LearningQC}, 
\item Adversarial Winograd (Winogrande) \citep{Sakaguchi2019WINOGRANDEAA}.
\end{itemize}

\subsection{Training Hyperparameters}
\label{sec:mcqa-hypers}

We use HuggingFace Transformers \citep{wolf-etal-2020-transformers} and PyTorch \citep{Paszke2019PyTorchAI} for the implementation of these models. 
For all our experiments, we use the AdamW optimizer \citep{Loshchilov2017FixingWD} with a learning rate of $10^{-4}$, a cosine decay schedule, and effective batch size $M = 32$. 
The training runs for $G= 10000$ with an initial linear warmup schedule for $1000$ steps. 

\section{Extended MMLU Results}
\label{sec:mmlu_task_breakdown}

\begin{figure*}[ht]
    \centering
    \includegraphics[width=\linewidth]{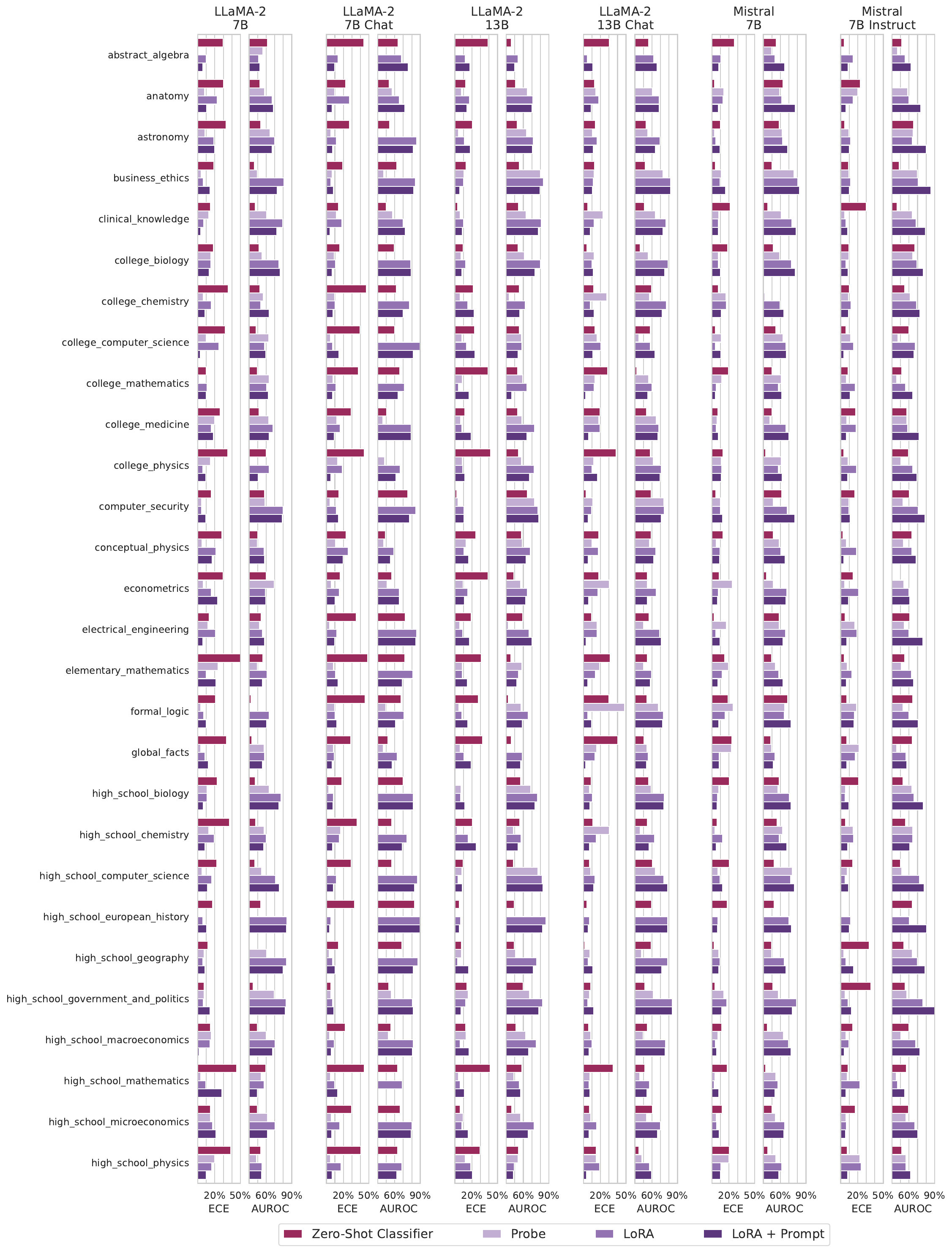}
    \caption{(Part 1) ECE and AUROC values for \texttt{Query}, \texttt{CT-Probe}, \texttt{CT-LoRA}, and \texttt{CT-Query} for each subset of MMLU in multiple-choice question answering (MCQA) setting.}
    \label{fig:mmlu_mcqa_bar1}
\end{figure*}

\begin{figure*}[ht]
    \centering
    \includegraphics[width=\linewidth]{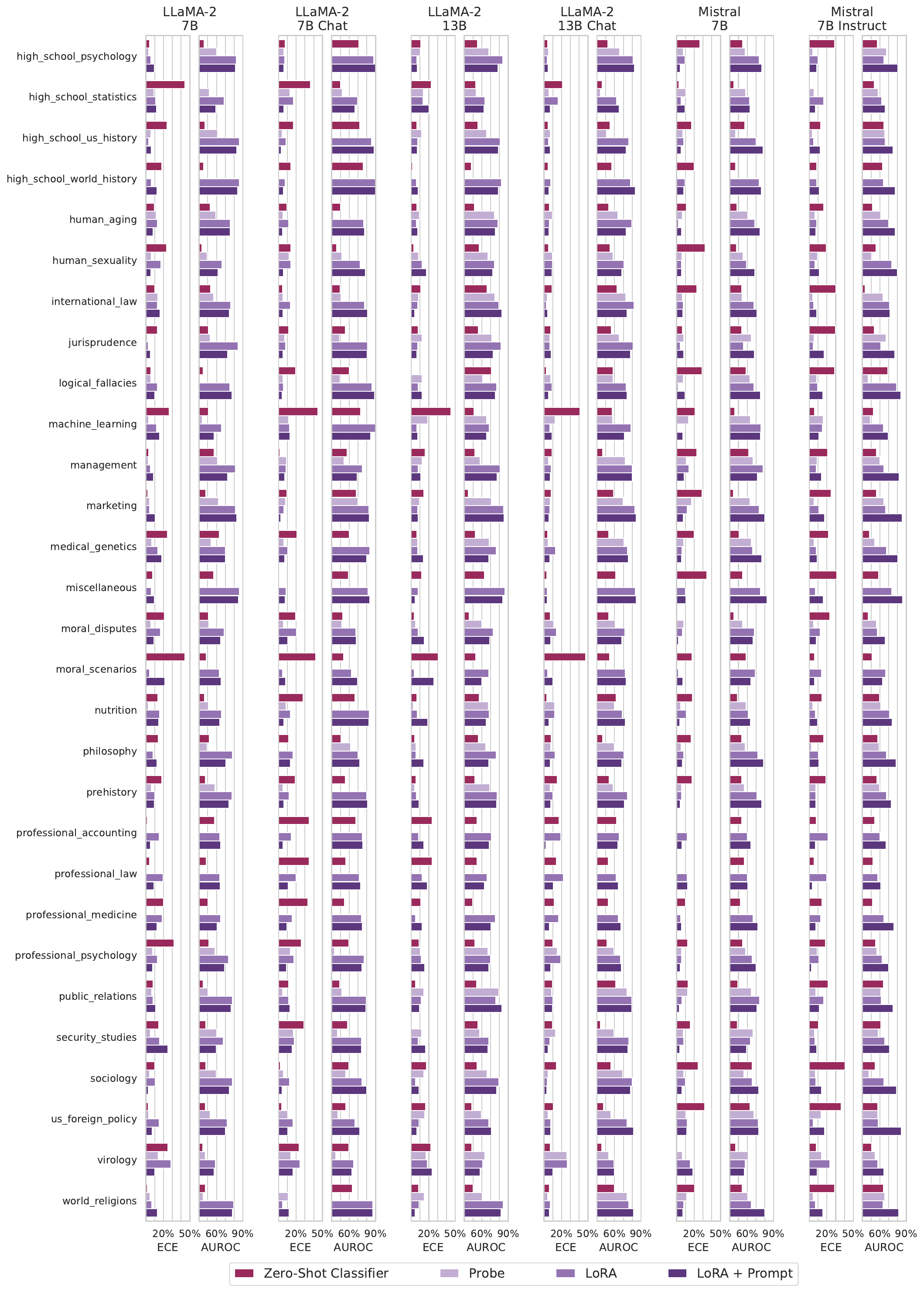}
    \caption{(Part 2) ECE and AUROC values for \texttt{Query}, \texttt{CT-Probe}, \texttt{CT-LoRA}, and \texttt{CT-Query} for each subset of MMLU in multiple-choice question answering (MCQA) setting.}
    \label{fig:mmlu_mcqa_bar2}
\end{figure*}

\begin{figure*}[ht]
    \centering
    \includegraphics[width=\linewidth]{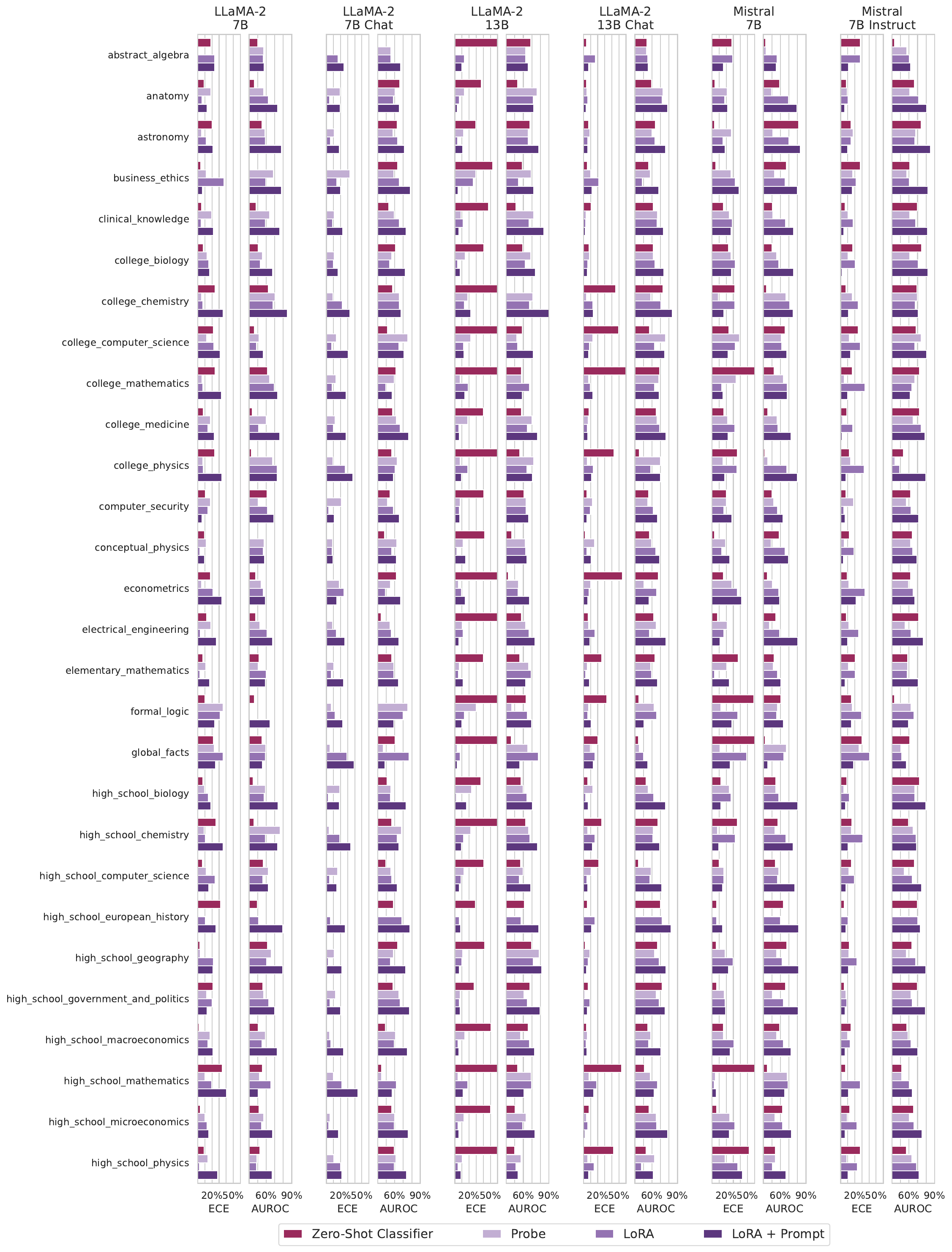}
    \caption{(Part 1) ECE and AUROC values for \texttt{Query}, \texttt{CT-Probe}, \texttt{CT-LoRA}, and \texttt{CT-Query} for each subset of MMLU in open-ended (OE) setting.}
    \label{fig:mmlu_oe_bar1}
\end{figure*}

\begin{figure*}[ht]
    \centering
    \includegraphics[width=\linewidth]{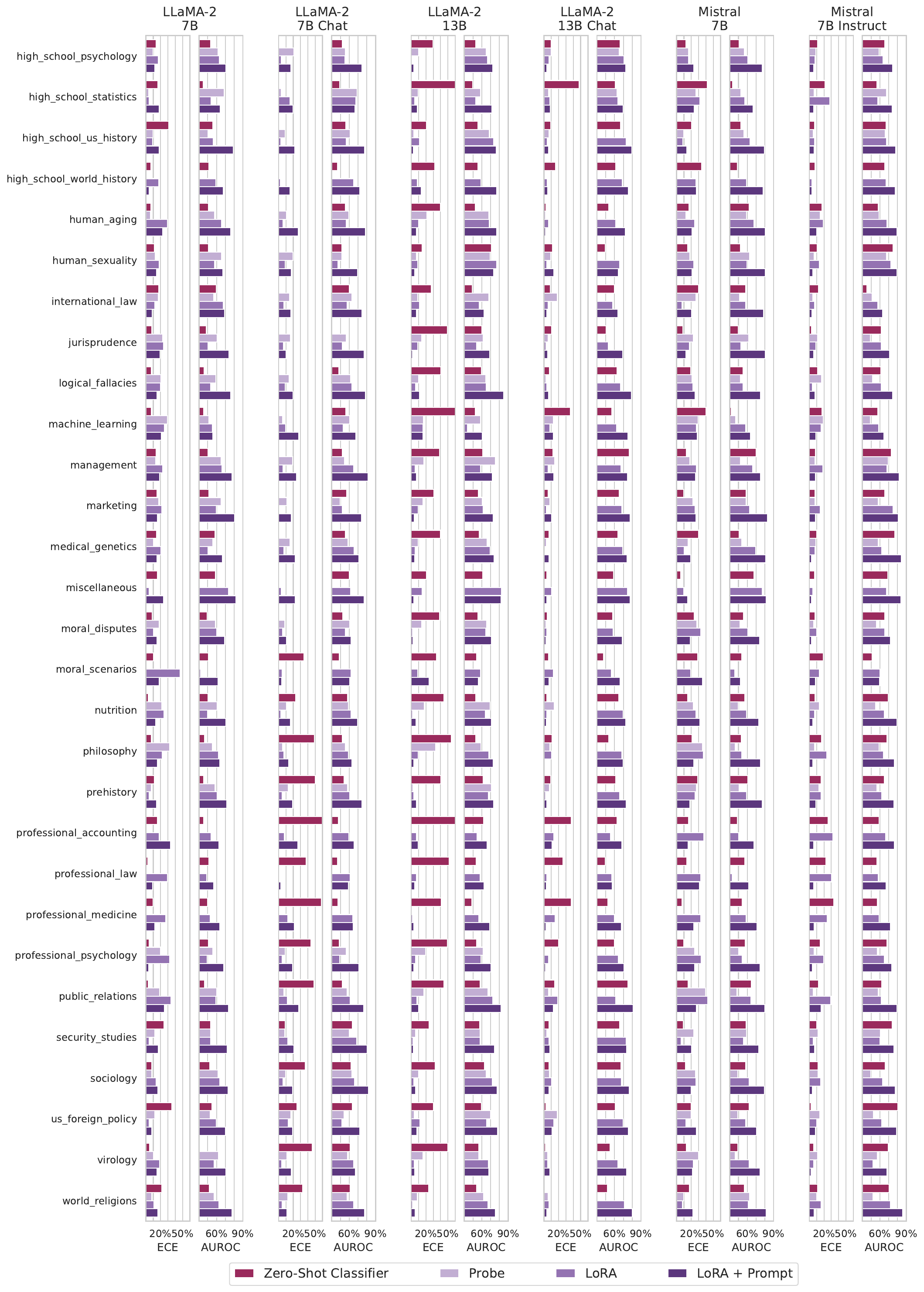}
    \caption{(Part 2) ECE and AUROC values for \texttt{Query}, \texttt{CT-Probe}, \texttt{CT-LoRA}, and \texttt{CT-Query} for each subset of MMLU in open-ended (OE) setting.}
    \label{fig:mmlu_oe_bar2}
\end{figure*}

We report the breakdown of uncertainty query accuracy and ECE on all MMLU tasks in \cref{fig:mmlu_mcqa_bar1,fig:mmlu_mcqa_bar2,fig:mmlu_oe_bar1,fig:mmlu_oe_bar1,fig:mmlu_oe_bar2}.

\section{Confidence as a Function of Target Length}
\label{sec:conf_vs_target_len}

As we noted when motivating calibration tuning, one limitation of sequence-level probabilities is their intrinsic connection to sequence length. The probability of a sequence decreases with increasing length, regardless of the correctness of the response. By contrast, we wouldn't expect concept-level probabilities to have any discernible relationship with sequence length. In \cref{sec:conf_vs_target_len}, we show there is no consistent relationship between the confidence estimated by the calibration-tuned model and target sequence length on MMLU tasks. 

\begin{figure*}[!t]
    \centering
\begin{tabular}{cccc}
   \includegraphics[width=.2\linewidth]{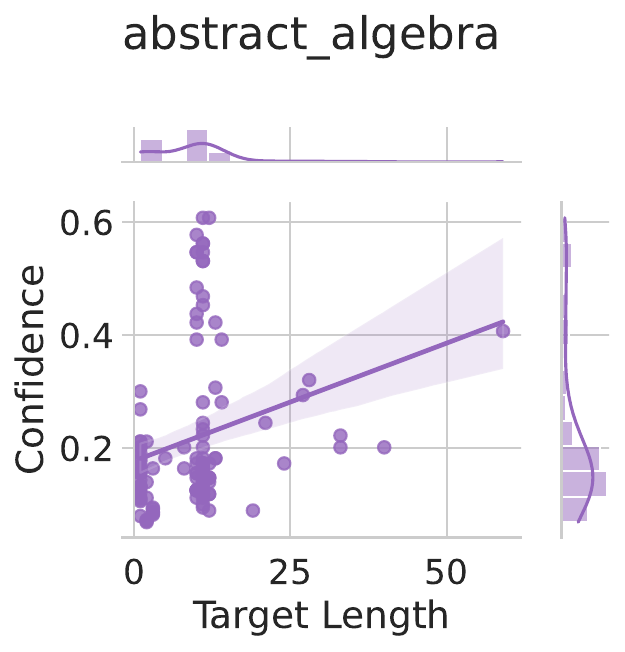} &
   \includegraphics[width=.2\linewidth]{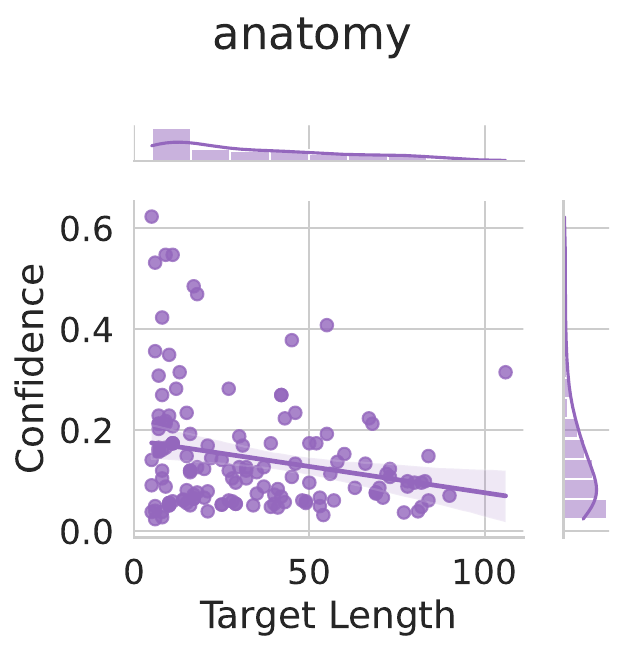} & 
   \includegraphics[width=.2\linewidth]{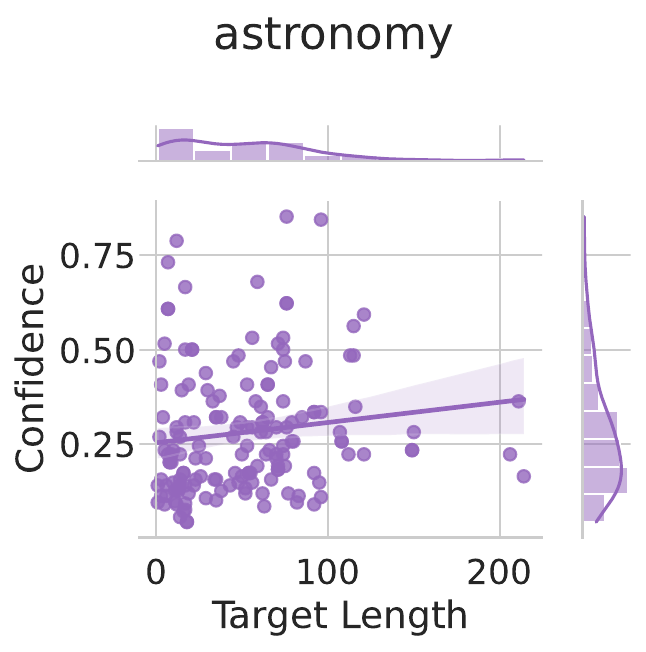} & 
   \includegraphics[width=.2\linewidth]{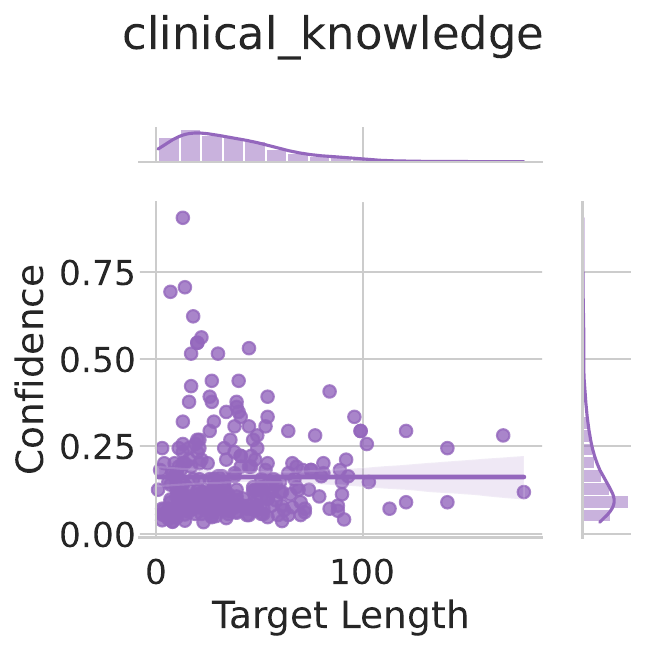} \\ 
   \includegraphics[width=.2\linewidth]{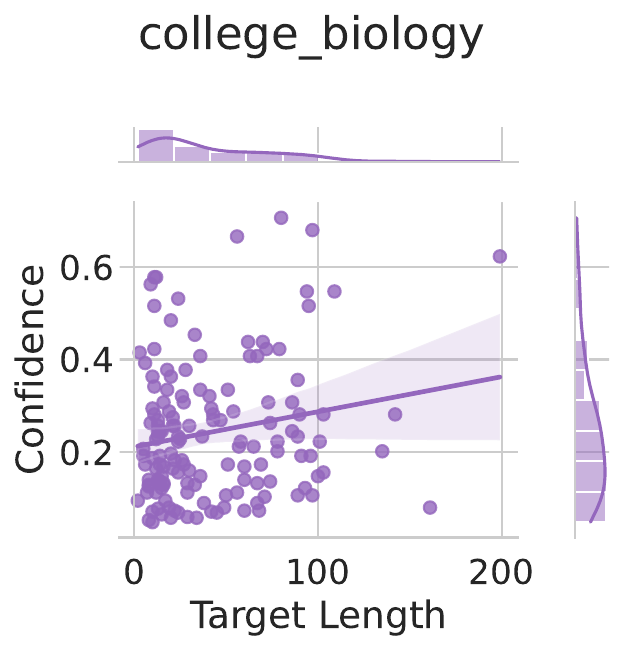} & 
   \includegraphics[width=.2\linewidth]{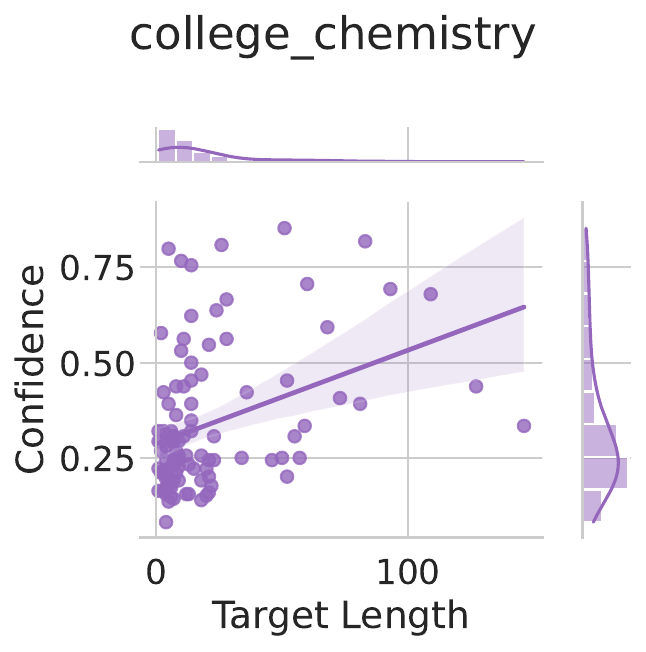} & 
   \includegraphics[width=.2\linewidth]{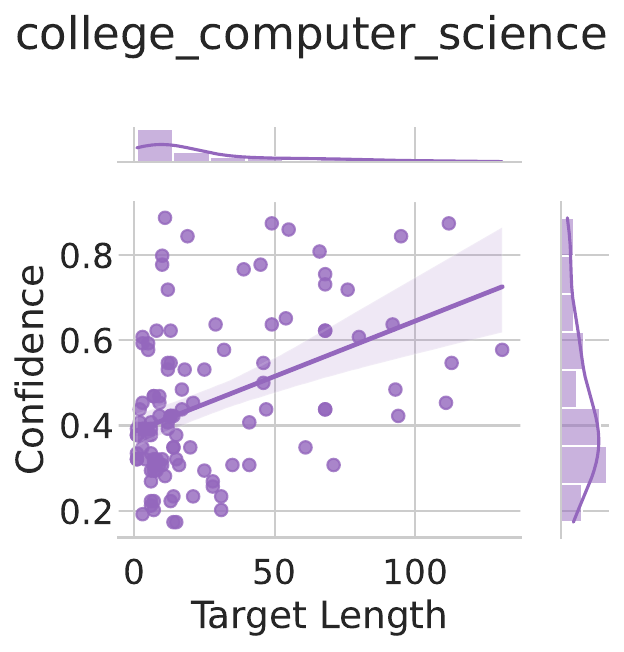} &
   \includegraphics[width=.2\linewidth]{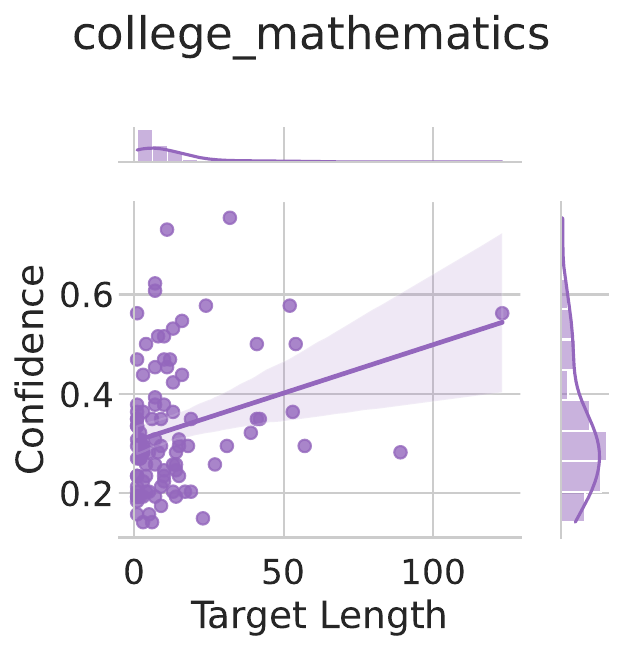} \\
   \includegraphics[width=.2\linewidth]{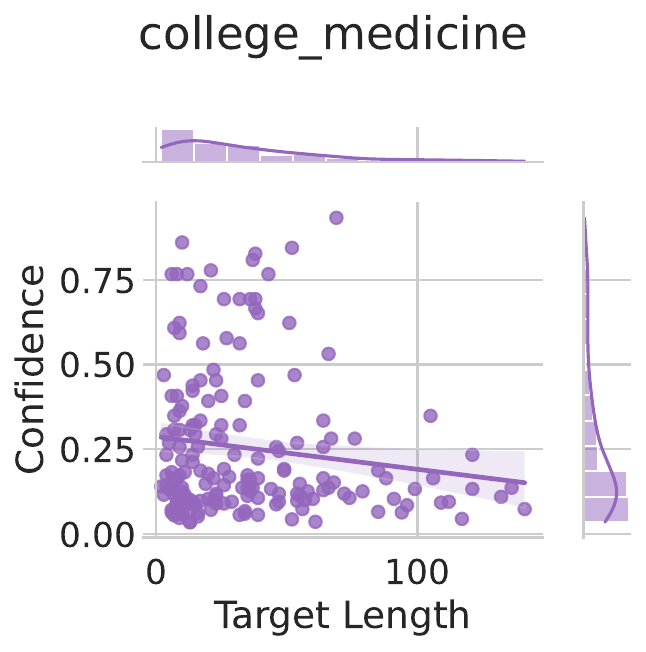} & 
   \includegraphics[width=.2\linewidth]{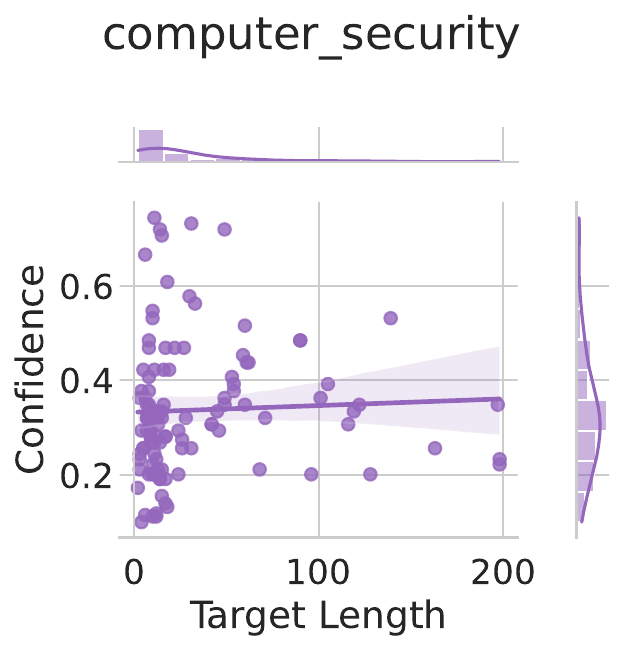} & 
   \includegraphics[width=.2\linewidth]{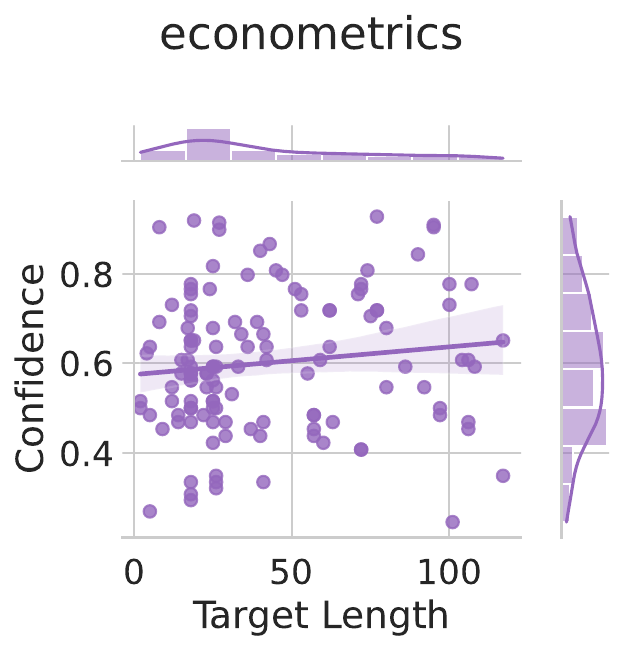} & 
   \includegraphics[width=.2\linewidth]{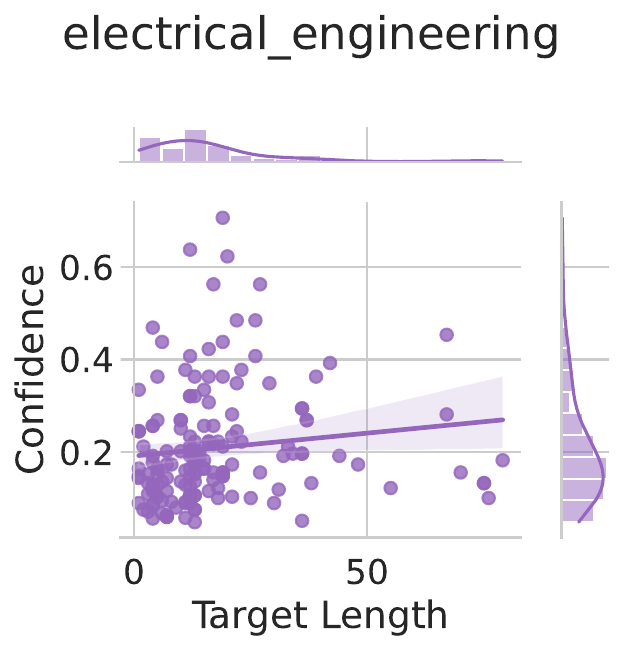} \\ 
   \includegraphics[width=.2\linewidth]{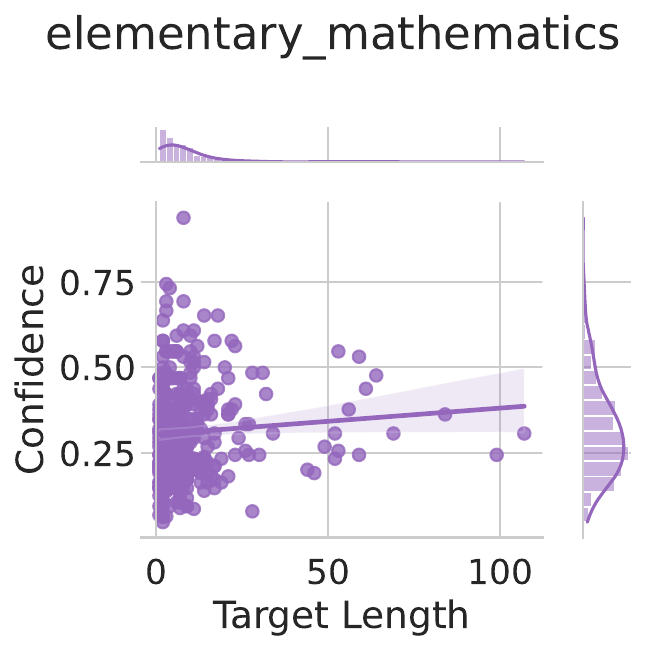} & 
   \includegraphics[width=.2\linewidth]{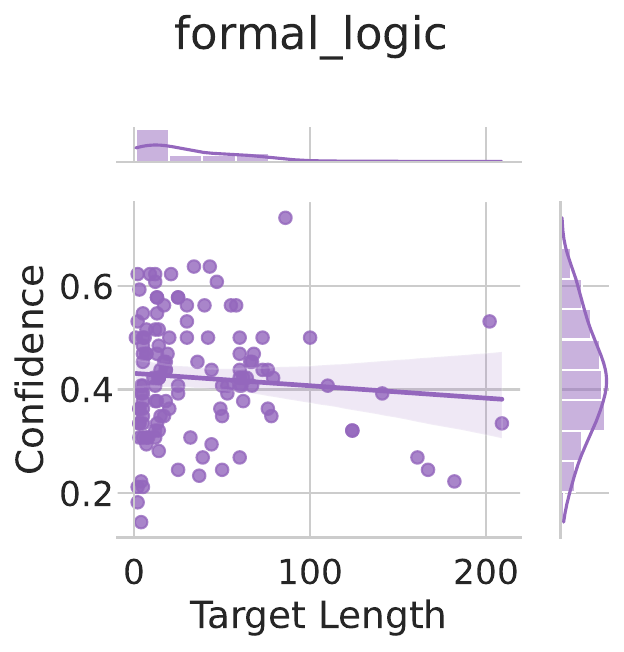} & 
   \includegraphics[width=.2\linewidth]{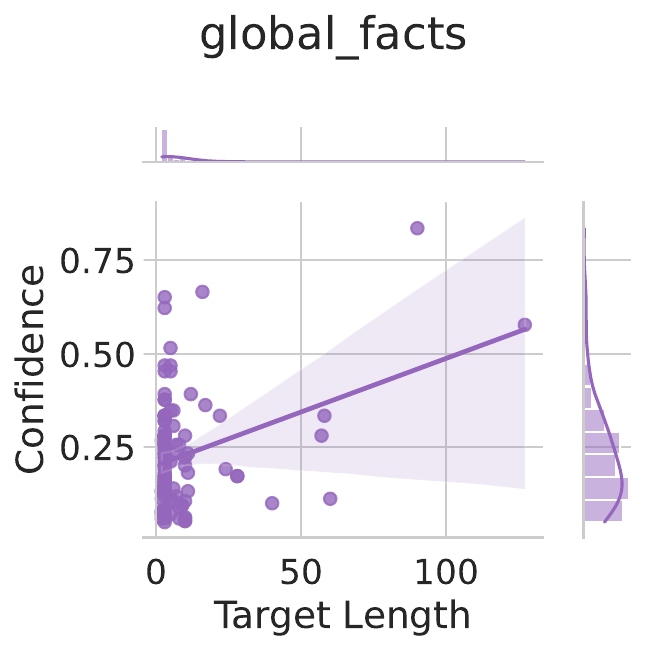} & 
   \includegraphics[width=.2\linewidth]{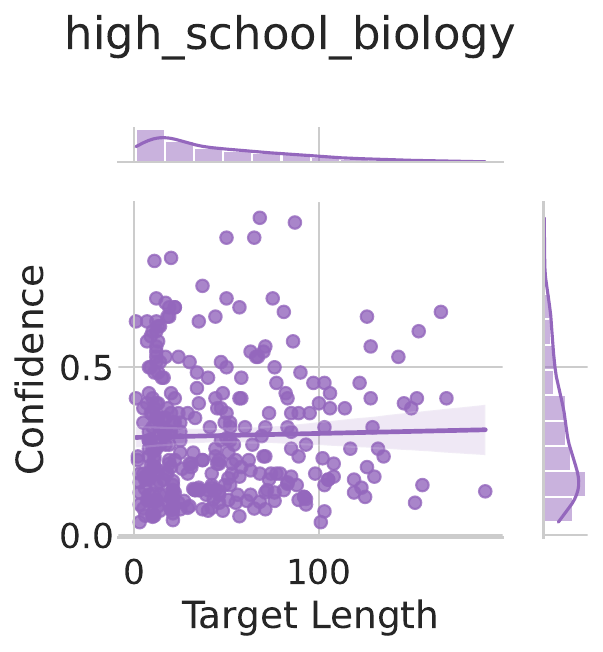} \\ 
   \includegraphics[width=.2\linewidth]{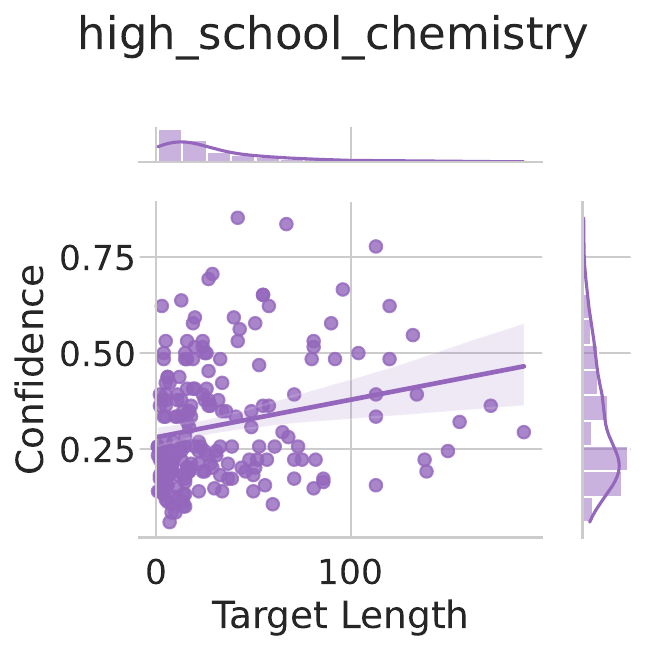} & 
   \includegraphics[width=.2\linewidth]{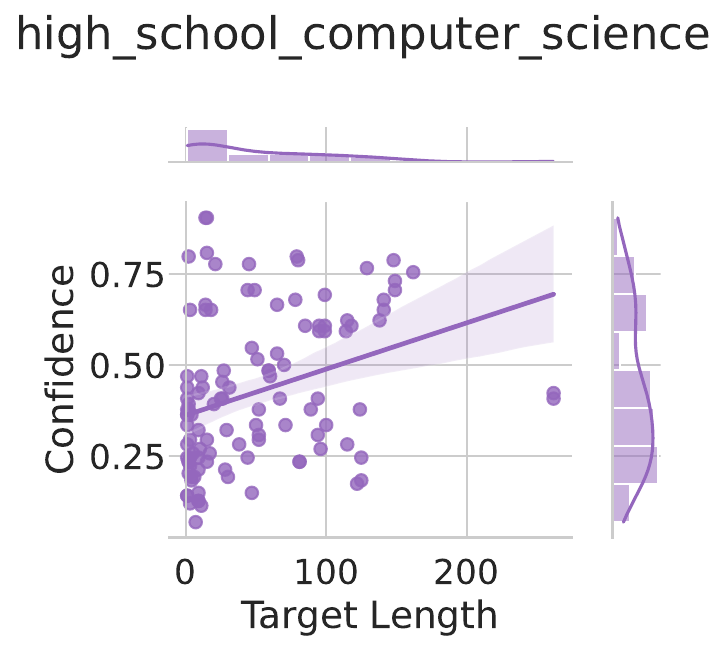} & 
   \includegraphics[width=.2\linewidth]{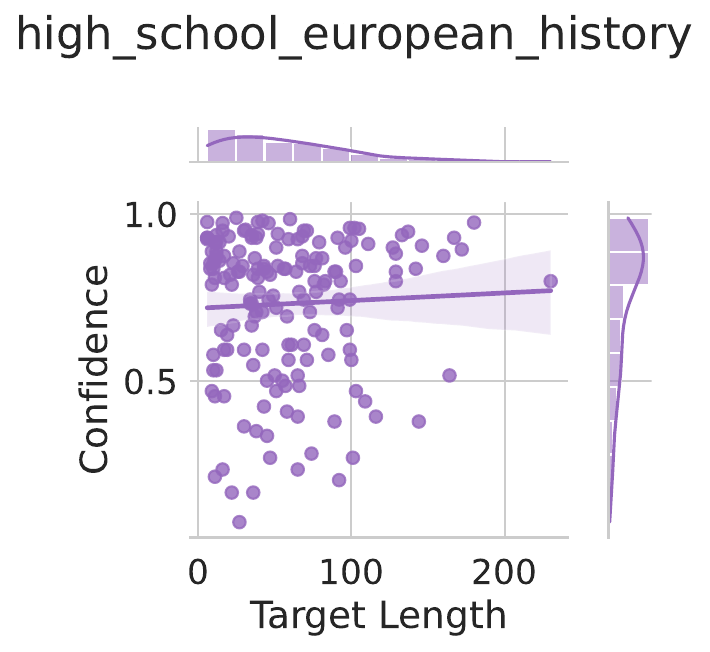} & 
   \includegraphics[width=.2\linewidth]{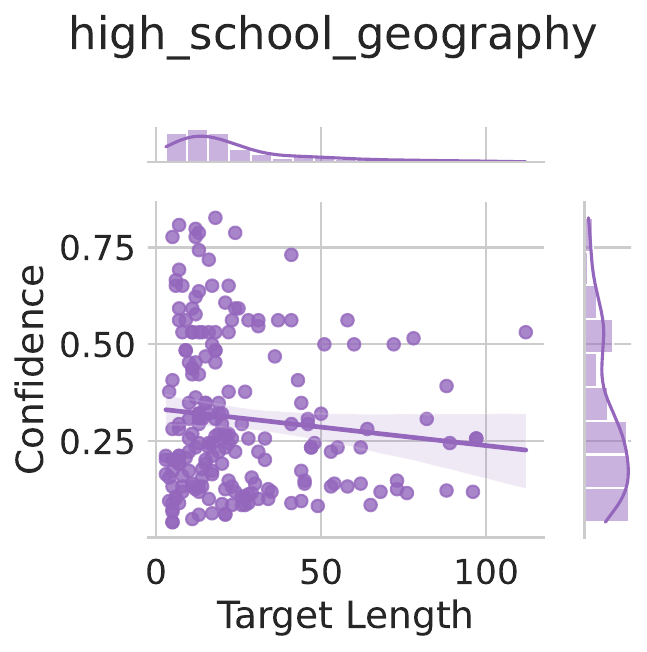} \\ 
   \includegraphics[width=.2\linewidth]{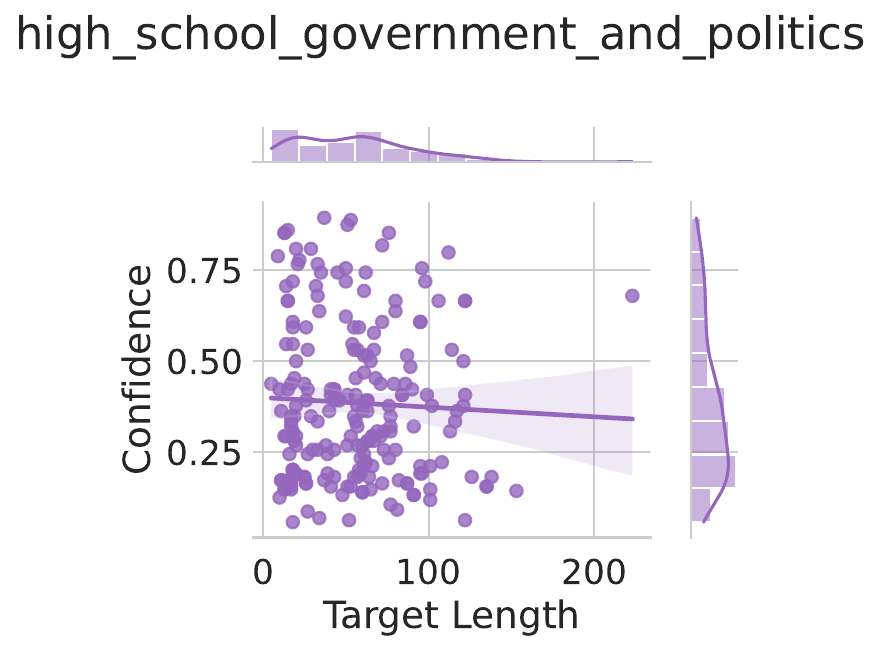} & 
   \includegraphics[width=.2\linewidth]{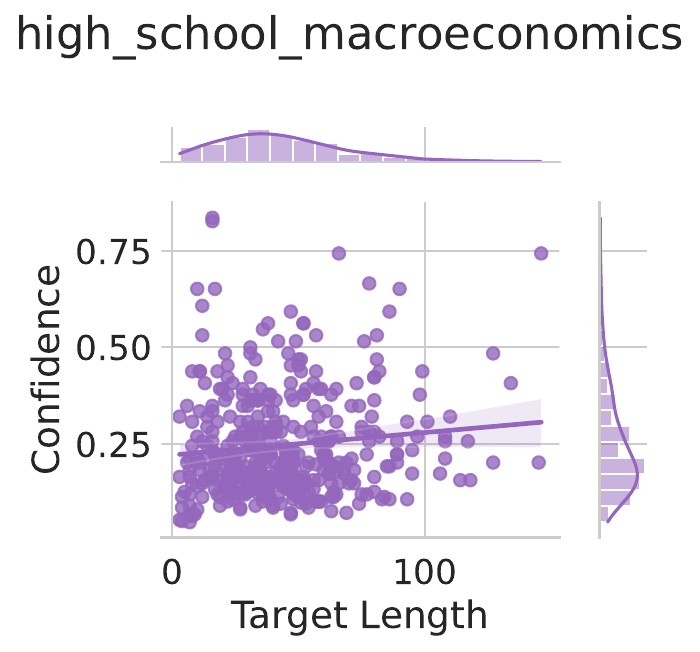} &
       \includegraphics[width=.2\linewidth]{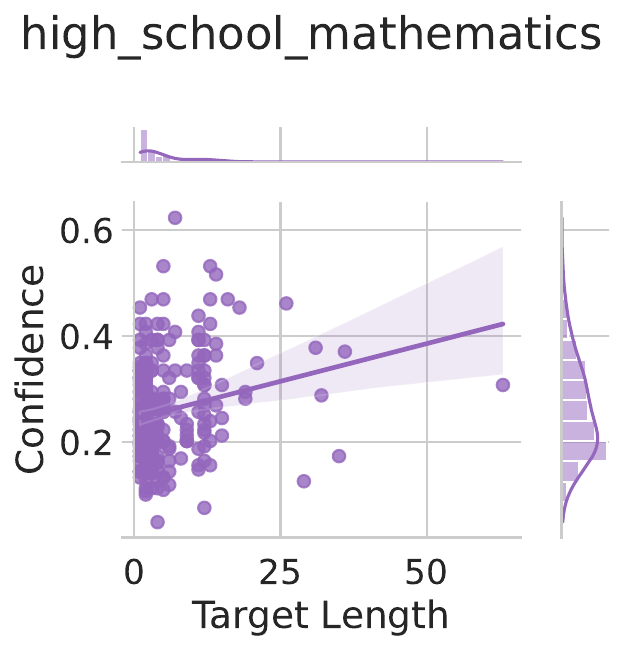} &
\includegraphics[width=.2\linewidth]{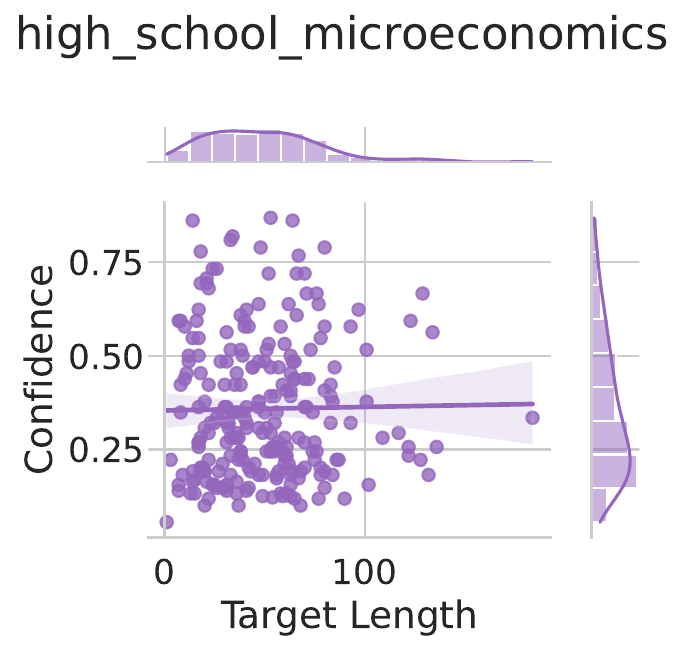}
\end{tabular}
    \caption{Confidence versus Target Length for various MMLU subsets. A horizontal regression line indicates weak correlation of confidence with the target length. See \cref{fig:conf_vs_target_len_2,fig:conf_vs_target_len_3} for other subsets.}
    \label{fig:conf_vs_target_len_1}
\end{figure*}

\begin{figure*}[!t]
    \centering
\begin{tabular}{cccc}
    \includegraphics[width=.2\linewidth]{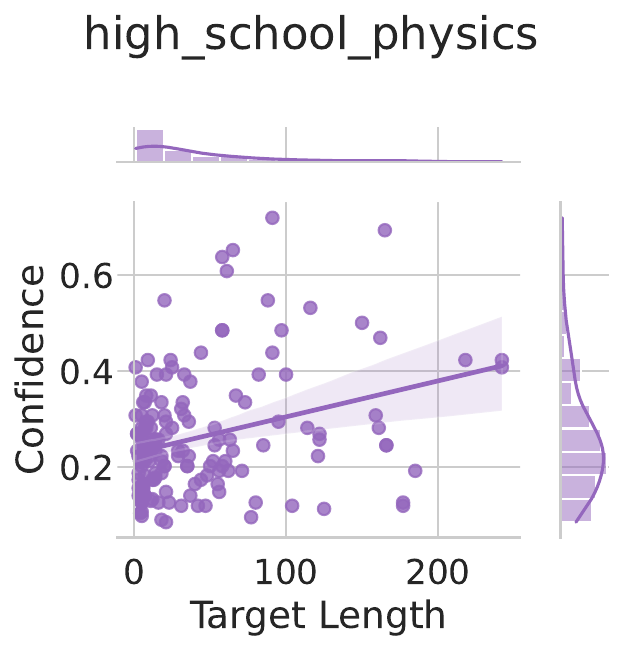} &
    \includegraphics[width=.2\linewidth]{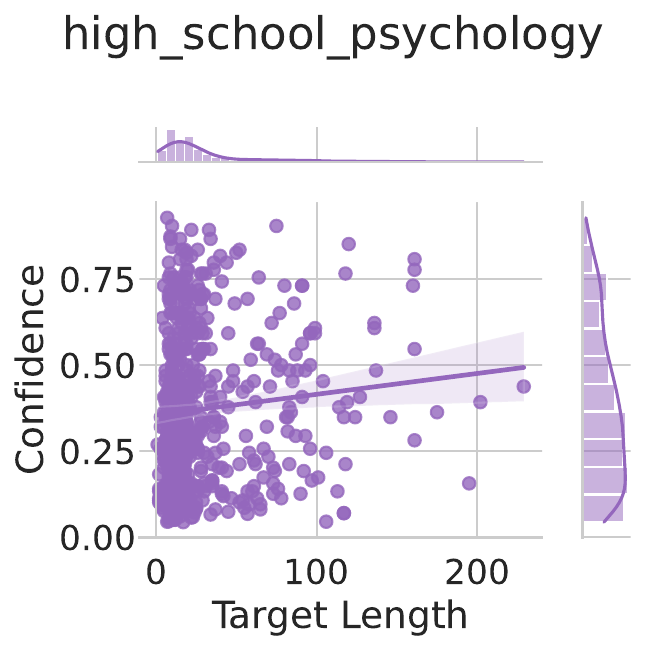} & 
    \includegraphics[width=.2\linewidth]{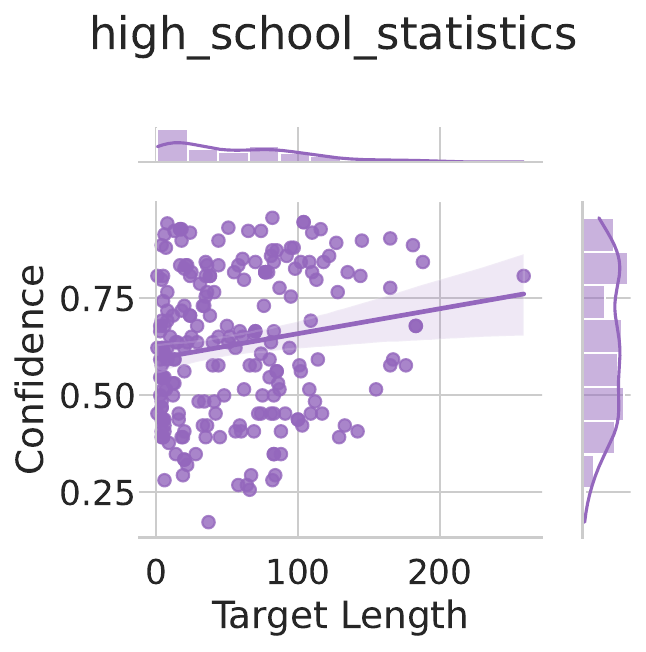} &
    \includegraphics[width=.2\linewidth]{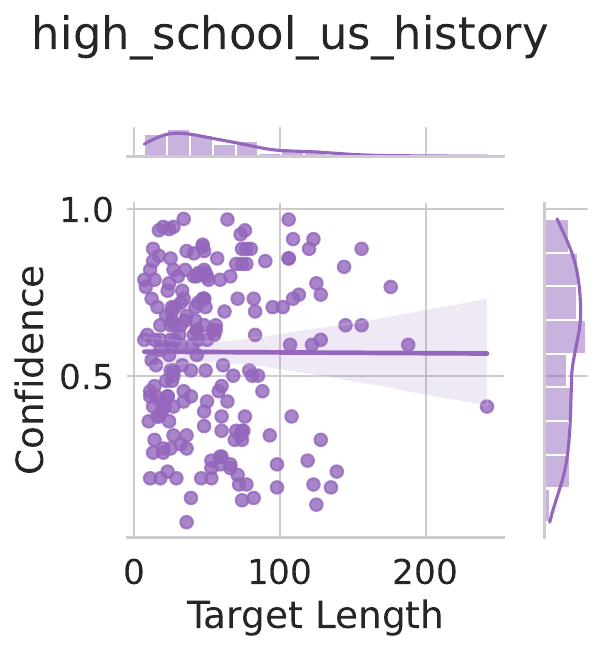} \\ 
    \includegraphics[width=.2\linewidth]{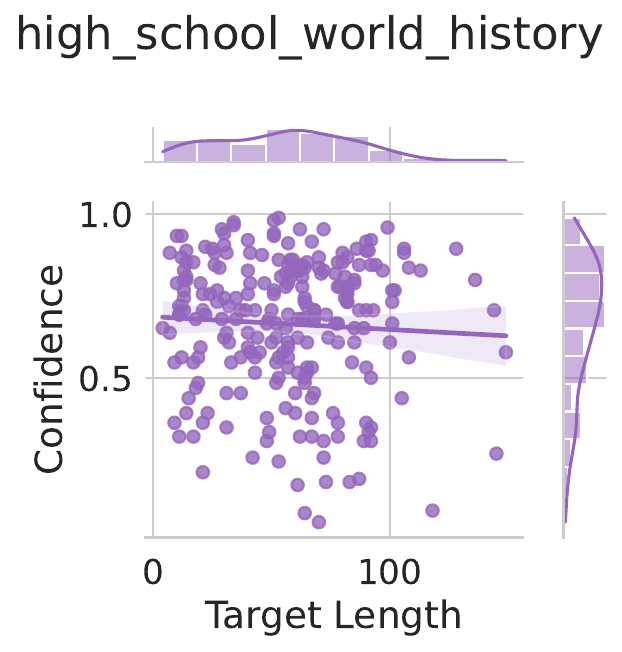} &
    \includegraphics[width=.2\linewidth]{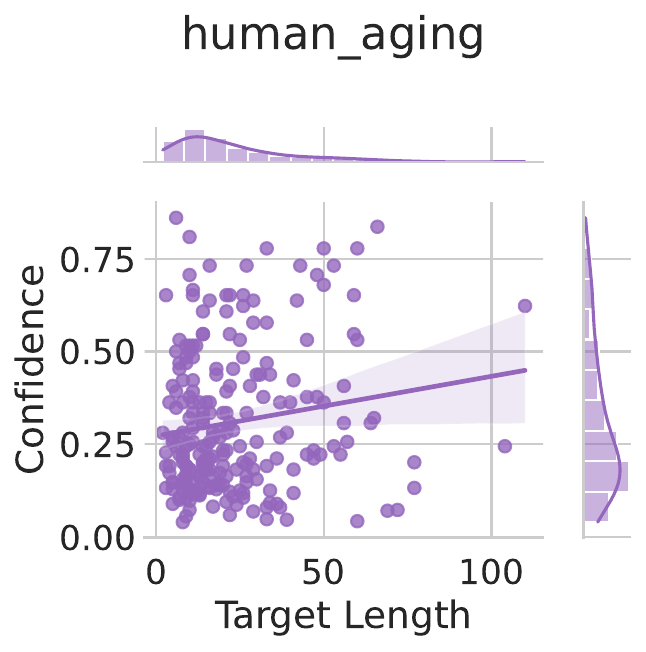} & 
    \includegraphics[width=.2\linewidth]{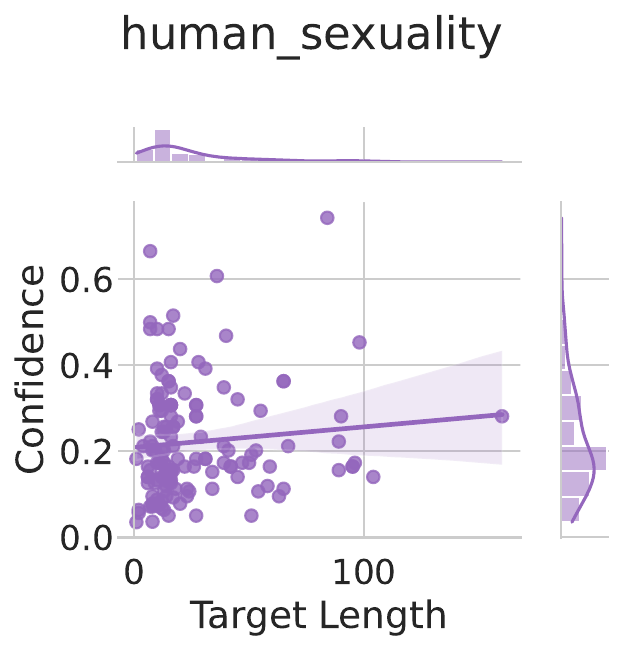} &
    \includegraphics[width=.2\linewidth]{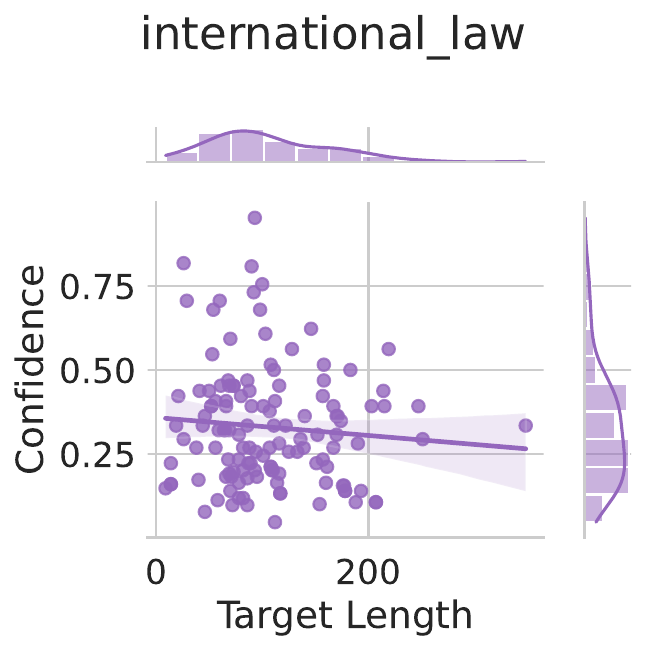} \\ 
    \includegraphics[width=.2\linewidth]{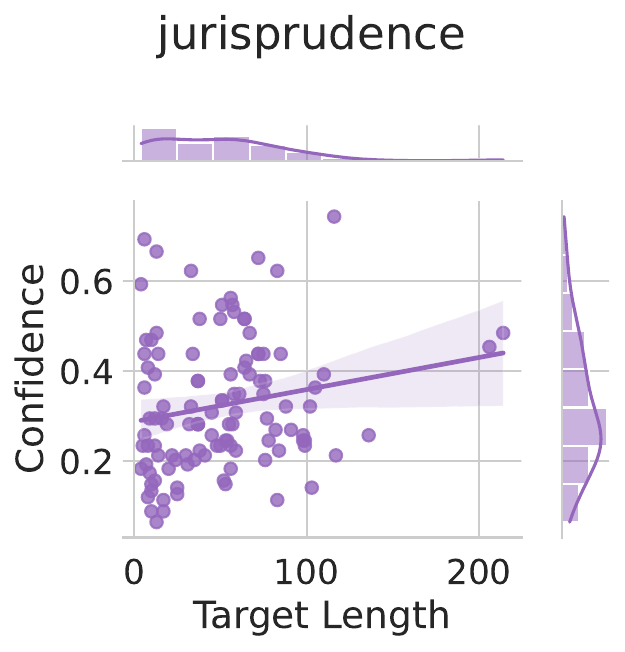} & 
    \includegraphics[width=.2\linewidth]{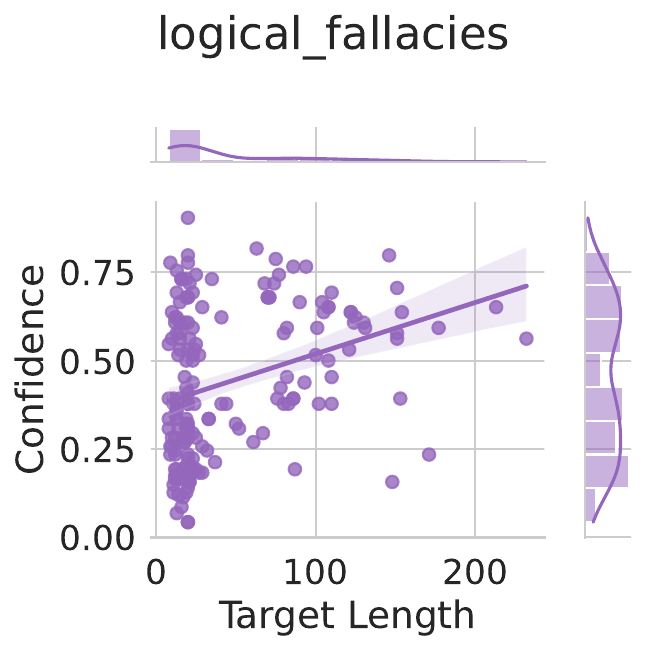} & 
    \includegraphics[width=.2\linewidth]{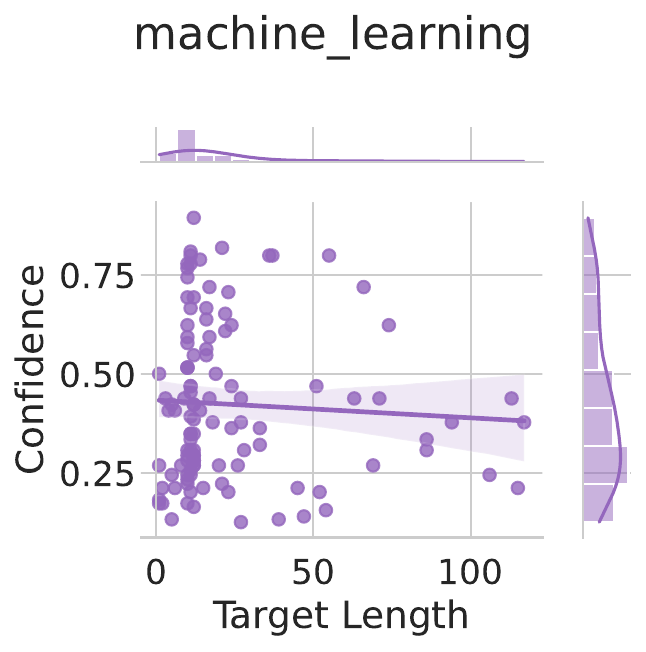} & 
    \includegraphics[width=.2\linewidth]{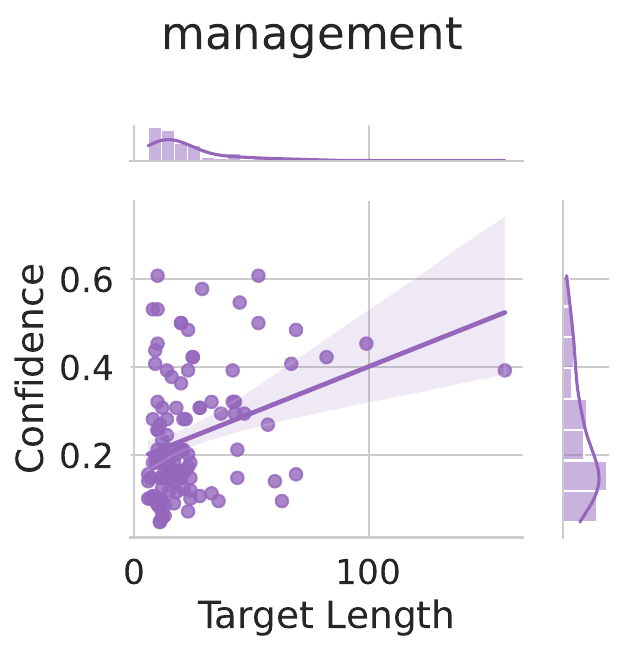} \\
    \includegraphics[width=.2\linewidth]{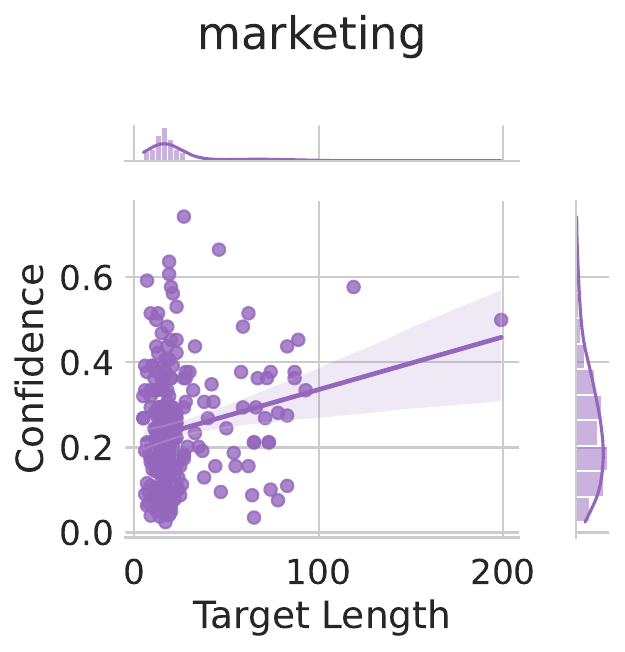} & 
    \includegraphics[width=.2\linewidth]{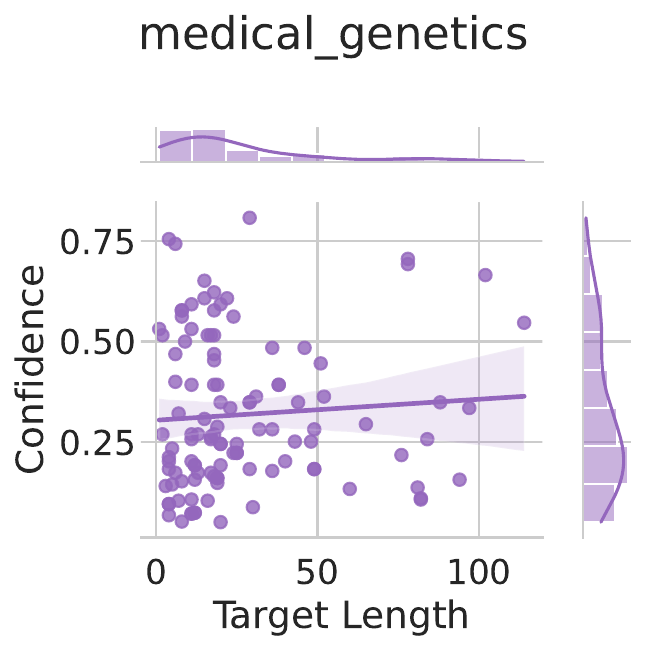} & 
    \includegraphics[width=.2\linewidth]{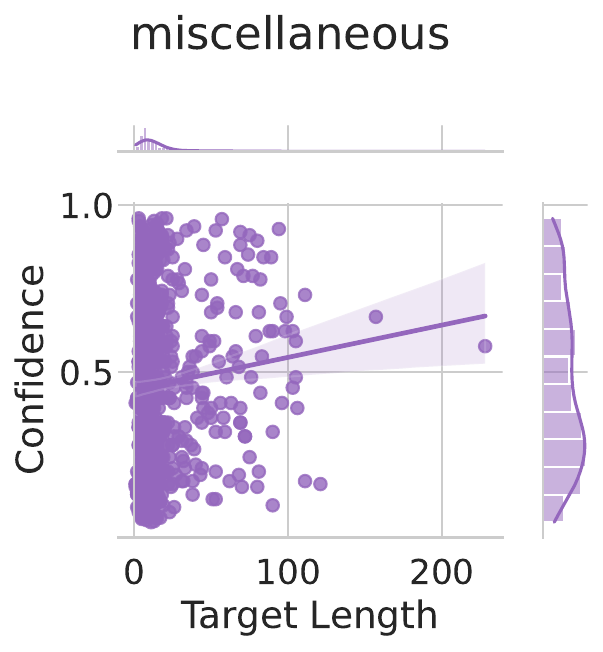} &
    \includegraphics[width=.2\linewidth]{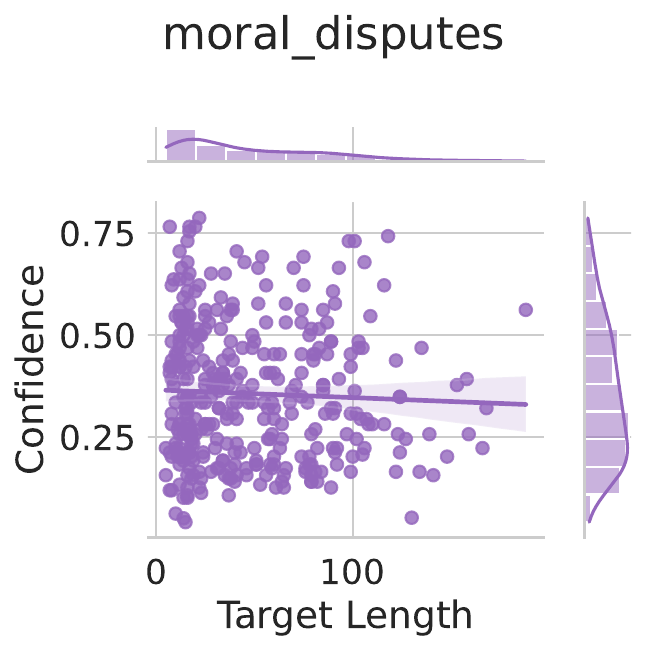} \\ 
    \includegraphics[width=.2\linewidth]{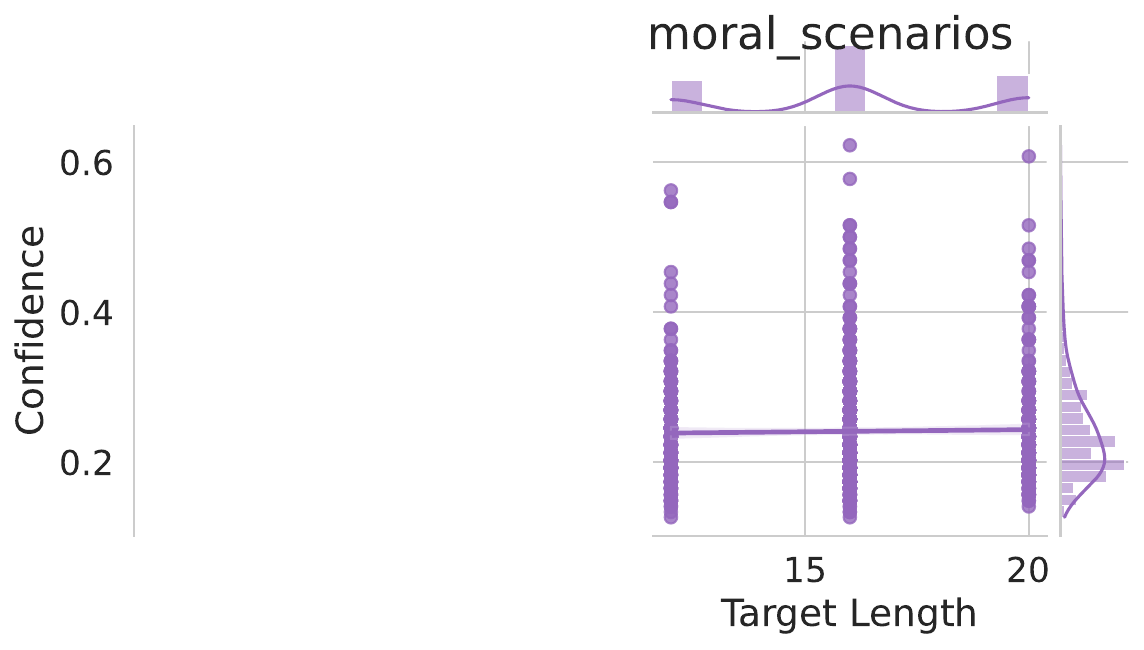} & 
    \includegraphics[width=.2\linewidth]{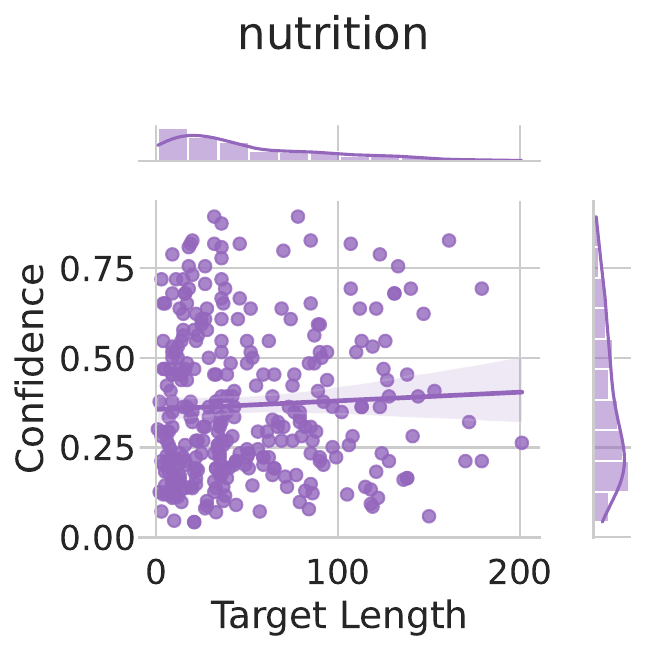} & 
    \includegraphics[width=.2\linewidth]{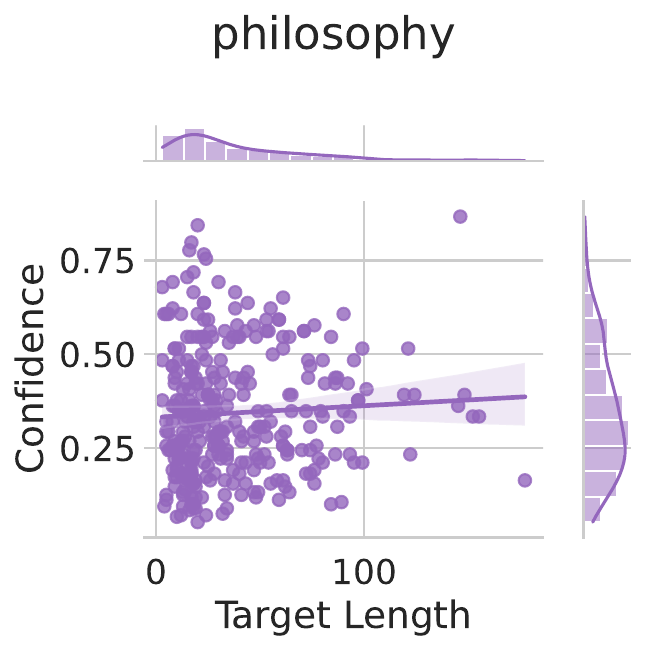} & 
    \includegraphics[width=.2\linewidth]{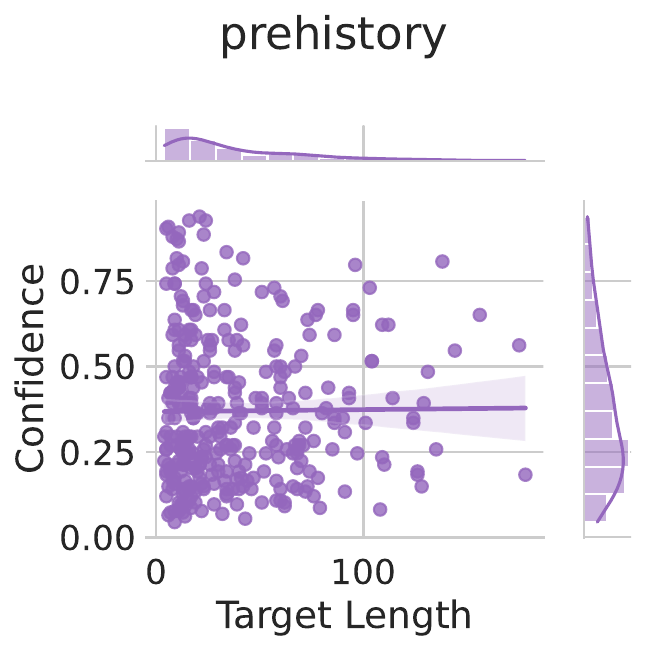} \\ 
    \includegraphics[width=.2\linewidth]{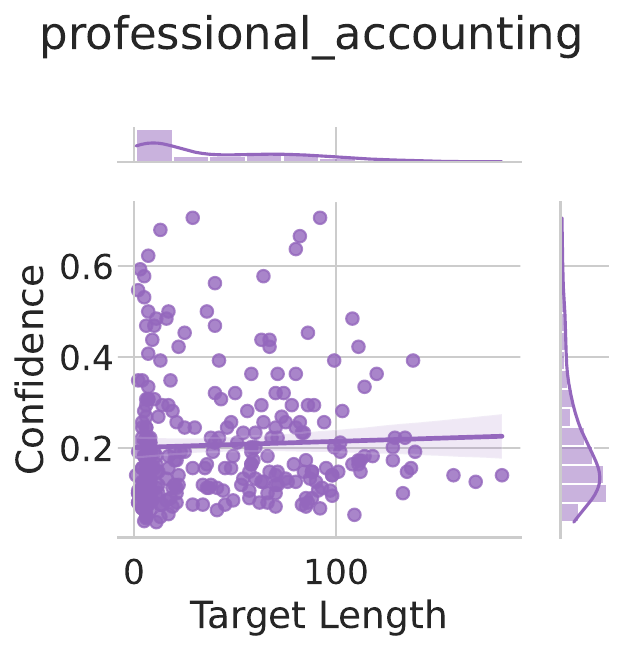} & 
    \includegraphics[width=.2\linewidth]{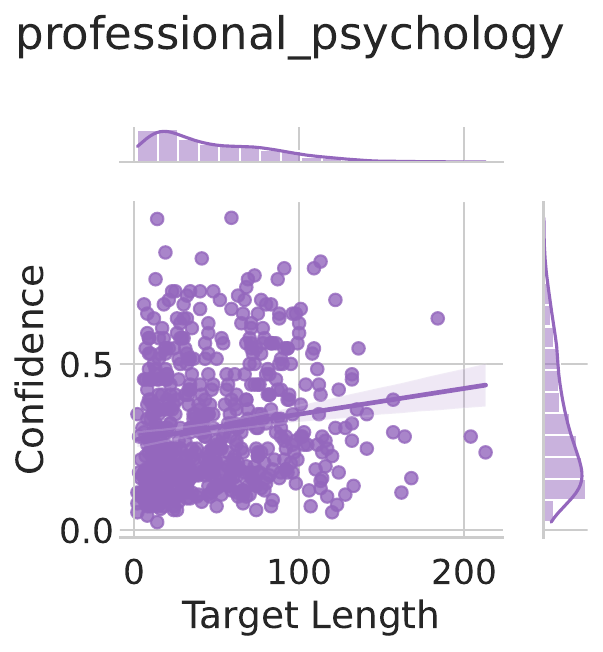} & 
    \includegraphics[width=.2\linewidth]{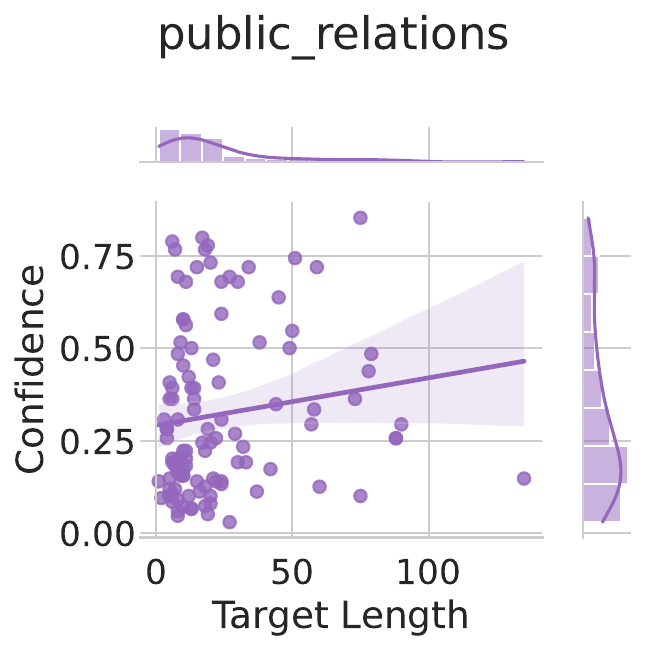} &
    \includegraphics[width=.2\linewidth]{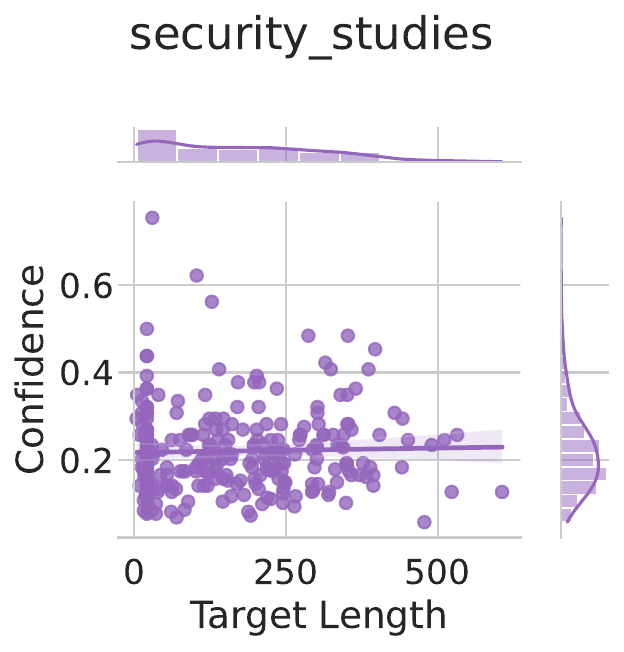}
\end{tabular}
    \caption{Continuing from \cref{fig:conf_vs_target_len_1}. See also \cref{fig:conf_vs_target_len_3}.}
    \label{fig:conf_vs_target_len_2}
\end{figure*}

\begin{figure*}[!t]
    \centering
\begin{tabular}{cccc}
    \includegraphics[width=.2\linewidth]{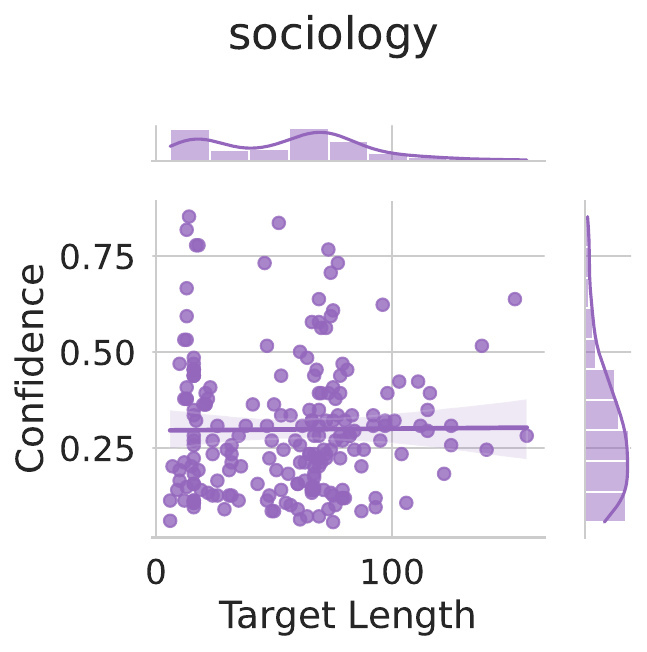} &
    \includegraphics[width=.2\linewidth]{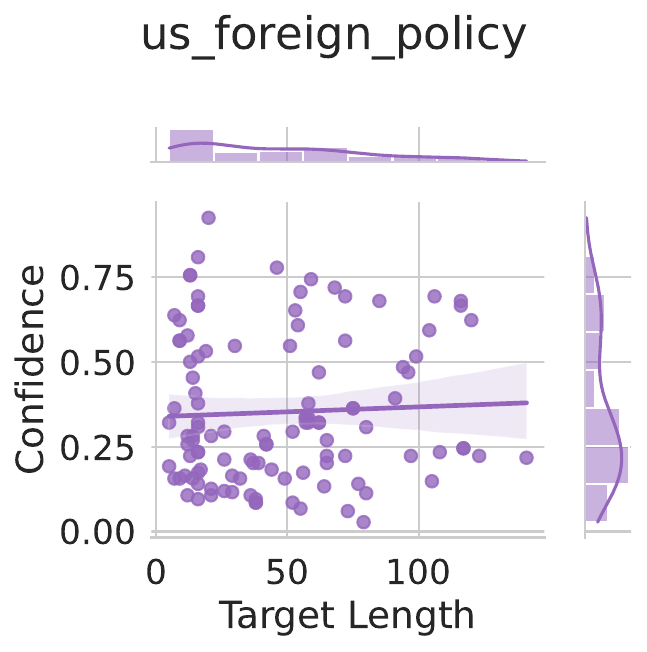} &
        \includegraphics[width=.2\linewidth]{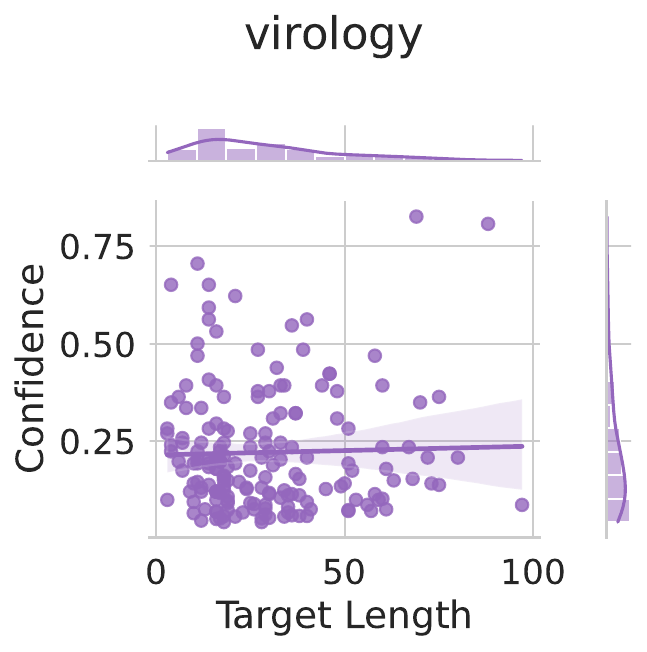} &
        \includegraphics[width=.2\linewidth]{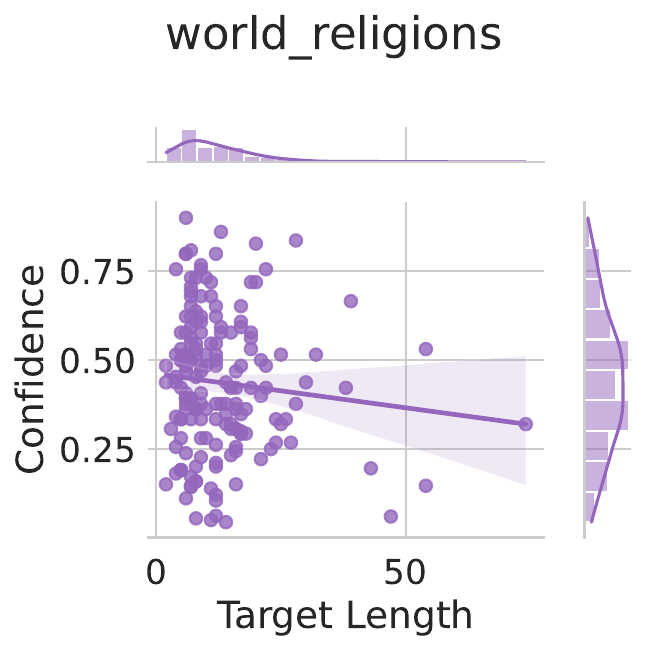}
\end{tabular}
    \caption{Continuing from \cref{fig:conf_vs_target_len_1,fig:conf_vs_target_len_2}.}
    \label{fig:conf_vs_target_len_3}
\end{figure*}

A key limitation of using token likelihoods is that they necessarily decay with the length of the generation. 
In \cref{fig:conf_vs_target_len_1,fig:conf_vs_target_len_2,fig:conf_vs_target_len_3}, we confirm over all subsets of MMLU that the length of the target does not strongly correlate with the confidence associated with the targets.
This behavior is an essential ingredient towards an effective confidence estimation in practice, such that longer sequences are not penalized in confidence despite being correct.

\section{Generalization to Coding Tasks}
\label{sec:generalization-coding}

Because there are no coding tasks in our training dataset, we can use a coding competition task introduced in LiveCodeBench~\citep{jain2024livecodebench} to assess how well finetuned uncertainty estimation methods perform on completely out of distribution tasks.

To conduct the analysis in \cref{tab:coding}, we evaluate several base models on the 62 LeetCode easy questions from the livecodebench\_generation\_lite task. We asking for the model to write a Python solution and grade the solution using test cases (marking it as correct iff it passes all test cases). We then apply \texttt{Lora + Prompt} and \texttt{Zero-Shot Classifier} uncertainty estimation methods---with these methods \emph{only} using training/temperature scaling data from our main dataset mixture which notably does not include any coding tasks \cref{app:training-data}. Accuracy is shown to contextualize the model's overall level of performance on the task. On Mistral-7B, the best performing model on the coding task, the supervised \texttt{Lora + Prompt} approach dramatically improves calibration and selective prediction as compared to \texttt{Zero-Shot Classifier}; on the worse-performing Mistral-7B-Instruct and LLaMa-2-7B, selective prediction improves but calibration slightly degrades.

\begin{table}
    \centering
    \begin{tabular}{c|c|c|c|c}
    \toprule
    Model & Method & Acc & ECE & AUROC \\
    \hline
    \multirow{2}{*}{LLaMa-2-7B}	&	Zero-Shot Classifier	&	3.2\%	&	41.0\%	&	56.9\%	\\	
    	&	Lora + Prompt	&	3.2\%	&	46.4\%	&	80.0\%	\\	
    \multirow{2}{*}{Mistral-7B}	&	Zero-Shot Classifier	&	27.4\%	&	70.2\%	&	66.2\%	\\	
   	&	Lora + Prompt	&	27.4\%	&	21.4\%	&	85.1\%	\\	
    \multirow{2}{*}{Mistral-7B-Instruct}	&	Zero-Shot Classifier	&	21.0\%	&	52.7\%	&	47.1\%	\\	
    	&	Lora + Prompt	&	21.0\%	&	56.1\%	&	70.2\%	\\	
    \bottomrule
    \end{tabular}
    \vspace{1mm}
    \caption{
    ECE and AUROC on livecodebench\_generation\_lite (LeetCode easy subset).
    ECE is shown after temperature scaling on a small hold-out set of the original dataset mixture \cref{app:training-data}. Acc is task accuracy (proportion of coding solutions that are correct). Supervised training (LoRA + Prompt) seems to always improve selective prediction, although supervised training only heavily improves calibration for Mistral-7B and in fact slightly degrades calibration for the two other models.}
    \label{tab:coding}
\end{table}

\section{User Studies}
\label{app:user-study}

\subsection{Additional Details on Setup}

\paragraph{Stimuli and Participant Selection} We closely followed the setup of \citep{bhatt2023learning}. We used the same 180 MMLU questions from which were pre-batched into three sets of 60 MMLU questions. Within each variant, we randomly assigned participants to one of the three batches. In total, we recruit $181$ participants (20 per variant\footnote{With the exception of one extra participant due to random batching allocation effects.}). All participants were recruited through the crowdsourcing platform Prolific \citep{palan2018prolific}; we restrict our participant pool to those based in the United States who speak English as a first language.

\paragraph{Compensation} Participants were told that the study would take approximately 30 minutes and were paid at a base rate of \$9/hr and informed that they would receive an optional bonus up to \$10 for answering questions correctly. We applied the bonus to all participants. 

\paragraph{LLM Answers and Uncertainty Elicitation}
\citeauthor{bhatt2023learning} originally used GPT-3.5 as their LLM. While at first, we explored user performance when provided with confidence scores modulated over the original GPT-3.5 responses that the authors had collected, the authors had filtered LLM performance to ensure the LLM achieved high performance on biology, computer science, and foreign policy and poor performance on mathematics. As such, we noticed that participants overwhelmingly uptook the LLM's answer (which was rational behaviour, given the model's high performance). To explore a more nuanced performance profile, we regenerated LLM answers using Mistral 7B Instruct via greedy decoding. We then generated confidence scores on top of the LLM responses. For our random baseline, we sample a confidence score uniformly between 0 and 100\% for each question. 

\subsection{Important considerations}

There are many reasons to heed caution in interpreting our results as definitive indications of the utility of displaying confidence to users in LLM assistive settings. In particular: (i) users are presented with feedback after each trial as in \citep{bhatt2023learning} -- as such, they can determine (potentially rapidly) whether or not a model is reliable, even without confidence scores. However, in practical settings users may not know whether or not the model was truly correct and therefore confidence scores could have an even larger impact; (ii) MMLU questions can be challenging for non-experts -- we see the biggest differences in performance for the no-LLM vs. any-LLM-assistance condition. We may see a wider range of reliance behaviors in settings wherein people have more confidence in their own abilities; (iii) we present users with numeric confidence; however, humans are not always able to reliably process confidence estimates nor appropriately calibrate uncertainty estimates themselves~\citep{keren1991calibration, vodrahalli2022uncalibrated, collins2023human, lichtenstein1977calibration}. It may be that alternate modes of communicating confidence improve users' ability to appropriately leverage the confidence scores in their decision making process. We see targeted exploration of each component  through interdisciplinary collaboration across AI, behavioral science, and human-computer interaction as ripe for future work.

\subsection{Extended Results}

\paragraph{Task Accuracy and Reliance Sensibility}

\begin{figure*}[!t]
    \centering
    \includegraphics[width=0.4\linewidth]{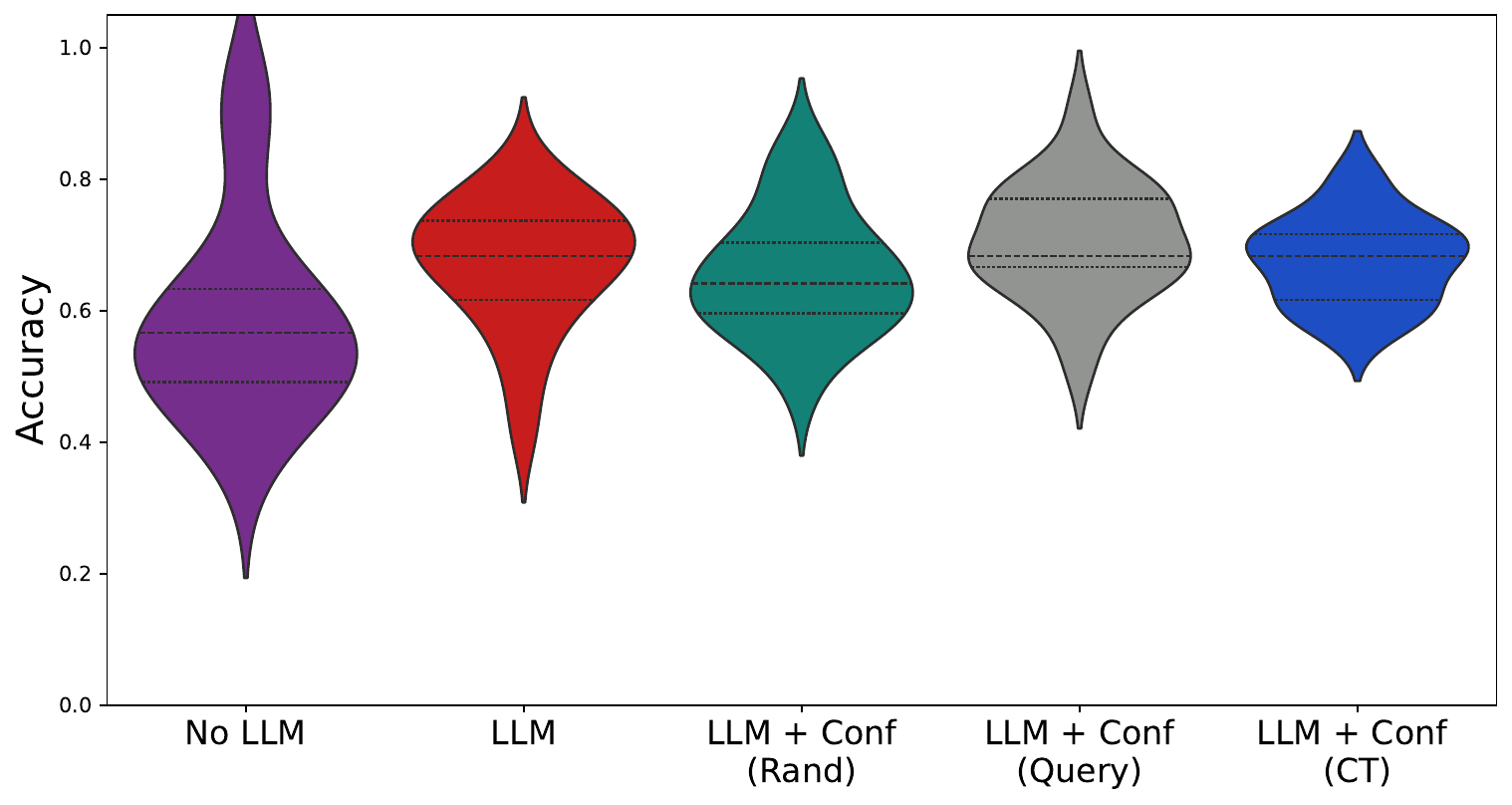}
    \hspace{2mm}
    \includegraphics[width=0.4\linewidth]{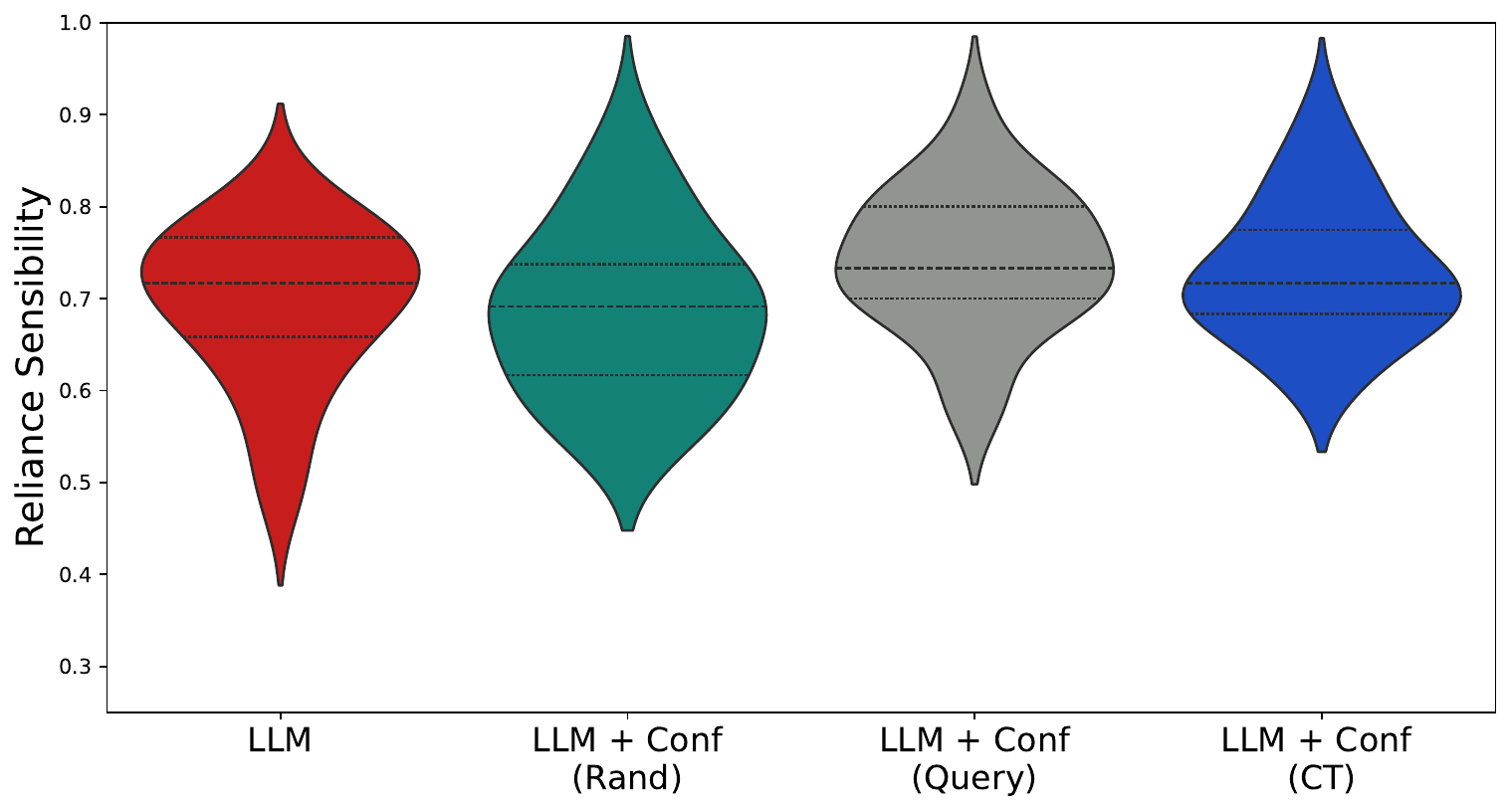}
    \caption{
    (\textbf{Left}) User accuracy on 60 MMLU questions per variant ($N=20$ users per variant); violin plots show quartiles as dashed lines (\textbf{Right}) Average reliance sensibility (proportion of instances where the user sided with the model when the model was correct, and overrode the model's prediction when the model was incorrect); higher indicates better reliance calibration.}
    \label{fig:user-study-extended}
\end{figure*}

We depict average user task accuracy and reliance sensibility across variants in Figure \ref{fig:user-study-extended}. We follow \citeauthor{bhatt2023learning} in computing reliance sensibility as the proportion of times the user appropriately sided with the model prediction when the model was correct and did not respond with the model's prediction when the model was incorrect. 

\begin{figure*}[!t]
    \centering
    \includegraphics[width=0.4\linewidth]{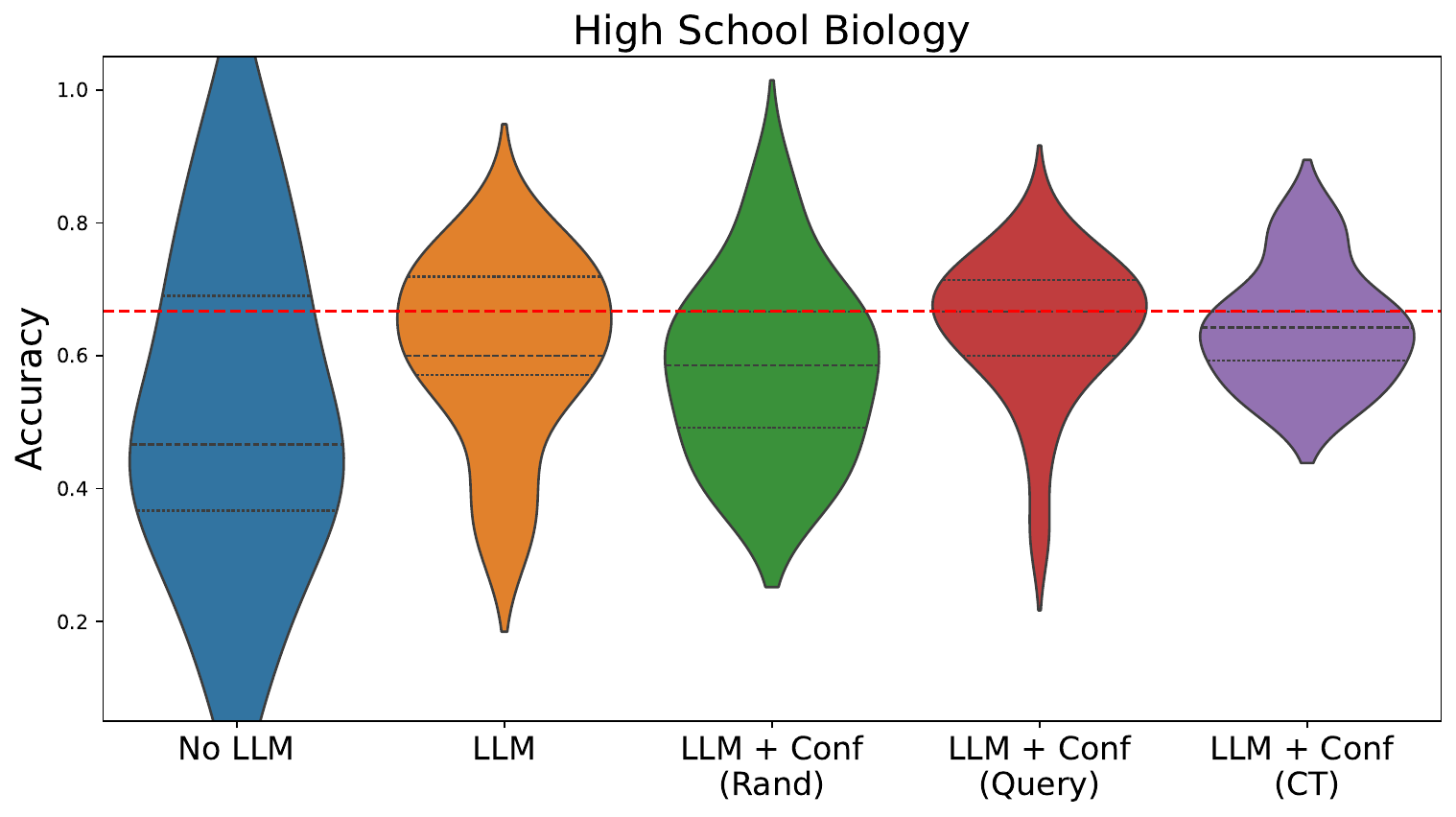}
    \hspace{3mm}
    \includegraphics[width=0.4\linewidth]{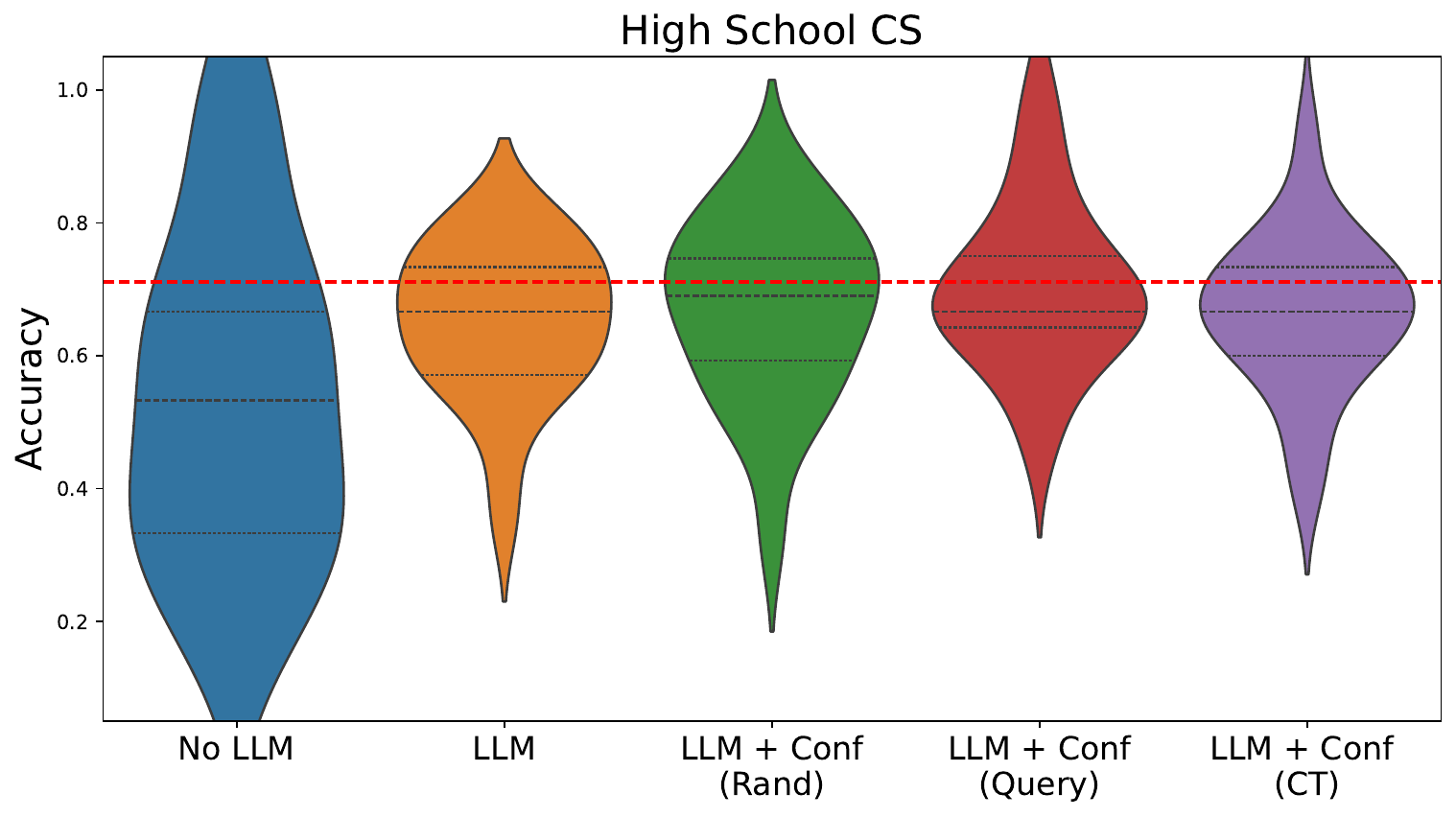} \\
    \vspace{3mm}
    \includegraphics[width=0.4\linewidth]{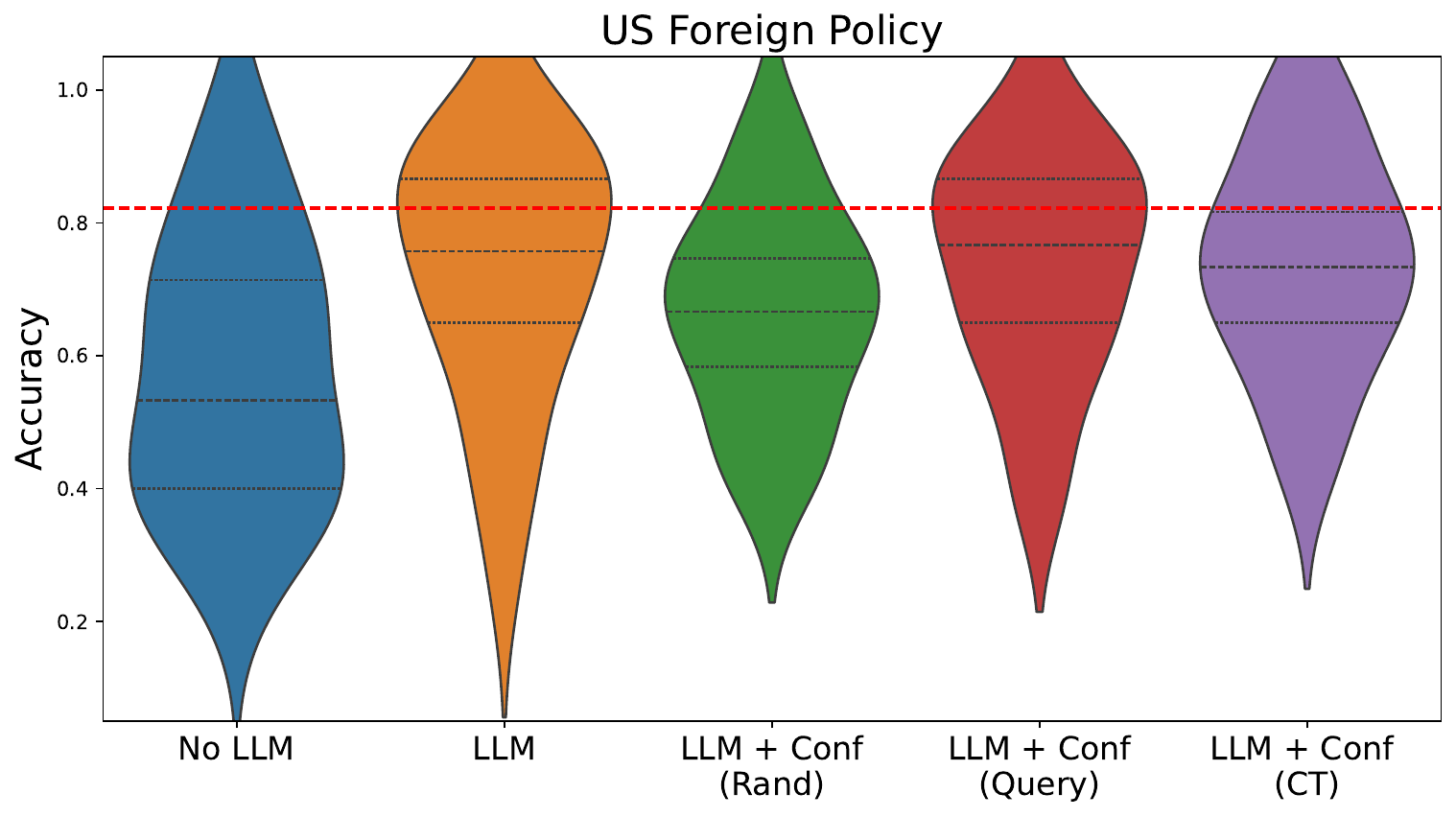}
    \hspace{3mm}
    \includegraphics[width=0.4\linewidth]{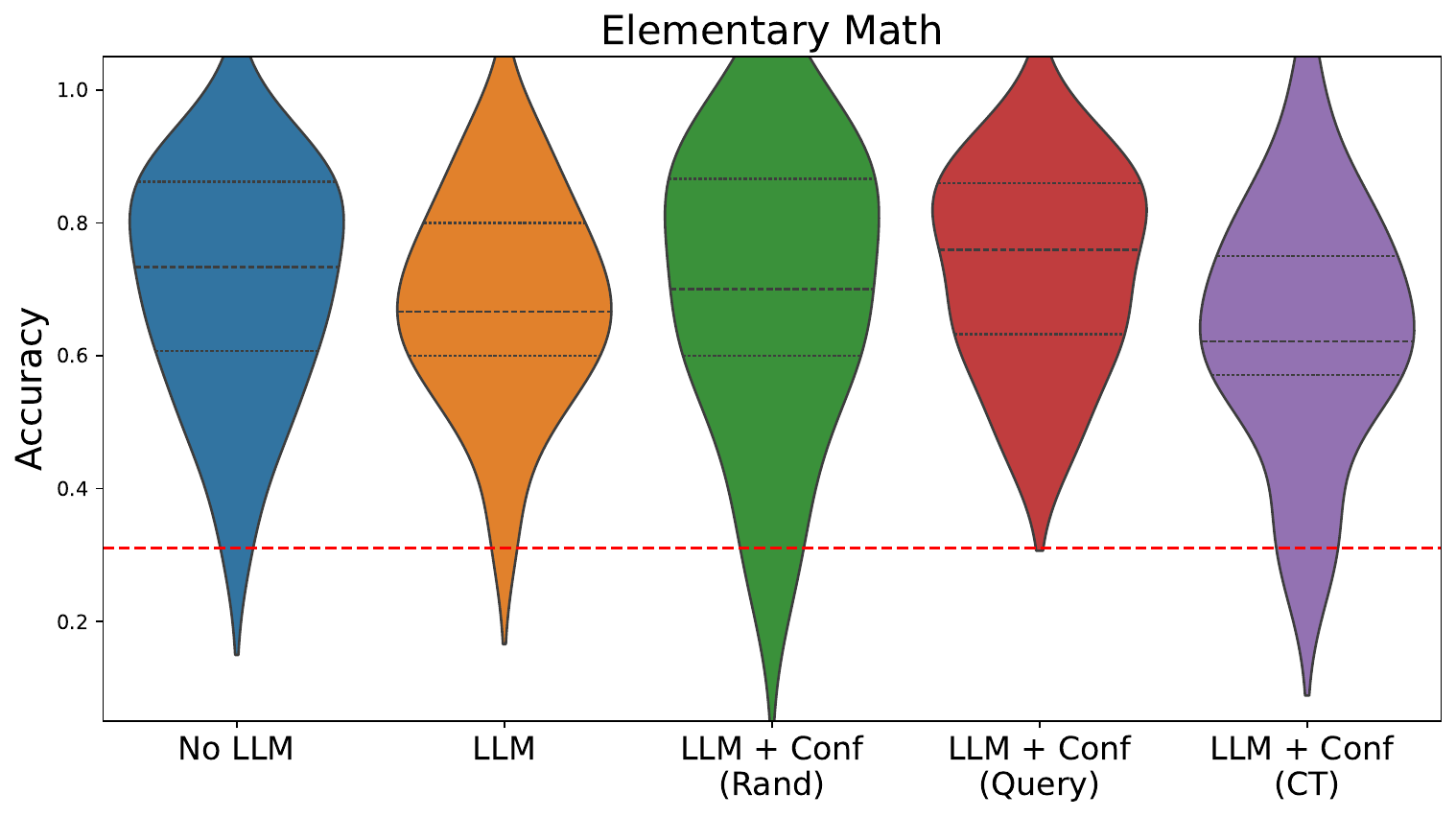}
    \caption{
    User accuracies per topic for the Mistral variants. Red line indicates the model's average accuracy.}
    \label{fig:per-topic-acc-mistral}
\end{figure*}

We depict per-topic accuracy, with the LLM's average performance in Figure \ref{fig:per-topic-acc-mistral}.

\paragraph{GPT-3.5 Confidence Generalization}

\begin{figure*}[!t]
    \centering
    \includegraphics[width=0.4\linewidth]{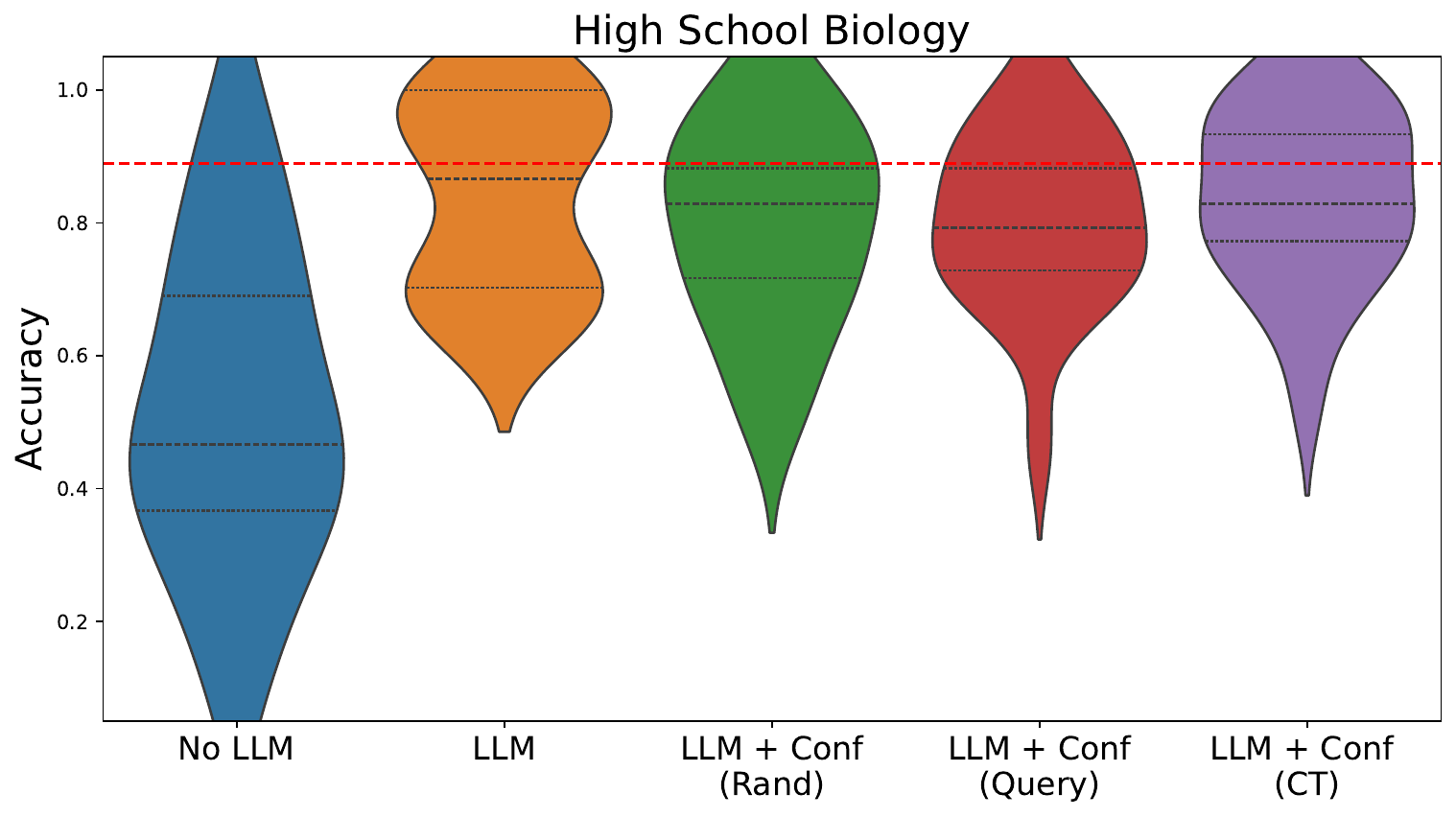}
    \hspace{3mm}
    \includegraphics[width=0.4\linewidth]{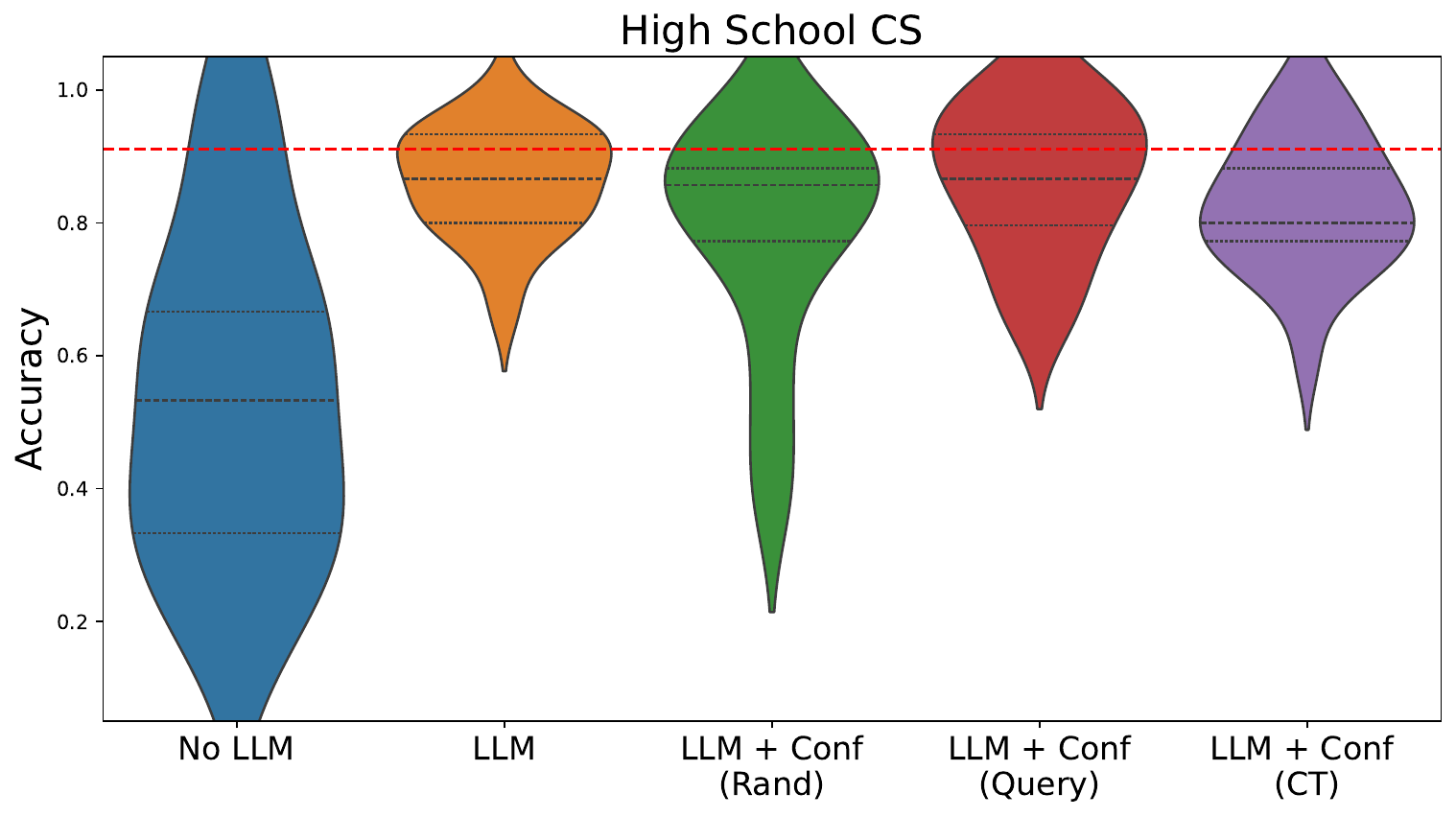} \\
    \vspace{3mm}
    \includegraphics[width=0.4\linewidth]{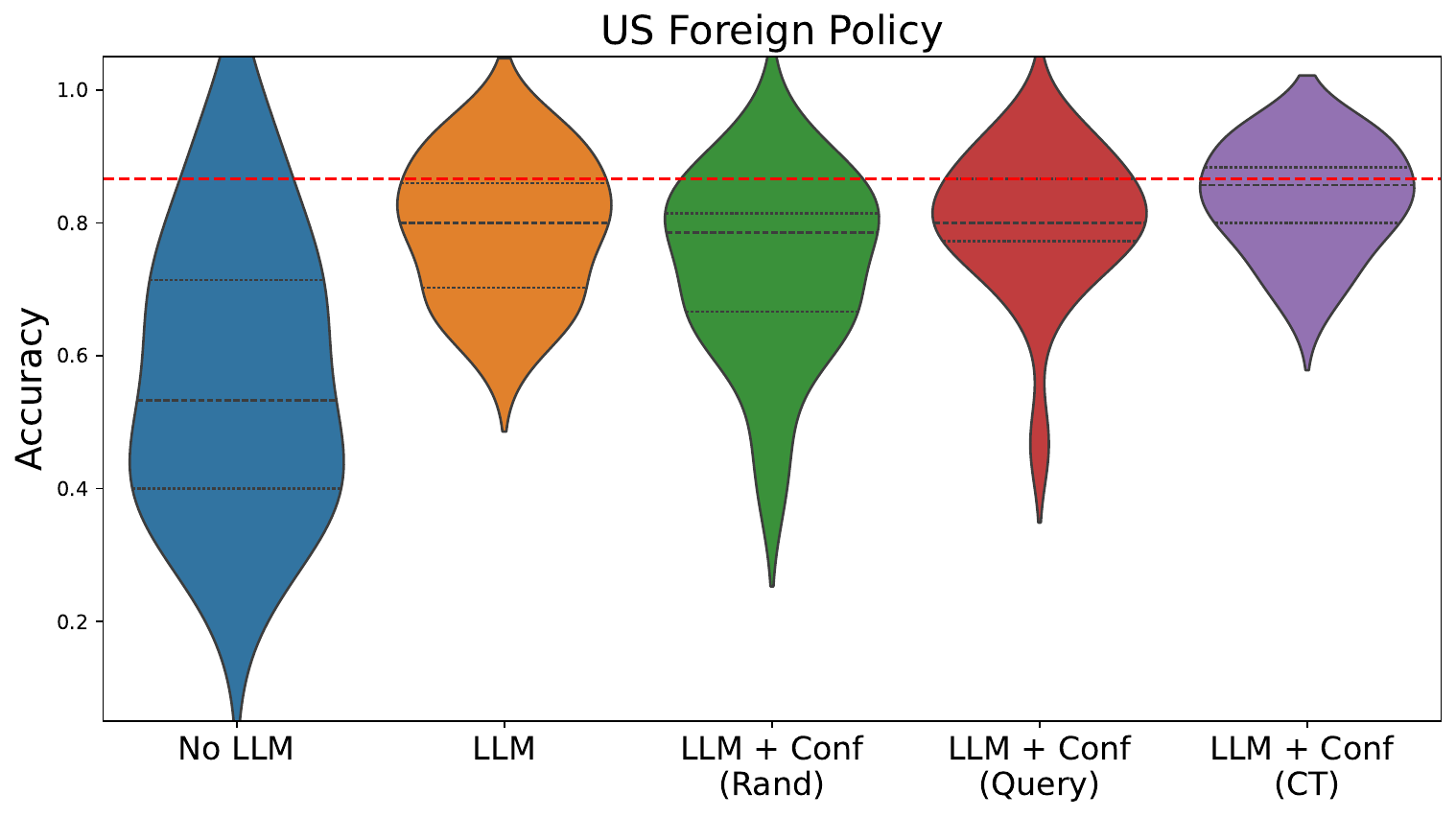}
    \hspace{3mm}
    \includegraphics[width=0.4\linewidth]{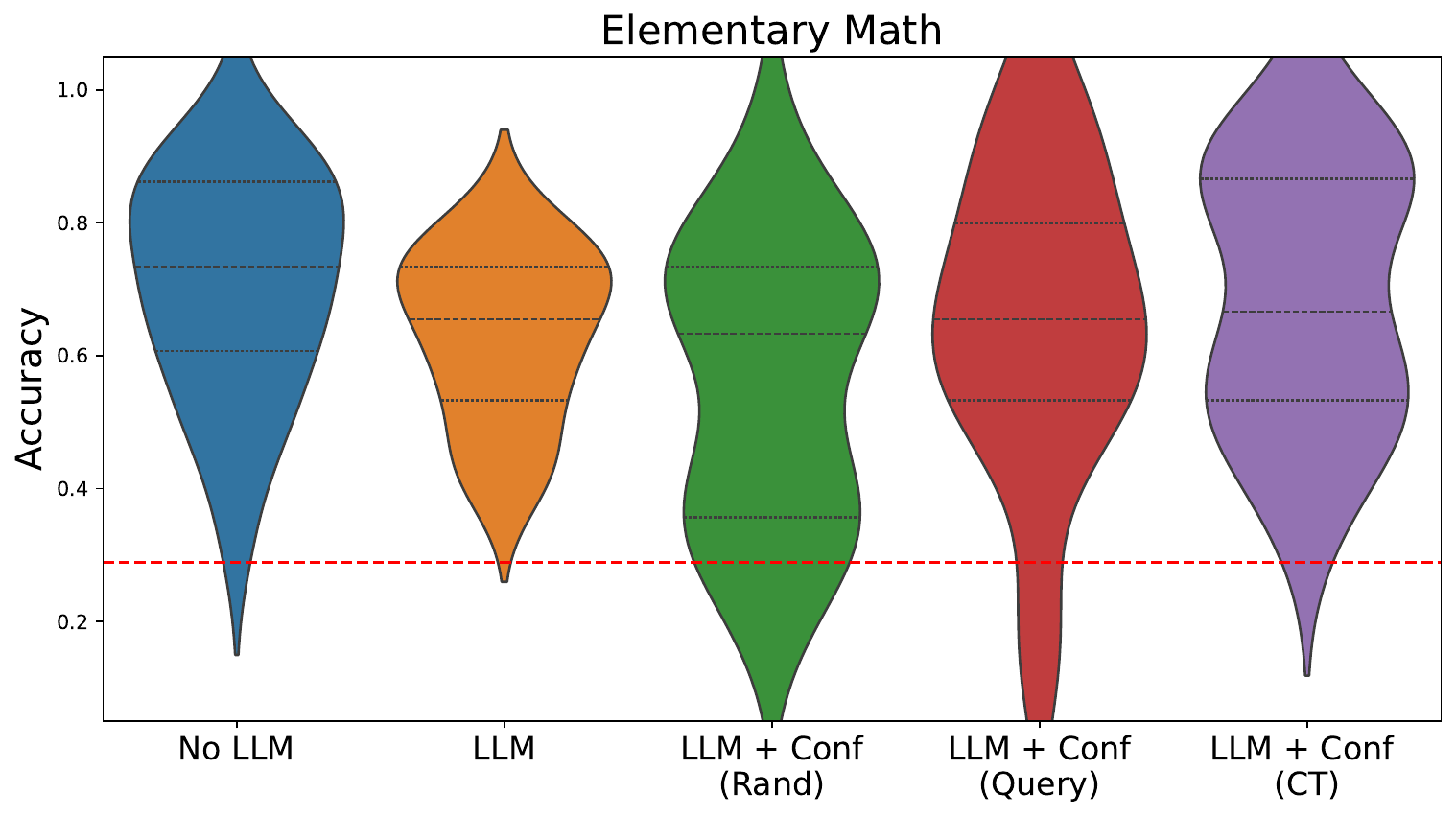}
    \caption{
    User accuracies per topic for the GPT-3.5 variants (with generalization confidence computed for the CT and Query cases). Red line indicates the model's average accuracy.}
    \label{fig:per-topic-acc-gpt35}
\end{figure*}

\begin{figure*}
    \centering
    \includegraphics[width=0.6\linewidth]{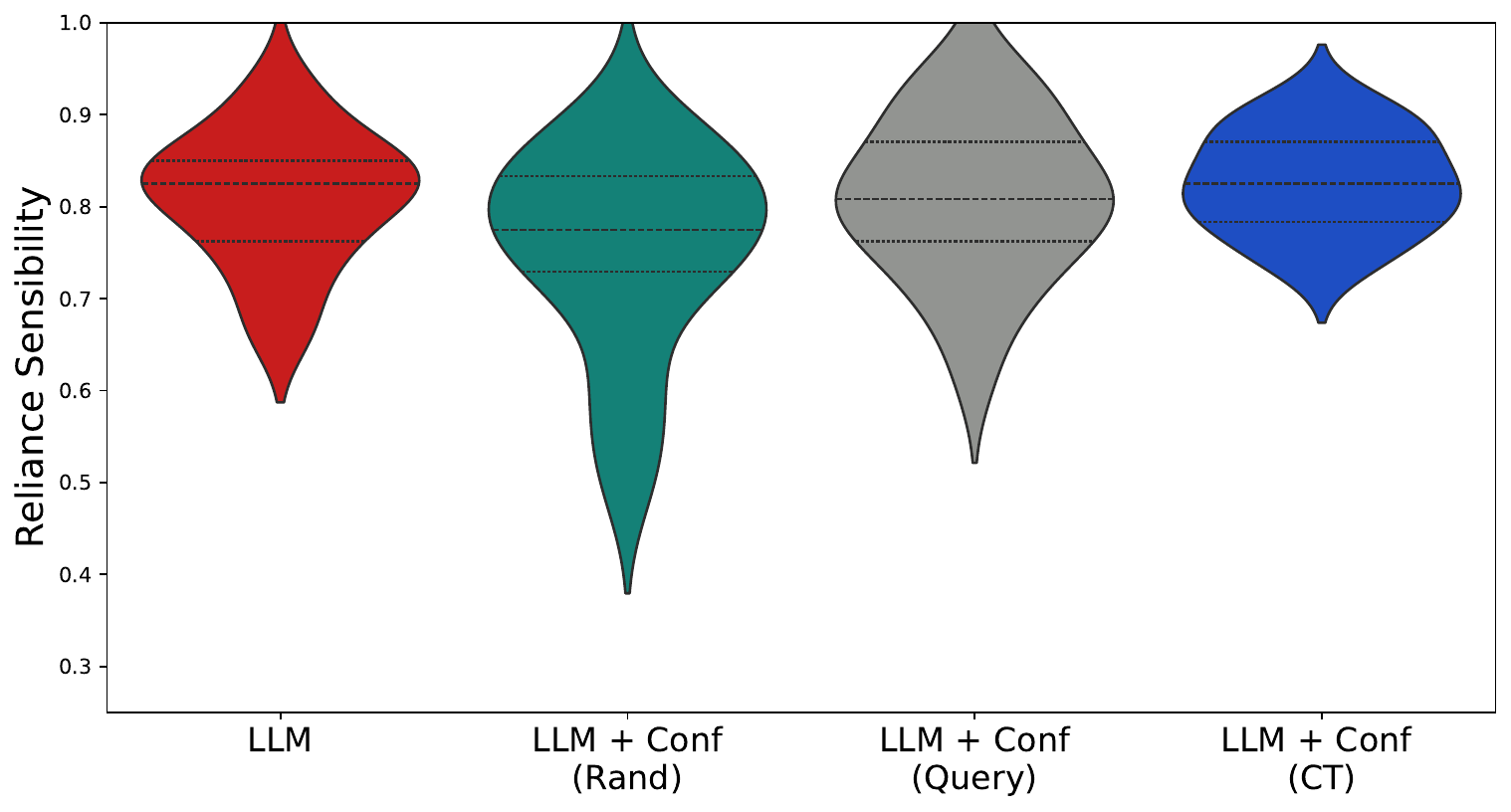}
    \caption{
    Reliance sensibility for the variants based on GPT-3.5}
    \label{fig:reliance-gpt35}
\end{figure*}

As noted, we ran variants using the same GPT-3.5 generations as \citep{bhatt2023learning}. We show aggregate and per-topic accuracy in \cref{fig:per-topic-acc-gpt35}, as well as reliance sensibility in \cref{fig:reliance-gpt35}. 

\paragraph{Freeform User Responses}

We permitted users to provide freeform responses at the end of the study. Some users were sensitive to confidence scores being reported and came up with their own heuristics for whether to rely on the model's output. We include a sampling of comments across confidence variants: 

\begin{itemize}
    \item ``if it had a confidence of less than 50\% it made me very skeptical.'' 
    \item ``The model\'s confidence indeed helped me choose and select my answer as I trusted in them most of the time.''
    \item ``I didn\'t really rely on the confidence level. If I had 0 confidence in the answer myself I relied on the AI regardless.''
    \item ``if the models confidence fell below 45 I decided to investigate it myself by remembering pieces of information. and also reasoning the question. If it was above 45 I would automatically agree to its prediction but there were some few cases I  challenged it even though it was above 45''
    \item ``At first I was hesistant to trust the model much because of the lower confidence levels but I still trusted it enough on topics I struggled with. As it went on, I was comfortable with confidence levels above 40.''
    \item ``If the model\'s confidence was low and I thought I knew the answer (and it was different) I chose my answer''
\end{itemize}

\subsection{Interface and Instructions}
\label{app:user-study-interfact}

\begin{figure}
    \centering
    \includegraphics[width=0.7\linewidth]{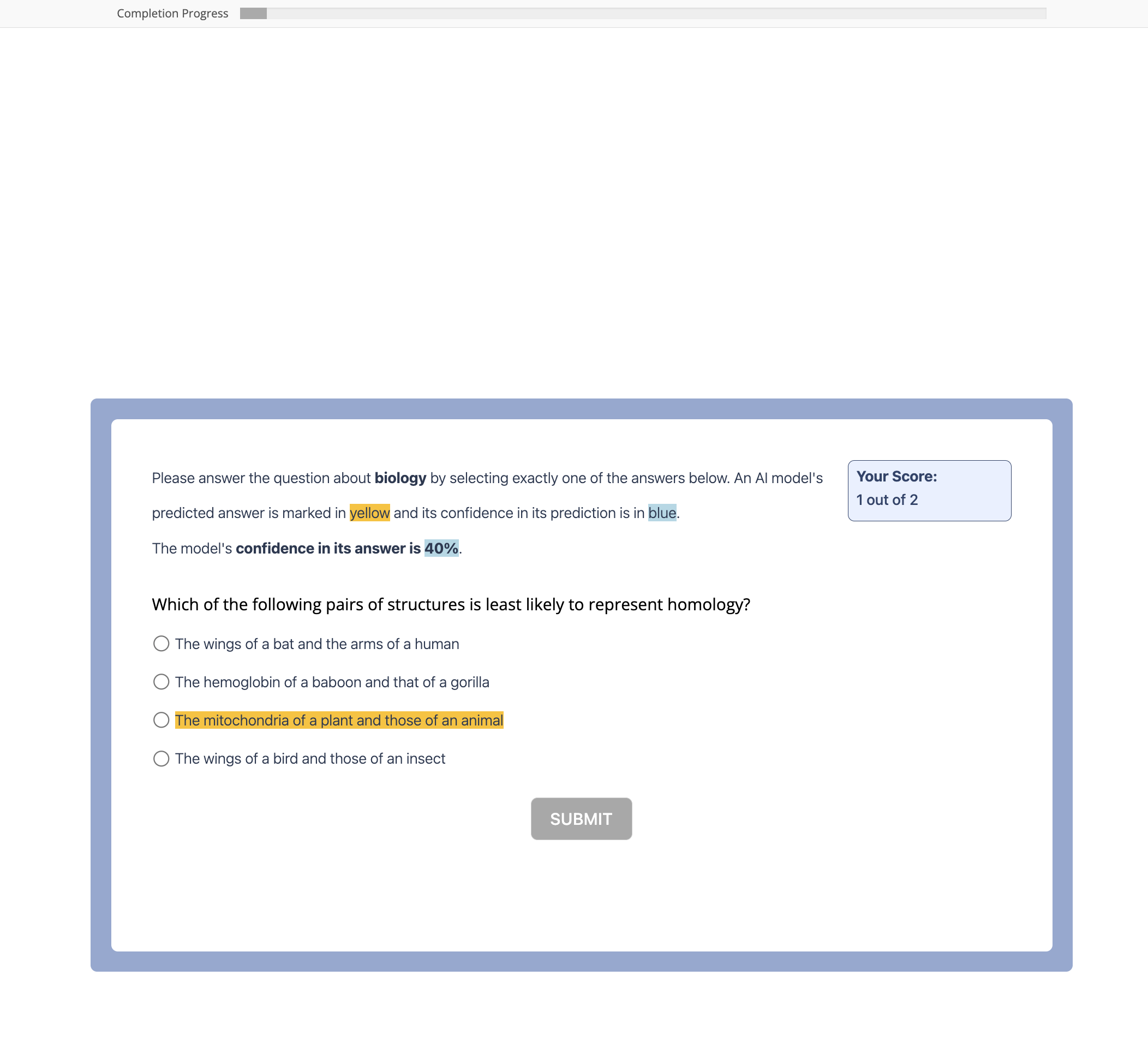}
    \caption{Example interface from \texttt{Modiste}. Participants are informed of the question (and topic), as well as the LLM prediction and confidence. Participants are informed of their running score throughout the experiment.}
    \label{fig:modiste-confidence}
\end{figure}

\begin{figure}
    \centering
    \includegraphics[width=0.8\linewidth]{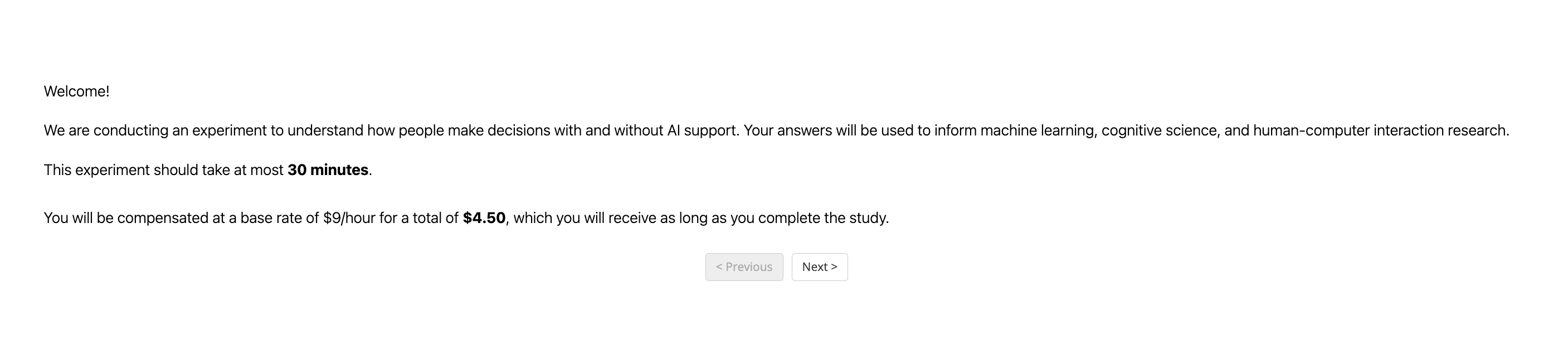}
    \includegraphics[width=0.8\linewidth]{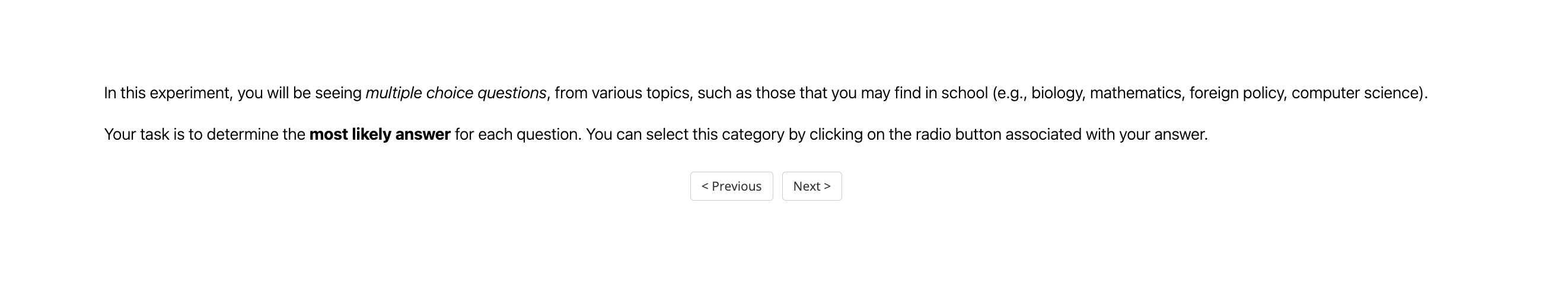}
    \includegraphics[width=0.8\linewidth]{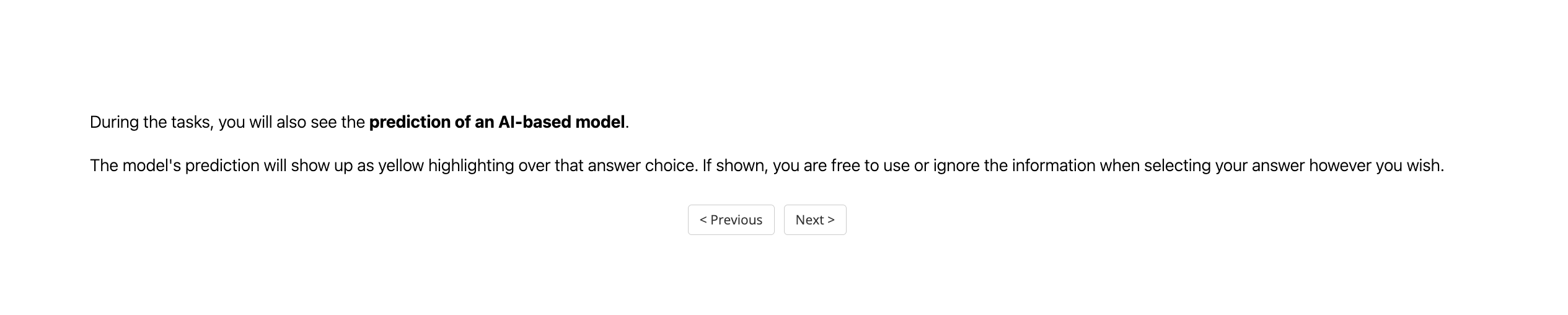}
    \includegraphics[width=0.8\linewidth]{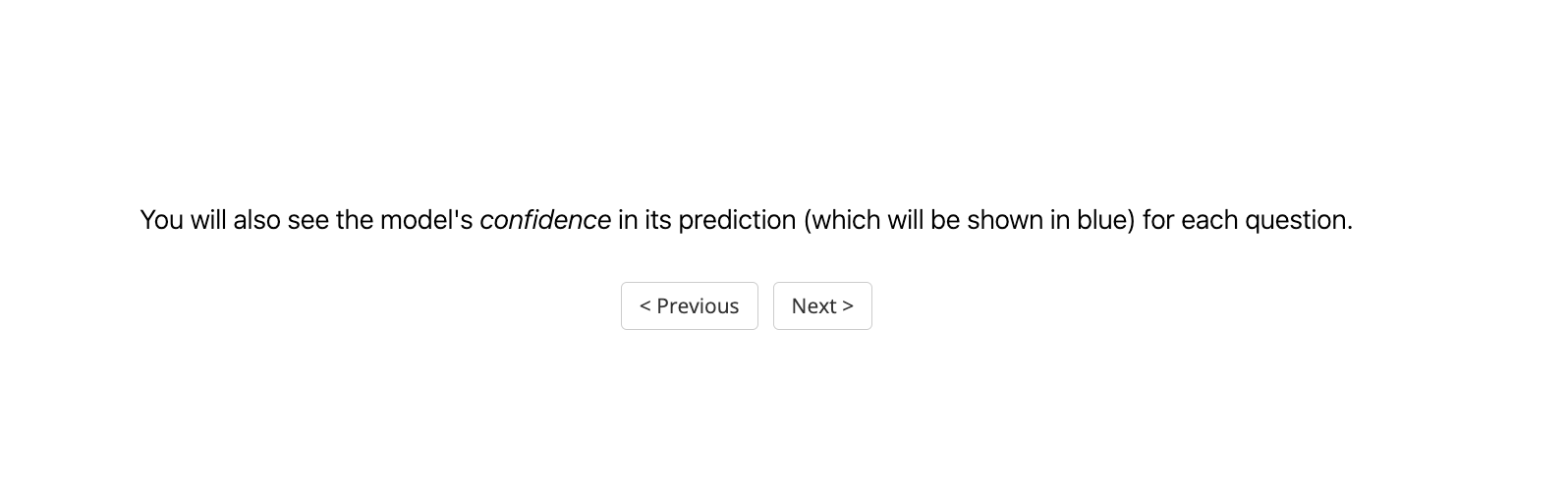}
    \includegraphics[width=0.8\linewidth]{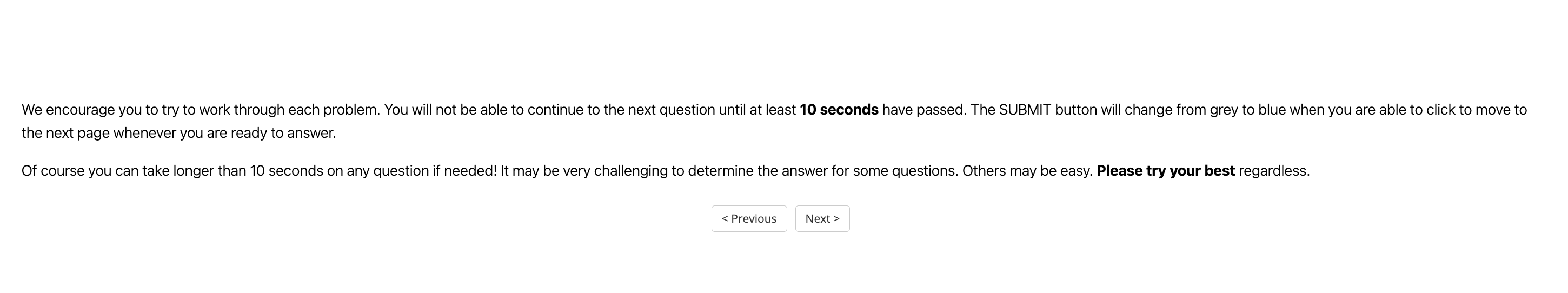}
    \includegraphics[width=0.8\linewidth]{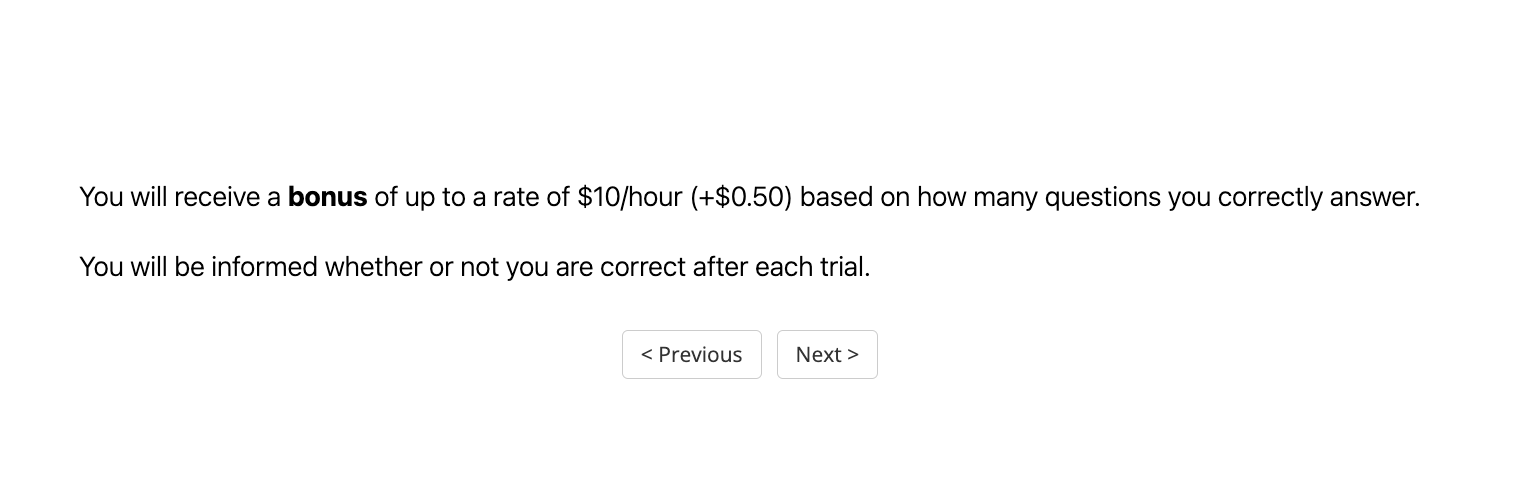}
    \caption{Experiment instructions for the confidence variants.}
    \label{fig:experiment-instructions}
\end{figure}

\begin{figure}
    \centering
    
    \includegraphics[width=0.8\linewidth]{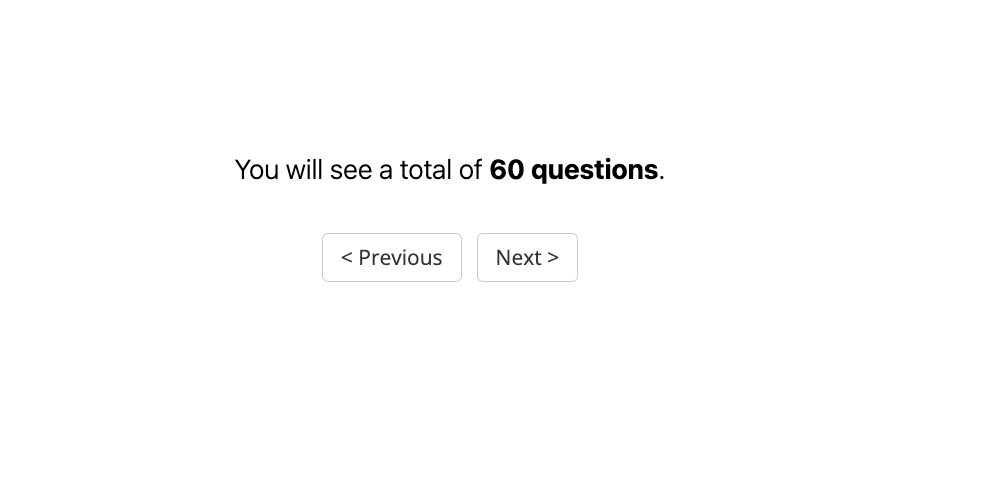}
    \includegraphics[width=0.8\linewidth]{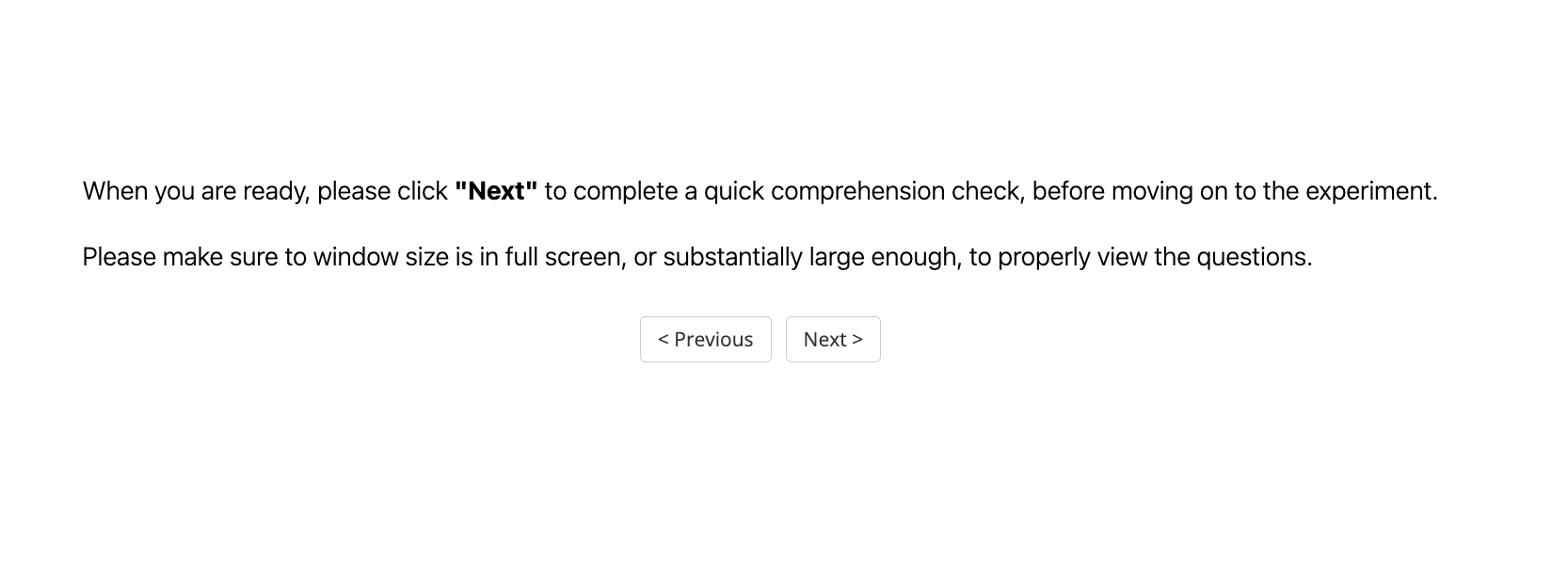}
    \includegraphics[width=0.8\linewidth]{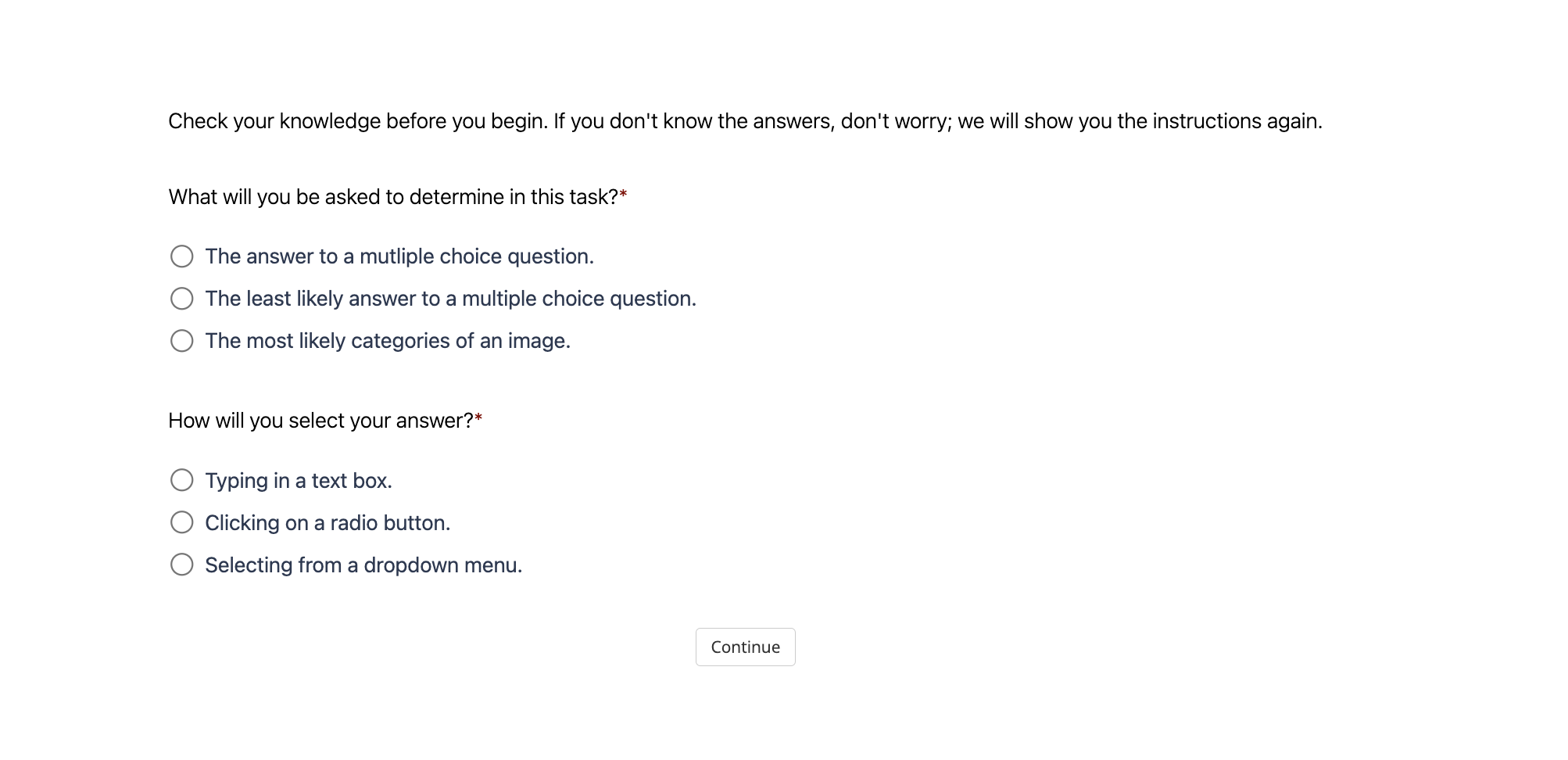}
    \caption{Experiment instructions for the confidence variants (continued).}
    \label{fig:experiment-instructions-2}
\end{figure}

\begin{figure}
    \centering
    \includegraphics[width=0.9\linewidth]{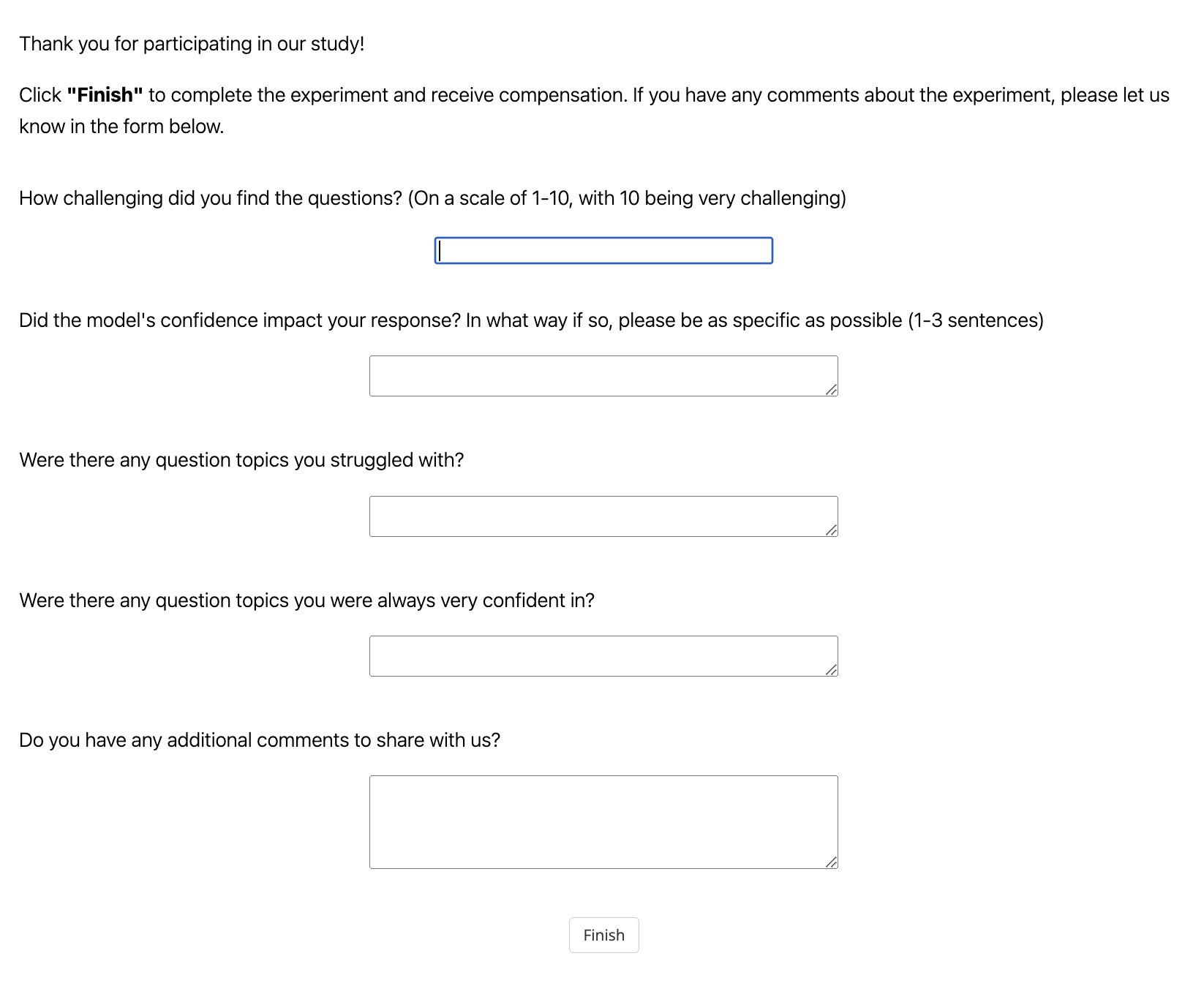}
    \caption{Sample pot-survey questionnaire for users who were allocated to a variant wherein they saw model confidence.}
    \label{fig:postquestionarre}
\end{figure}

We show a sample interface of our extension of \texttt{Modiste} with user confidence in Figure \ref{fig:modiste-confidence}, and present the the full set of instructions provided to users in Figures \ref{fig:experiment-instructions} and \ref{fig:experiment-instructions-2}. Note, for the LLM-only and no-LLM conditions, we followed the instruction text from \citep{bhatt2023learning} directly, i.e., participants who saw only the LLM did not see the instruction page about model confidence, and participants in the ``No-LLM'' variant were not instructed about any model variant and were just instructed to answer the questions as best as they could by themselves. Participants also responded to a post survey questionarre after completing the user study, which we depict in Figure \ref{fig:postquestionarre}.

\section{Broader Impact and Implications}
\label{app:broader-impact}

The goal of this work is to make LLM outputs have better confidence values associated with them. 
With successful, calibrated confidence values, the machine systems ultimately become more interpretable and trustworthy by a user~\citep{janssen2008updating}. 
When applied correctly, our advancements will help users be able to make decisions based off of LLM outputs in a more informed way.
Similar examples in other domains, like AlphaFold~\cite{terwilliger2023alphafold}, have shown how well-calibrated confidence scores can be useful in complex decision-making domains.
Our hope is to replicate those broad findings in LLMs.

We acknowledge the ongoing debate over the appropriateness, limitations, and harms of LLMs. 
We do highlight that the development of more confident, interpretable, and trustworthy LLMs can lead to continued techno-solutionism in unintended applications. 
Specifically, we highlight that our work is limited to use-cases with fact-based questions.
Many applications of text-based LLMs are generative, meaning that there is no way for our paradigm to be applied appropriately, and the use of a confidences from calibration-tuned models could be misleading or damaging without checks and guardrails.
Additionally, even within the fact-based paradigm, what is true can be subjective, with ground truth in machine learning being a contested topic~\citep{aroyo2015truth, uma2021learning}.

The philosophical debate on these topics is beyond the expertise of the authors; nonetheless, we believe that the ongoing debate over the appropriateness of LLMs should be considered in context with the benefits of our approach in making LLMs more interpretable and useful.

\clearpage
\section{NeurIPS Paper Checklist}

\begin{enumerate}

\item {\bf Claims}
    \item[] Question: Do the main claims made in the abstract and introduction accurately reflect the paper's contributions and scope?
    \item[] Answer: \answerYes{}
    \item[] Justification: We describe and link all claims in \cref{sec:intro}.
    
\item {\bf Limitations}
    \item[] Question: Does the paper discuss the limitations of the work performed by the authors?
    \item[] Answer: \answerYes{}
    \item[] Justification: We provide a discussion on the limitations in \cref{sec:discussion}.
    \begin{itemize}
        \item The answer NA means that the paper has no limitation while the answer No means that the paper has limitations, but those are not discussed in the paper. 
        \item The authors are encouraged to create a separate "Limitations" section in their paper.
        \item The paper should point out any strong assumptions and how robust the results are to violations of these assumptions (e.g., independence assumptions, noiseless settings, model well-specification, asymptotic approximations only holding locally). The authors should reflect on how these assumptions might be violated in practice and what the implications would be.
        \item The authors should reflect on the scope of the claims made, e.g., if the approach was only tested on a few datasets or with a few runs. In general, empirical results often depend on implicit assumptions, which should be articulated.
        \item The authors should reflect on the factors that influence the performance of the approach. For example, a facial recognition algorithm may perform poorly when image resolution is low or images are taken in low lighting. Or a speech-to-text system might not be used reliably to provide closed captions for online lectures because it fails to handle technical jargon.
        \item The authors should discuss the computational efficiency of the proposed algorithms and how they scale with dataset size.
        \item If applicable, the authors should discuss possible limitations of their approach to address problems of privacy and fairness.
        \item While the authors might fear that complete honesty about limitations might be used by reviewers as grounds for rejection, a worse outcome might be that reviewers discover limitations that aren't acknowledged in the paper. The authors should use their best judgment and recognize that individual actions in favor of transparency play an important role in developing norms that preserve the integrity of the community. Reviewers will be specifically instructed to not penalize honesty concerning limitations.
    \end{itemize}

\item {\bf Theory Assumptions and Proofs}
    \item[] Question: For each theoretical result, does the paper provide the full set of assumptions and a complete (and correct) proof?
    \item[] Answer: \answerNA{} 
    \item[] Justification: \answerNA{}
    \begin{itemize}
        \item The answer NA means that the paper does not include theoretical results. 
        \item All the theorems, formulas, and proofs in the paper should be numbered and cross-referenced.
        \item All assumptions should be clearly stated or referenced in the statement of any theorems.
        \item The proofs can either appear in the main paper or the supplemental material, but if they appear in the supplemental material, the authors are encouraged to provide a short proof sketch to provide intuition. 
        \item Inversely, any informal proof provided in the core of the paper should be complemented by formal proofs provided in appendix or supplemental material.
        \item Theorems and Lemmas that the proof relies upon should be properly referenced. 
    \end{itemize}
    
    \item {\bf Experimental Result Reproducibility}
    \item[] Question: Does the paper fully disclose all the information needed to reproduce the main experimental results of the paper to the extent that it affects the main claims and/or conclusions of the paper (regardless of whether the code and data are provided or not)?
    \item[] Answer: \answerYes{}
    \item[] Justification: We provide the complete code, and the complete list of datasets used for all experiments in \cref{sec:ut} to reproduce all our experiments with instructions. All hyperparameters are described in \cref{sec:ut}.

\item {\bf Open access to data and code}
    \item[] Question: Does the paper provide open access to the data and code, with sufficient instructions to faithfully reproduce the main experimental results, as described in supplemental material?
    \item[] Answer: \answerYes{}
    \item[] Justification: We provide the complete code, and the complete list of datasets used for all experiments in \cref{app:training-data} to reproduce all our experiments with instructions. All hyperparameters are described in \cref{sec:ut}.

\item {\bf Experimental Setting/Details}
    \item[] Question: Does the paper specify all the training and test details (e.g., data splits, hyperparameters, how they were chosen, type of optimizer, etc.) necessary to understand the results?
    \item[] Answer: \answerYes{}
    \item[] Justification: We provide the complete code, and the complete list of datasets used for all experiments in \cref{app:training-data} to reproduce all our experiments with instructions. All hyperparameters are described in \cref{sec:ut}.

\item {\bf Experiment Statistical Significance}
    \item[] Question: Does the paper report error bars suitably and correctly defined or other appropriate information about the statistical significance of the experiments?
    \item[] Answer: \answerYes{}
    \item[] Justification: All figures are appropriately labeled with the error bars.

\item {\bf Experiments Compute Resources}
    \item[] Question: For each experiment, does the paper provide sufficient information on the computer resources (type of compute workers, memory, time of execution) needed to reproduce the experiments?
    \item[] Answer: \answerYes{}
    \item[] Justification: We provide an estimate of the compute resources required in \cref{sec:ut}.
    
\item {\bf Code Of Ethics}
    \item[] Question: Does the research conducted in the paper conform, in every respect, with the NeurIPS Code of Ethics \url{https://neurips.cc/public/EthicsGuidelines}?
    \item[] Answer: \answerYes{}
    \item[] Justification: We have read the ethics guidelines

\item {\bf Broader Impacts}
    \item[] Question: Does the paper discuss both potential positive societal impacts and negative societal impacts of the work performed?
    \item[] Answer: \answerYes{}
    \item[] Justification: We provide a broader impact statement in \cref{app:broader-impact}
    
\item {\bf Safeguards}
    \item[] Question: Does the paper describe safeguards that have been put in place for responsible release of data or models that have a high risk for misuse (e.g., pretrained language models, image generators, or scraped datasets)?
    \item[] Answer: \answerNA{} 
    \item[] Justification: We train on open-access models with open-source datasets. We do not change their generation behavior, and all existing safeguards (if any) remain.

\item {\bf Licenses for existing assets}
    \item[] Question: Are the creators or original owners of assets (e.g., code, data, models), used in the paper, properly credited and are the license and terms of use explicitly mentioned and properly respected?
    \item[] Answer: \answerYes{}
    \item[] Justification: We explicitly cite all models in \cref{sec:ut}. All datasets used are listed and cited in \cref{app:training-data}.

\item {\bf New Assets}
    \item[] Question: Are new assets introduced in the paper well documented and is the documentation provided alongside the assets?
    \item[] Answer: \answerYes{}
    \item[] Justification: We release our trained models for easy use via Hugging Face.

\item {\bf Crowdsourcing and Research with Human Subjects}
    \item[] Question: For crowdsourcing experiments and research with human subjects, does the paper include the full text of instructions given to participants and screenshots, if applicable, as well as details about compensation (if any)? 
    \item[] Answer: \answerYes{} 
    \item[] Justification: We provide screenshots of our instructions, as well as details of compensation in \cref{app:user-study}.

\item {\bf Institutional Review Board (IRB) Approvals or Equivalent for Research with Human Subjects}
    \item[] Question: Does the paper describe potential risks incurred by study participants, whether such risks were disclosed to the subjects, and whether Institutional Review Board (IRB) approvals (or an equivalent approval/review based on the requirements of your country or institution) were obtained?
    \item[] Answer: \answerYes{} 
    \item[] Justification: We received prior approval from our respective institutional ethics review body for our user study. All users provided consent before partaking in the study. 

\end{enumerate}

\end{document}